%% file: iclr2023_conference.tex
\documentclass{article} 
\usepackage{iclr2023_conference,times}

\input{math_commands.tex}

\usepackage[hidelinks]{hyperref}
\usepackage{url}

\usepackage{cleveref}
\crefname{figure}{Fig.}{Fig.}
\crefname{subfigure}{Fig.}{Fig.}
\crefformat{section}{\S#2#1#3}
\crefformat{subsection}{\S#2#1#3}
\crefformat{subsubsection}{\S#2#1#3}
\crefmultiformat{section}{\S#2#1#3}{ and~\S#2#1#3}{, \S#2#1#3}{, and~\S#2#1#3}
\crefmultiformat{subsection}{\S#2#1#3}{ and~\S#2#1#3}{, \S#2#1#3}{, and~\S#2#1#3}
\crefmultiformat{subsubsection}{\S#2#1#3}{ and~\S#2#1#3}{, \S#2#1#3}{, and~\S#2#1#3}
\crefrangeformat{section}{\mbox{\S#3#1#4--#5#2#6}}
\crefrangeformat{subsection}{\mbox{\S#3#1#4--#5#2#6}}
\crefrangeformat{subsubsection}{\mbox{\S#3#1#4--#5#2#6}}
\crefname{appendix}{Appendix}{Appendices}
\crefname{table}{Table}{Tables}
\crefname{equation}{Eq.}{Eq.}
\usepackage{graphicx}
\usepackage{subcaption}
\usepackage{multirow}
\usepackage{booktabs}
\usepackage{notoccite}
\title{A Curriculum View of Robust Loss Functions}

\author{
Zebin Ou\textsuperscript{1}\thanks{Work mainly done at Westlake University.} \ \& Yue Zhang\textsuperscript{2} \\
\textsuperscript{1}Zhihu Inc., \textsuperscript{2}School of Engineering, Westlake University\\
\texttt{ouzebin@zhihu.com, yuezhang@westlake.edu.cn}
}

\iclrfinalcopy 
\begin{document}

\maketitle
\begin{abstract}
  Robust loss functions are designed to combat the adverse impacts of label noise, whose robustness is typically supported by theoretical bounds agnostic to the training dynamics. However, these bounds may fail to characterize the empirical performance as it remains unclear why robust loss functions can underfit. We show that most loss functions can be rewritten into a form with the same class-score margin and different sample-weighting functions. The resulting curriculum view provides a straightforward analysis of the training dynamics, which helps attribute underfitting to diminished average sample weights and noise robustness to larger weights for clean samples. We show that simple fixes to the curriculums can make underfitting robust loss functions competitive with the state-of-the-art, and training schedules can substantially affect the noise robustness even with robust loss functions. Code is available at \url{github}.
\end{abstract}
\vspace{-1mm}
\section{Introduction}
\vspace{-1mm}
Label noise is non-negligible in automatic annotation \citep{ner_noise}, crowd-sourcing \citep{imagenet} and expert annotation \citep{treebank_noise}. Their adverse impacts can be mitigated with loss functions that are theoretically robust against label noise \citep{noise_survey}, whose robustness is supported by bounding the difference between expected risk minimizers obtained with clean or noisy labels \citep{mae,asymmetric}. However, these bounds do not consider how minimizers are approached or whether they apply to empirically obtained local optimums, leaving open questions on the empirical performance. For example, robust loss functions can underfit \citep{gce,sce} difficult tasks, but the underlying reason cannot be derived from these bounds.

To address such limitations, we analyze training dynamics towards the risk minimizers to characterize the empirical performance of different loss functions. We rewrite loss functions into a form with equivalent gradients, which consists of \emph{the same} class-score margin and \emph{different} sample-weighting functions. The former is a lower bound of the score difference between the labeled class and any other classes, which determines the \emph{direction} of the sample gradient. The latter \emph{weight} samples based on their class-score margins. Interactions between distributions of class-score margins and sample weighting functions thus reveal aspects of the training dynamics. Notably, loss functions with our derived form \emph{implicitly} define different \emph{sample-weighting curriculums} (\cref{sec:curriculum}). Here a curriculum, by definition \citep{curriculum}, specifies a sequence of re-weighting for the distribution of training samples, e.g., sample weighting \citep{method:weight1} or sample selection \citep{method:sample}, based on a metric for sample difficulty.

We first attribute the underfitting issue to diminished average sample weights during training (\cref{sec:dynamics:underfit}). In particular, classification with more classes lead to smaller class-score margins at initialization, which can lead to minimal sample weights given some robust loss functions. Robust loss functions that severely underfit can become competitive with the best-performing ones by adapting the sample-weighting functions to the number of classes. We then attribute the noise robustness of loss functions to larger weights for clean samples than noise samples (\cref{sec:dynamics:robust}). We find that dynamics of SGD suppress the learning of noise samples or even get them unlearned. The sample-weighting functions of robust loss functions then magnify differences in the learning pace between clean and noise samples and neglect the unlearned noise samples. Finally, to support our understanding of the training dynamics, we present surprising results that deviate from existing theoretical bounds: by simply modifying the learning rate schedule, (1) robust loss functions can become vulnerable to label noise, and (2) cross entropy can appear robust.

\section{Related Work}

Most existing studies on robust loss functions \citep{mae,gce,sce,taylor,peer,peer2,asymmetric} focus on deriving bounds of the difference between risk minimizers obtained with noisy and clean labels, which are agnostic to the training dynamics. We instead analyze the training dynamics of robust loss functions for reasons behind their underfitting and noise robustness. Although the underfitting issue has been heuristically mitigated with loss combination \citep{gce,sce,actpass}, we aim to explicitly identify the cause and support it with our fixes.

Our curriculum view connects existing robust loss functions to the seemingly distinct \citep{noise_survey} curriculum learning. To mitigate the adverse impacts of noisy labels, curriculum-based approaches use either sample selection \citep{method:sample2,o2u,method:sample} or sample weighting \citep{method:weight1,method:weight3,imae,dm}. Our work is related to studies of sample-weighting curriculums but differs in four perspectives. First, the sample weights analyzed in our work are \emph{implicitly} defined by robust loss functions rather than \emph{explicitly} designed \citep{method:weight1,imae,dm} or predicted by a model \citep{method:weight2,method:weight3}. Second, the sample difficulty metric of the implicit sample-weighting curriculums is the class-score margin we derived rather than common ones based on loss values \citep{curriculum:easy,curriculum:hard2} or gradient magnitudes \citep{curriculum:hardness}.
Third, instead of designing better sample-weighting curriculums \citep{method:weight1,imae,dm}, we focus on characterizing the performance of \emph{existing} robust loss functions by analyzing their implicit sample-weighting curriculums. Finally, although existing work \citep{method:weight2} provides a robust-loss-function view of some sample-weighting curriculums by identifying their effective loss functions, we focus on the curriculum view of \emph{existing} robust loss functions.

Our work is also related to the ongoing debate \citep{curriculum:easyhard,curriculum} on the strategies to select or weight samples in curriculum learning: either easier first \citep{curriculum_bengio,curriculum:easy} or harder first \citep{curriculum:hard2,curriculum:hard}. In contrast, the sample-weighting functions we identified in robust loss functions can be viewed as a combination of both strategies, emphasizing samples with moderate difficulty.

\vspace{-1mm}
\section{Background}
\vspace{-1mm}
\label{sec:background}

We formulate classification with label noise and noise robustness and briefly review existing research on robust loss functions before diving into our curriculum view.

\textbf{Classification} $k$-ary classification with input $\bm{x} \in \mathbb{R}^{d}$ can be solved with classifier $\argmax_i s_i$, where $s_i$ is the score of the $i$-th class in scoring function $\bm{s}_{\bm{\theta}}: \mathbb{R}^{d} \to \mathbb{R}^{k}$ parameterized by $\bm{\theta}$. $s_i$ can be converted into probability $p_i$ with the softmax function $p_i = e^{s_i}/\sum_{j} e^{s_j}$. Given the ground truth label $y^* \in \{ 1 .. k\}$ for $\bm{x}$ and a loss function $L(\bm{s}_{\bm{\theta}}(\bm{x}), y^*)$, we can estimate $\bm{\theta}$ with risk minimization $\argmin_{\bm{\theta}} \mathbb{E}_{\bm{x}, y^*} [L(\bm{s}_{\bm{\theta}}(\bm{x}), y^*)]$.

\textbf{Noise robustness\ } Labeling errors can corrupt the ground truth label $y^*$ into a noisy one,
\begin{equation*}
y = \left\{\begin{array}{l l}
                     y^*,            & \textrm{with probability } P(y=y^* | \bm{x}, y^*) \\
                     i, i \neq y^* & \textrm{with probability } P(y=i | \bm{x}, y^*)
                   \end{array}
            \right.
\end{equation*}
Samples $(\bm{x}, y)$ with noisy label $y$ are clean samples if $y = y^*$ and noise samples otherwise. Following \citet{mae}, label noise is \emph{symmetric} (or uniform) if $P(\tilde{y}=i | \bm{x}, y) = \eta / (k-1), \forall i \neq y$ and \emph{asymmetric} (or class-conditional) when $P(\tilde{y}=i | \bm{x}, y) = P(\tilde{y}=i | y)$, where $\eta = P(\tilde{y} \neq y)$ is the noise rate. Given a noisy label $\tilde{y}$ for $\bm{x}$, a loss function $L$ is robust against label noise if
\begin{equation*}
\argmin\nolimits_{\bm{\theta}} \mathbb{E}_{\bm{x}, \tilde{y}} L(\bm{s}_{\bm{\theta}}(\bm{x}),\tilde{y}) \approx \argmin\nolimits_{\bm{\theta}} \mathbb{E}_{\bm{x}, y} L(\bm{s}_{\bm{\theta}}(\bm{x}), y)
\label{eq:robustness}
\end{equation*}
approximately holds. We rewrite $\bm{s}_{\bm{\theta}}(\bm{x})$ into $\bm{s}$ for notation simplicity hereafter.

\subsection{Typical Robust Loss Functions}
\label{sec:background:loss}
We briefly review typical robust loss functions besides cross entropy (CE) that is vulnerable \citep{mae} to label noise. See \cref{table:weight} for the formulae and \cref{app:losses} for the derivations and extra loss functions.

\textbf{Symmetric\ } A loss function $L$ is \emph{symmetric} \citep{mae} if
\begin{equation*}
\sum_{i=1}^k L(\bm{s}, i) = C, \ \forall \bm{s} \in \mathbb{R}^{k},
\label{eq:symmetric}
\end{equation*}
with a constant $C$. It is robust against \emph{symmetric} label noise when $\eta < (k-1)/k$. Mean absolute error (MAE; \citealp{mae}) and the equivalent reverse cross entropy (RCE; \citealp{sce}) are both symmetric. \citet{actpass} make loss functions with $L(\bm{s}, i) > 0, \forall i \in \{1 .. k\}$ symmetric by normalizing them, $L'(\bm{s}, y) = L(\bm{s}, y)/\sum_{i} L(\bm{s}, i)$. We include normalized cross entropy (NCE; \citealt{actpass}) as an example.

\textbf{Asymmetric\ } $L$ as a function of the softmax probability $p_i$, $L(\bm{s}, i) = l(p_i)$, is \emph{asymmetric} \citep{asymmetric} if
\begin{equation*}
  \tilde{r} = \max_{i\neq y} \frac{P(\tilde{y}=i | \bm{x}, y)}{P(\tilde{y}=y | \bm{x}, y)} \leq \inf_{\substack{0 \leq p_i, p_j \leq 1\\p_i + p_j \leq 1}} \frac{l(p_i) - l(p_i +  p_j)}{l(0) - l(p_j)},
\label{eq:asymmetric}
\end{equation*}
where $p_j$ is a valid increment of $p_i$. An asymmetric loss function is robust against \emph{generic} label noise when $\tilde{r} < 1$, i.e., there are more clean samples than noisy samples. We include asymmetric generalized cross entropy (AGCE) and asymmetric unhinged loss (AUL) from \citet{asymmetric}.

\textbf{Combined\ } Robust loss functions that underfit \citep{gce} can be combined with loss functions like CE to balance robustness and sufficient learning. For example, generalized cross entropy (GCE; \citealp{gce}) is a smooth interpolation between CE and MAE. Alternatively, symmetric cross entropy (SCE; \citealp{sce}) is a weighted average of CE and RCE (MAE). \citet{actpass} propose to combine active and passive loss functions with weighted average. We include NCE+MAE as an example.

\subsection{Explaining Underfitting of Robust Loss Functions}
\label{sec:background:question}
Despite the theoretical bounds for noise robustness, robust loss functions such as MAE can underfit difficult tasks \citep{gce}. Existing explanations can be limited. \citet{gce} attribute underfitting of MAE to the lack of the $1/p_y$ term in the sample gradient compared to CE, thus ``treating every sample equally'' and hampering learning. However, we show that MAE instead emphasizes samples with moderate class-score margins after factoring sample gradients into weights and directions. \citet{actpass} attribute underfitting to failure in balancing the active-passive components. They rewrite loss functions into $L(\bm{s}, y) = \sum_{i} l(\bm{s}, i)$ where $i \in \{1 .. k\}$ is an arbitrary class, and define active loss functions with $\forall i \neq y, \ l(\bm{s}, i) = 0$, which emphasizes learning the labeled class $y$. In contrast, passive loss functions defined with $\exists i \neq y, \ l(\bm{s}, i) \neq 0$ can be improved by unlearning other classes $i \neq y$. However, since there is no canonical guideline to specify $l(\bm{s}, i)$, different specifications can lead to ambiguities. Given
\begin{equation*}
L_{\mathrm{MAE}}(\bm{s}, y) \propto \sum\nolimits_{i} |\mathbb{I}(i=y) - p_i| \propto \sum\nolimits_{i} \mathbb{I}(i=y) (1-p_i)
\end{equation*}
with indicator function $\mathbb{I}(\cdot)$, MAE is active if $l(\bm{s}, i) = \mathbb{I}(i=y) (1-p_i)$ but passive if $l(\bm{s}, i) = |\mathbb{I}(i=y) - p_i|$. Finally, \citet{imae} view $\|\nabla_{\bm{s}} L(\bm{s}, y)\|_1$ as weights for sample gradients and attribute underfitting to their low variance, making clean and noise samples less distinguishable. However, as shown in \cref{sec:dynamics:underfit}, MAE also underfits data with clean labels. We provide an alternative explanation with our curriculum view in \cref{sec:dynamics:underfit}, which leads to effective fixes for the issue.

\vspace{-0.8mm}
\section{Implicit Curriculums of Loss Functions}
\vspace{-0.8mm}
\label{sec:curriculum}
We derive the main results of our curriculum view for later analysis. The softmax probability $p_y$ of the (noisily) labeled class $y$ can be written into a sigmoid form,
\begin{equation*}
p_y =  \frac{e^{s_y}}{\sum_{i} e^{s_i}} = \frac{1}{e^{-(s_y -\log \sum_{i \neq y} e^{s_i})} +  1} = \frac{1}{e^{-\Delta_y} +  1},
\end{equation*}
where
\begin{equation*}
\Delta_y = s_y - \log \sum_{i \neq y} e^{s_i} \leq s_y - \max_{i \neq y}s_i
\end{equation*}
is the soft score margin between the labeled class $y$ and any other classes.
A large $\Delta_y$ indicates a well-learned sample as $\Delta_y \geq 0$ leads to successful classification with $y = \argmax_i s_i$. Note that loss functions in \cref{table:weight} except NCE and NCE+MAE are functions of $p_y$, $L(\bm{s}, y) = l(p_y)$. Given
\begin{align*}
\nabla_{\boldsymbol{s}} L(\boldsymbol{s}, y) &= \nabla_{\boldsymbol{s}}l(p_y) = \frac{\mathrm{d} l}{\mathrm{d} p_y} \frac{\mathrm{d} p_y}{\mathrm{d} \Delta_y} \cdot \nabla_{\boldsymbol{s}}\Delta_{y} = \nabla_{\boldsymbol{s}} \left [ \rho \left ( \frac{\mathrm{d} l}{\mathrm{d} p_y} \frac{\mathrm{d} p_y}{\mathrm{d} \Delta_y} \right ) \cdot \Delta_{y} \right ] = \nabla_{\boldsymbol{s}} \left [ w(\Delta_y) \cdot \Delta_{y} \right ],
\end{align*}

where $\rho(\cdot)$ is the stop-gradient operator, we can rewrite $L$ into a form with equivalent gradient,
\begin{equation}
  \tilde{L}(\boldsymbol{s}, y) = w(\Delta_y) \cdot \Delta_y,
\label{eq:normalized}
\end{equation}
where $w(\Delta_y)=\rho ( \frac{\mathrm{d} l}{\mathrm{d} p_y}\frac{\mathrm{d} p_y}{\mathrm{d} \Delta_y} ) \leq 0$ is the \emph{sample-weighting function}. A larger $|w(\Delta_y)|$ emphasizes more on increasing $\Delta_y$ of the sample. Since $\|\nabla_{\bm{s}} \Delta_y\|_1 = 2$, $\Delta_y$ determines the direction of sample gradients. Thus \cref{eq:normalized} essentially factorizes the weight and direction of sample gradients.
Loss functions in the form of \cref{eq:normalized} differ only in the sample-weighting functions $w(\Delta_y)$, each implicitly defines a \emph{sample-weighting curriculum} based on $\Delta_y$ that reflects sample difficulty. Compared to existing sample-difficulty metrics like loss value \citep{curriculum:easy} or gradient magnitude \citep{curriculum:hardness}, $\Delta_y$ factors out the nonlinear preference of $w(\Delta_y)$ and can be a better drop-in replacement in curriculum-based approaches. The interactions between $w(\Delta_y)$ and distributions of $\Delta_y$ further reveal training dynamics with different loss functions, which facilitate our analysis in \cref{sec:dynamics}. See \cref{table:weight} for $w(\Delta_y)$ of the reviewed loss functions, and \cref{app:losses} for how hyperparameters affect $w(\Delta_y)$.

\begin{table}[t]
  \centering
  \small
  \renewcommand{\arraystretch}{1.8}
  \begin{tabular}{ c | c | c | c | c }
    Type                   & Name    & Function                                                   & Sample Weight $w$                                                                           & Constraints      \\
    \midrule
           /                & CE      & $-\log p_y$                                                & $1-p_y$                                                                                     &           /       \\
    \midrule
    \multirow{2}{*}{Sym.}  & MAE/RCE & $1- p_y$                                                   & $p_y(1-p_y)$                                                                                &    /              \\
                           & NCE     & $ -\log p_y/ \left(\sum_{i=1}^{k}-\log p_i\right)$              &  / &              /    \\
    \midrule
    \multirow{2}{*}{Asym.} & AUL     & $[(a - p_y)^q - (a-1)^q]/q$                          & $p_y(1-p_y)(a - p_y)^{q-1}$                                                                 & $a > 1$, $q > 0$ \\
                           & AGCE    & $[(a+1) - (a + p_y)^q]/q$                            & $p_y(a+p_y)^{q-1}(1-p_y) $                                                                  & $a > 0$, $q > 0$ \\
    \midrule
    \multirow{3}{*}{Comb.} & GCE     & $(1 - p_y^q)/q$                                      & $p_y^q(1-p_y)$                                                                              & $ 0 < q \leq 1$  \\
                           & SCE     & $ (1-q) \cdot L_{\mathrm{CE}} + q \cdot L_{\mathrm{MAE}}$  & $ (1-q + q \cdot p_y )(1-p_y)$                                                              & $0 < q < 1$      \\
                           & NCE+MAE & $ (1-q) \cdot L_{\mathrm{NCE}} + q \cdot L_{\mathrm{MAE}}$ &       /                             & $0 < q < 1$      \\
  \end{tabular}
  \caption{Expressions, constraints and sample-weighting functions (\cref{sec:curriculum}) for loss functions in \cref{sec:background:loss}.}
  \renewcommand{\arraystretch}{1}
  \label{table:weight}
\end{table}
\subsection{The Additional Regularizer of NCE}
\label{sec:curriculum:nce}
NCE does not exactly follow \cref{eq:normalized} as it additionally depends on $p_i$ with class $i \neq y$.
However, with equivalent gradients, it can be rewritten into
\begin{equation}
\tilde{L}_{\mathrm{NCE}}(\bm{s}, y) = \gamma \cdot  L_{\mathrm{CE}}(\bm{s}, y) + \gamma \cdot  \epsilon \cdot  R_{\mathrm{NCE}}(\bm{s}),
\label{eq:curriculum:nce}
\end{equation}
with $\gamma = \rho(-1/\sum_{i}\log p_i)$ and $\epsilon = \rho(k \log p_y/\sum_{i}\log p_i)$ the weights and $\rho(\cdot)$ the stop-gradient operator. Both $\gamma$ and $\epsilon$ decrease as $\Delta_y$ increases. The first additive term in \cref{eq:curriculum:nce}
 is a \emph{primary loss function} following \cref{eq:normalized}, which defines a sample-weighting curriculum. The second is a \emph{regularizer}
\begin{equation*}
R_{\mathrm{NCE}}(\bm{s}) = \sum_{i=1}^{k} \frac{1}{k} \log p_i
\end{equation*}
that reduces the entropy of softmax outputs and decreases $\gamma$ and $\epsilon$.
Although training dynamics of NCE are complicated by the extra regularizer, the L1 norm of sample gradients can be bounded with
\begin{equation}
\|\nabla_{\bm{s}} L_{\mathrm{NCE}}(\bm{s}, y)\|_1  \leq 2 \gamma \cdot \left(1 + \epsilon \right) \cdot w_{\mathrm{CE}}
\label{eq:nce:weight}
\end{equation}
which helps explain why NCE can underfit in \cref{sec:dynamics:underfit}. See \cref{app:losses:regularizer} for  discussions on similar loss functions with an additional regularizer and detailed derivations.

\vspace{-0.8mm}
\section{Loss Functions with the Curriculum View}
\vspace{-0.8mm}
\label{sec:dynamics}
We examine the interaction between $w(\Delta_y)$ and $\Delta_y$ distributions to address questions in \cref{sec:background:question}. Results are reported on MNIST \citep{mnist} and CIFAR10/100 \citep{cifar} with synthetic symmetric and asymmetric label noise following \citet{actpass,asymmetric}. For real-world scenarios, we include CIFAR10/100 with human label noise \citep{cifarn} and the large-scale noisy dataset WebVision \citep{webvision}, which exhibit more complex noise patterns than symmetric and asymmetric label noise. Unlike standard settings, we scale $w(\Delta_y)$ to unit maximum to avoid complications, since hyperparameters of loss functions can change the scale of $w(\Delta_y)$, essentially adjusting the learning rate of SGD. See \cref{app:setting} for more experimental details.

\subsection{Understanding Underfitting of Robust Loss Functions}
\label{sec:dynamics:underfit}

We reproduce the underfitting issue without label noise in \cref{table:ulearn}. The hyperparameters of loss functions are tuned on CIFAR100 and listed in \cref{table:underfit_param} of \cref{app:dynamics:underfit}. CE outperforms NCE, AGCE, AUL and MAE by a nontrivial margin on CIFAR100. The less performant loss functions has smaller gap between training and testing performance, suggesting the issue of underfitting. In contrast, all loss functions performs equally well on CIFAR10.

\textbf{Marginal sample weights explains underfitting.} Since the same model fits each dataset well with CE in \cref{table:ulearn}, underfitting should result from insufficient parameter updates with the altered loss functions rather than inadequate model size. Based on our derivation, the average scale of parameter update up to $t$-th step can be estimated with
\begin{equation*}
  \alpha_t = \frac{\sum_{i=1}^{t} \sum_{\bm{s} \in  \mathcal{B}_i} \eta_i \cdot \|\nabla_{\bm{s}} \Delta_y\|_1}{\sum_{i=1}^{t} \eta_i \cdot |\mathcal{B}_{i}|}
\end{equation*}
where $\eta_i$ is the learning rate and $\mathcal{B}_i$ the sampled batch at training step $i$. As shown in \cref{table:ulearn}, at the final training step, a small $\alpha_t$ highly correlates with underfitting.

\begin{figure}
  \centering
  \begin{subfigure}[b]{0.48\textwidth}
    \centering
    \includegraphics[width=0.8\textwidth]{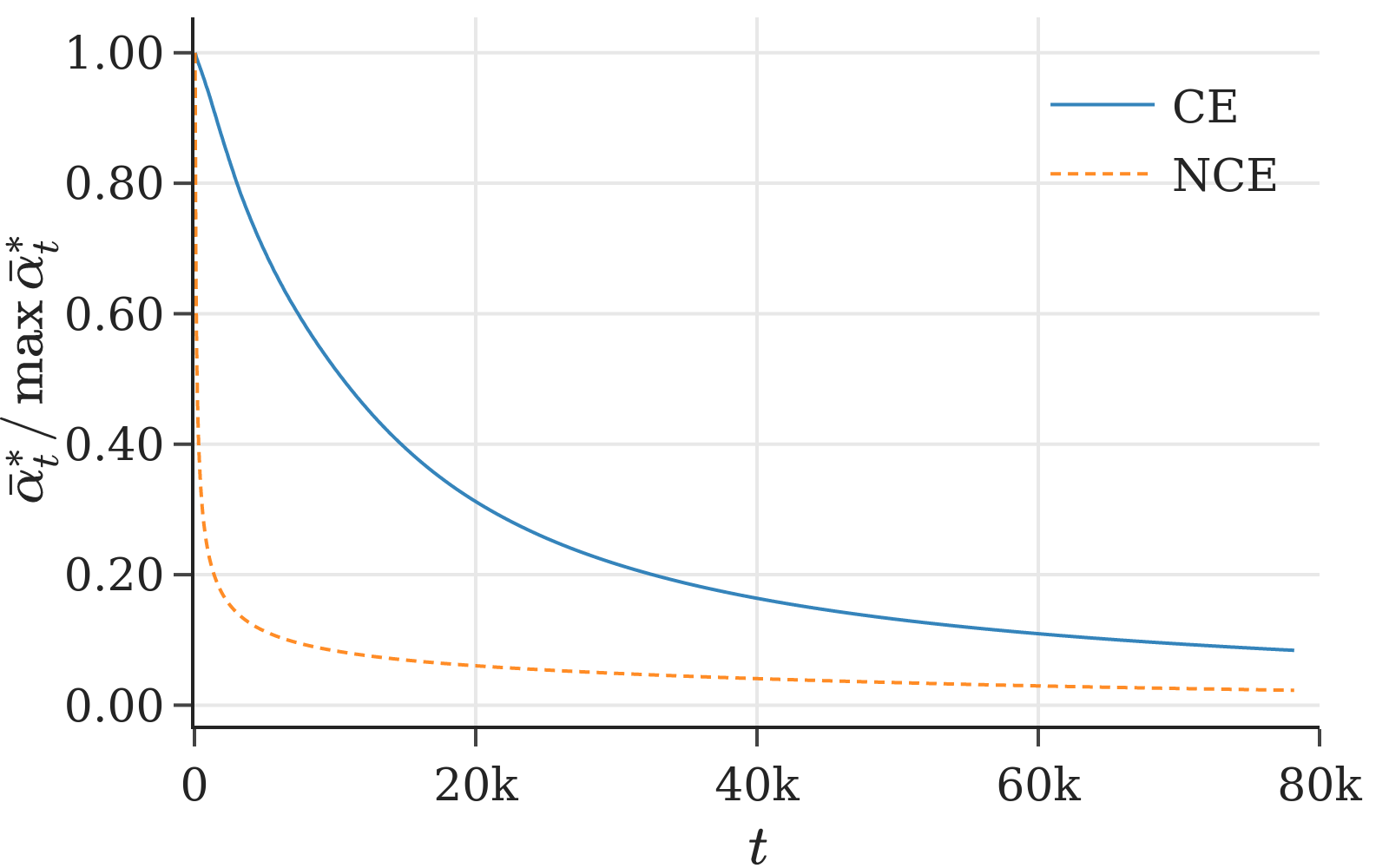}
    \caption{NCE with estimated weight upperbound.}
    \label{fig:elr:nce}
  \end{subfigure}
  \ \ \
  \begin{subfigure}[b]{0.48\textwidth}
    \centering
    \includegraphics[width=0.8\textwidth]{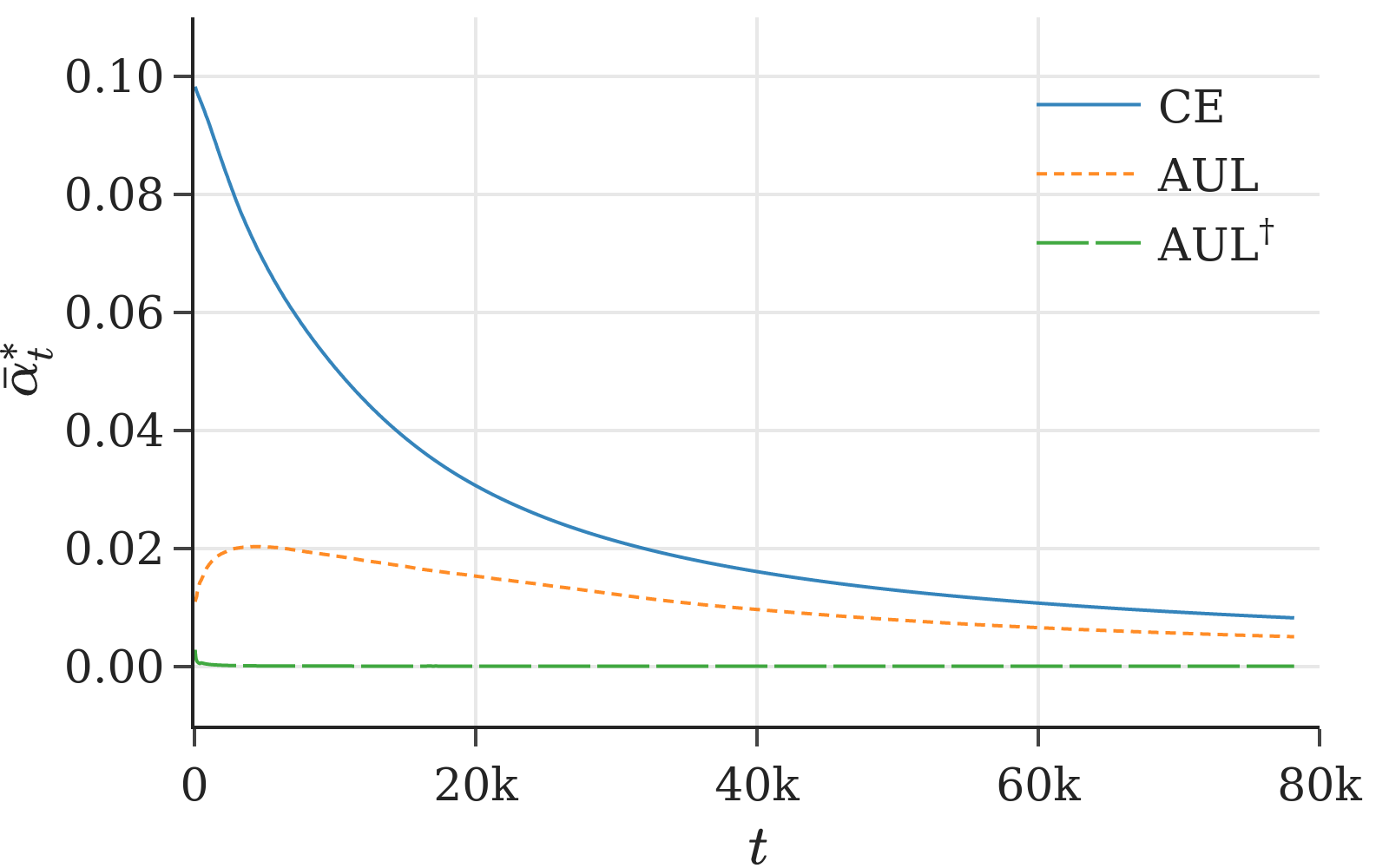}
    \caption{AUL with inferior/superior hyperparameters.}
    \label{fig:elr:gce}
  \end{subfigure}
  \caption{Different explanations for underfitting: (a) fast diminishing sample weights; (b) marginal initial sample weights. We plot the variation of $\alpha_t$ with training step $t$ on CIFAR100 without label noise for each loss function. $\alpha_t$ in (a) is normalized with its maximum to emphasize its variation during training.}
  \label{fig:elr}
\end{figure}

\textbf{Fast diminishing sample weights lead to underfitting.} In \cref{fig:elr:nce}, $\alpha_t$ of NCE peaks at initialization similar to CE. However, it decreases much faster than CE since both $\gamma$ and $\epsilon$ decrease with improved $\Delta_y$. The regularizer $R_{\mathrm{NCE}}(\bm{s})$ further reduces the entropy of softmax output and thus $\gamma$. As a result, the fast decreasing $\alpha_t$ hampers the learning of training samples and leads to underfitting.

\begin{table}[t]
  \centering
  \small
  \addtolength{\tabcolsep}{-1pt}
  \begin{tabular}{ l | r r r | r r r}
    \multicolumn{1}{c|}{}              & \multicolumn{3}{c|}{CIFAR10} & \multicolumn{3}{c}{CIFAR100}                                                 \\
    Loss                              & \multicolumn{1}{c}{Train}  & \multicolumn{1}{c}{Test}  & $\alpha_t$          & \multicolumn{1}{c}{Train} & \multicolumn{1}{c}{Test} & $\alpha_t$ \\
    \midrule
    CE      & 99.98 & 92.96 & 45.74 & 99.97 & 71.02 & 86.79 \\
    \midrule
    SCE     & 99.99 & 93.20 & 19.72 & 99.97 & 71.11 & 30.63 \\
    GCE     & 99.97 & 92.78 & 23.72 & 99.90 & 70.14 & 30.86 \\
    NCE+MAE  & 99.68 & 92.35 & /  & 92.30 & 68.28 & / \\
    \midrule
    AUL     & 99.93 & 91.80 & 4.49  & 88.19 & 59.62 & 2.11 \\
    AGCE    & 99.83 & 92.88 & 17.20 & 66.99 & 51.33 & 9.24 \\
    NCE     & 99.92 & 91.17 & /  & 31.64 & 29.67 & / \\
    \midrule
    MAE     & 98.67 & 91.93 & 4.05  & 10.32 & 9.85  & 0.28 \\
    AUL\textsuperscript{\textdagger}  & 98.80 & 92.06 & 3.25  & 9.22  & 8.67  & 0.21 \\
    AGCE\textsuperscript{\textdagger} & 91.60 & 86.30 & 3.98  & 3.82  & 3.82  & 0.12 \\
  \end{tabular}
  \caption{With clean labels, robust loss functions can underfit CIFAR100 but CIFAR10. We report the average accuracies and $\alpha_t$ (scaled by $10^4$) at the final training step with learning rate $\eta=0.1$ from 3 different runs. See \cref{app:dynamics:underfit} for hyperparameters of loss functions tuned on CIFAR100 in \cref{table:underfit_param}. Settings with inferior hyperparameters are denoted with \textdagger.}
  \addtolength{\tabcolsep}{1pt}
  \label{table:ulearn}
  \end{table}

\begin{figure}
  \centering
  \begin{subfigure}[b]{0.325\textwidth}
    \centering
    \includegraphics[width=\textwidth]{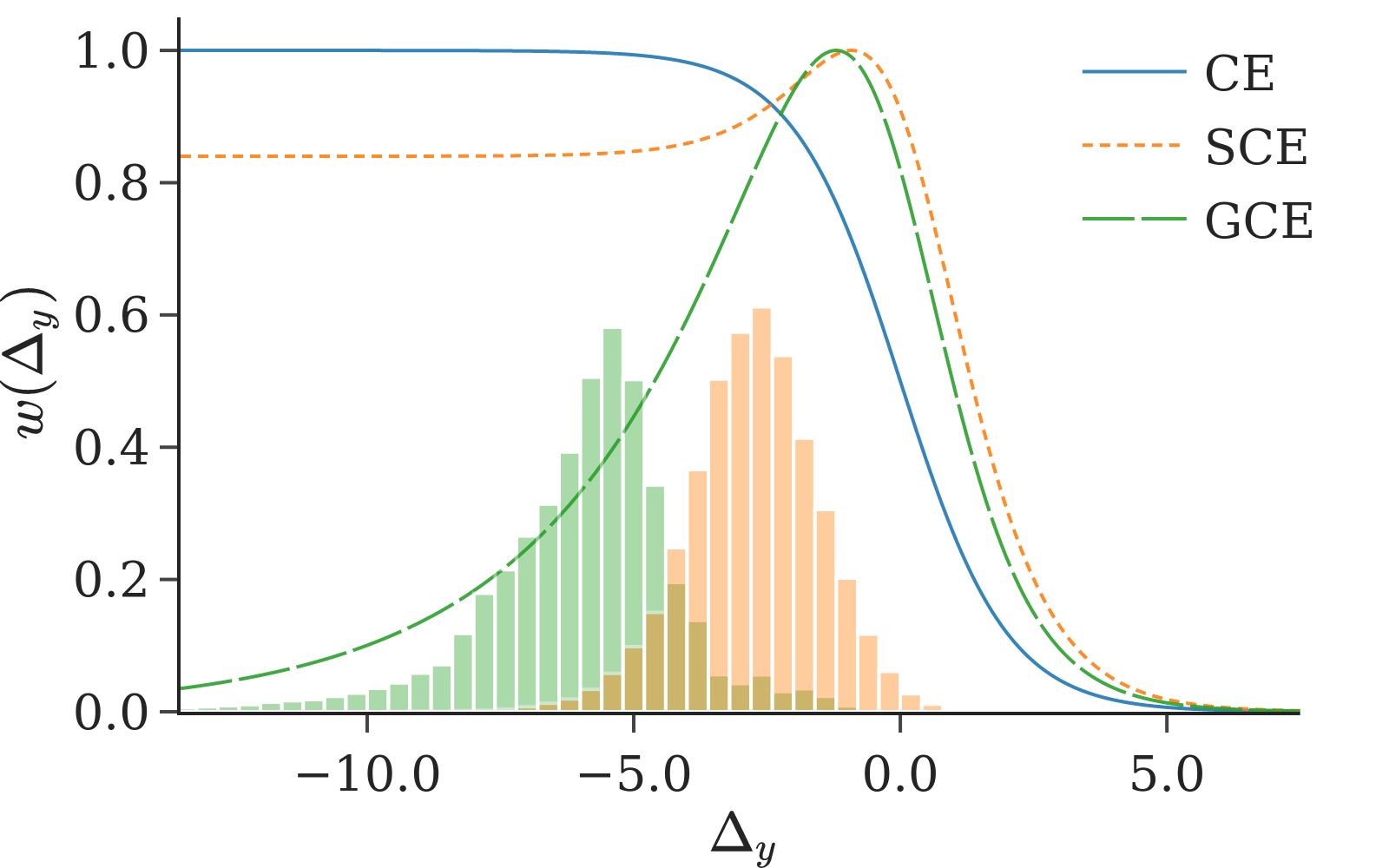}
    \caption{No underfitting}
    \label{fig:loss_weight:good}
  \end{subfigure}
  \begin{subfigure}[b]{0.325\textwidth}
    \centering
    \includegraphics[width=\textwidth]{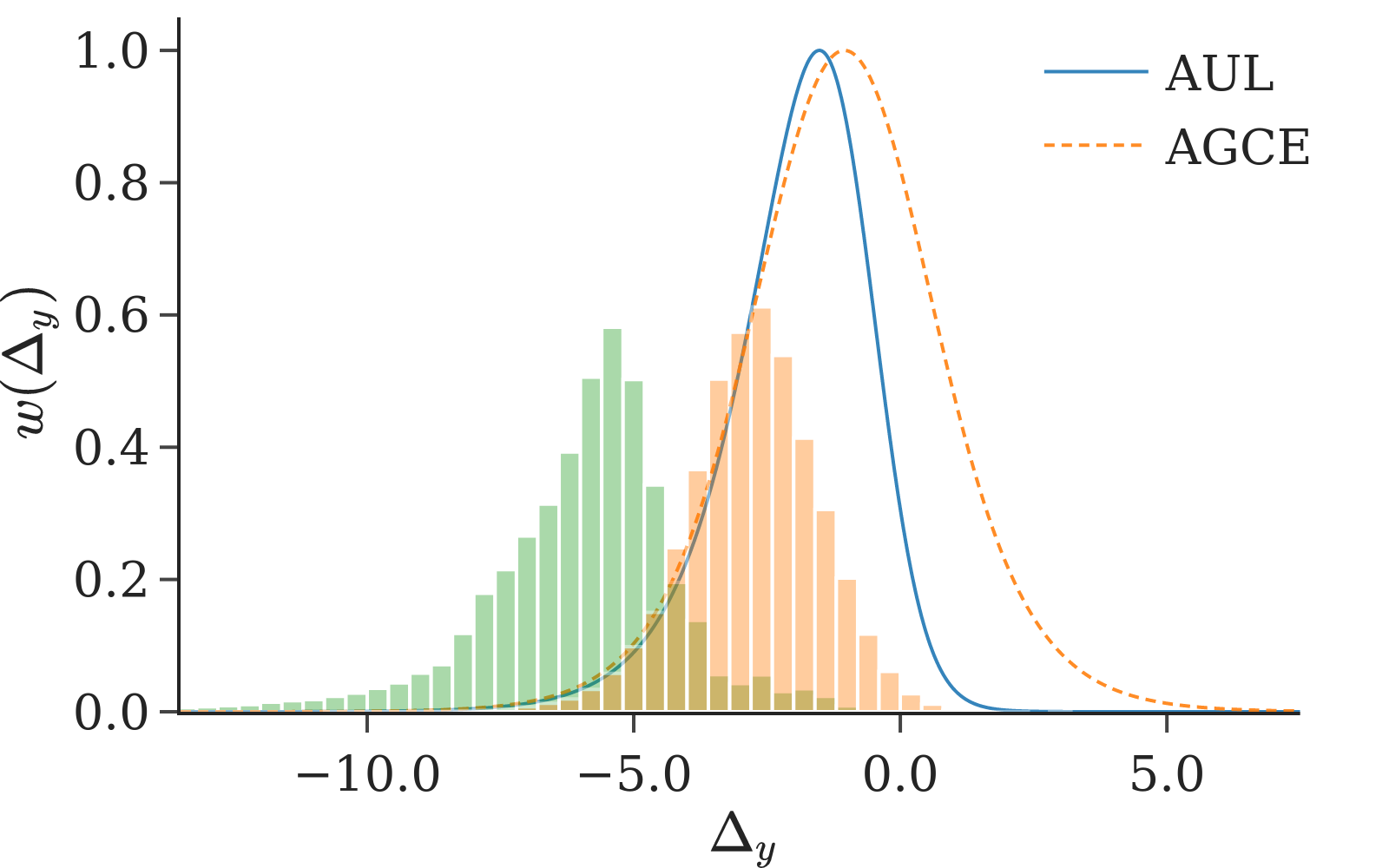}
    \caption{Mild underfitting}
    \label{fig:loss_weight:moderate}
  \end{subfigure}
  \begin{subfigure}[b]{0.325\textwidth}
    \centering
    \includegraphics[width=\textwidth]{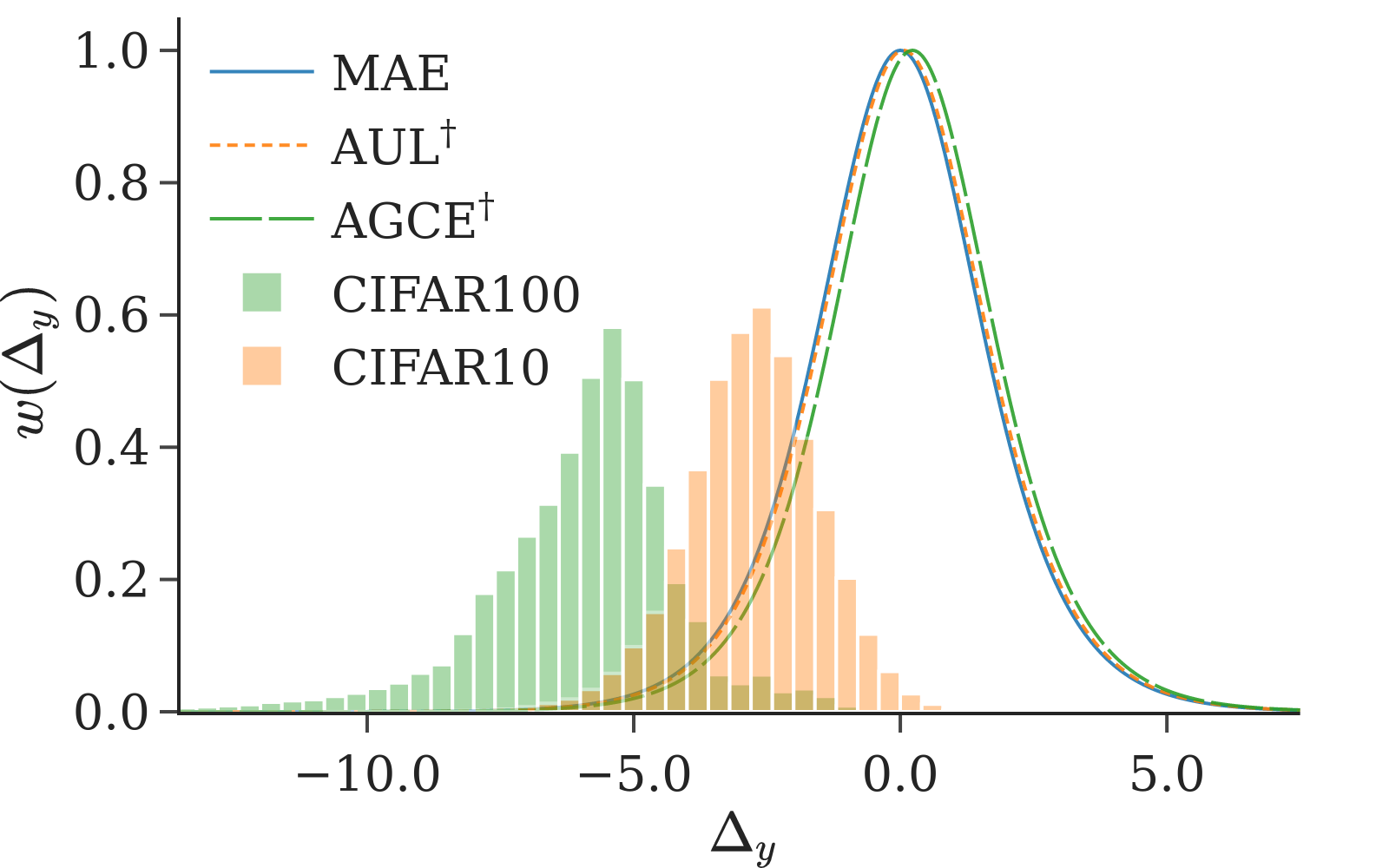}
    \caption{Severe underfitting}
    \label{fig:loss_weight:bad}
  \end{subfigure}
  \caption{Sample-weighting functions $w(\Delta_y)$ of loss functions in \cref{table:ulearn} with hyperparameters in \cref{table:underfit_param}.  We include the initial $\Delta_y$ distributions of training samples on CIFAR10 and CIFAR100 for reference, which are extracted with a randomly initialized model.}
  \label{fig:loss_weight}
  \vspace{-1mm}
\end{figure}

\textbf{Marginal initial sample weights lead to underfitting.} Unlike NCE, loss functions that severely underfit in \cref{table:ulearn} assign marginal initial weights (\cref{fig:loss_weight:bad}) to samples in CIFAR100, which leads to marginal initial $\alpha_t$. $\Delta_y$ of these samples can barely improve before the learning rate vanishes, thus leading to underfitting. In contrast, loss functions with non-trivial initial sample weights (\cref{fig:loss_weight:moderate,fig:loss_weight:good}) result in moderate or no underfitting. As further corroboration, we plot $\alpha_t$ of AUL with superior and inferior hyperparameters (AUL and AUL\textsuperscript{\textdagger} in \cref{table:ulearn}) in \cref{fig:elr:gce}. $\alpha_t$ stays marginal with AUL\textsuperscript{\textdagger}, but quickly increases to a non-negligible value before gradually decreasing with AUL.

\textbf{Loss combination can mitigate underfitting.} As $\alpha_t$ of NCE peaks at initialization but quickly diminishes while $\alpha_t$ of MAE is marginal at initialization but peaks later during training, combining NCE with MAE can mitigate the underfitting issue of each other. In \cref{table:ulearn}, combining NCE and MAE suffers less from underfitting compared to both individuals.

\textbf{Increased number of classes leads to marginal initial sample weights.} Unlike CIFAR100, loss functions in \cref{table:ulearn} perform equally well on CIFAR10. The difference has been vaguely attributed to the increased task difficulty of CIFAR100 \citep{gce,noise_survey}. Intuitively, the more classes, the more subtle differences to be distinguished. In addition, the number of classes $k$ determines the initial distribution of $\Delta_y$. Assume that class scores $s_i$ at \emph{initialization} are i.i.d. normal variables $s_i \sim \mathcal{N}(\mu, \sigma)$. In particular, $\mu = 0$ and $\sigma = 1$ for most neural networks with standard initializations \citep{xavier,kaiming} and normalization layers \citep{bn,ln}. The expected $\Delta_y$ can be approximated with
\begin{equation}
  \mathbb{E}[\Delta_y] \approx -\log(k-1) - \sigma^2/2 + \frac{e^{\sigma^2}-1}{2 (k-1)}
  \label{eq:difficult}
\end{equation}
We leave derivations and comparisons between our assumptions and real settings to \cref{app:dynamics:underfit}. A large $k$ results in small initial $\Delta_y$; with sample-weighting functions in \cref{fig:loss_weight:bad} it further leads to marginal initial sample weights, which results in underfitting on CIFAR100 as discussed above.

\begin{table}[t]
  \centering
  \small
  \begin{tabular}{ l | r | r r | r | r }
    \multicolumn{1}{c|}{}                   & \multicolumn{1}{c|}{Clean}    & \multicolumn{2}{c|}{Symmetric} & \multicolumn{1}{c|}{Asymmetric} & \multicolumn{1}{c}{Human}                                        \\
    Loss                                    & \multicolumn{1}{c|}{$\eta=0$} & \multicolumn{1}{c}{$\eta=0.4$} & \multicolumn{1}{c|}{$\eta=0.8$} & \multicolumn{1}{c|}{$\eta=0.4$} & \multicolumn{1}{c}{$\eta=0.4$} \\
    \midrule
    CE\textsuperscript{\textdaggerdbl}      & 71.33 $\pm$ 0.43              & 39.92 $\pm$ 0.10               & 7.59 $\pm$ 0.20                 & 40.17 $\pm$ 1.31                &                       \multicolumn{1}{c}{/}         \\
    \midrule
    GCE\textsuperscript{\textdaggerdbl}     & 63.09 $\pm$ 1.39              & 56.11 $\pm$ 1.35               & 17.42 $\pm$ 0.06                & 40.91 $\pm$ 0.57                &                        \multicolumn{1}{c}{/}        \\
    NCE\textsuperscript{\textdaggerdbl}     & 29.96 $\pm$ 0.73              & 19.54 $\pm$ 0.52               & 8.55 $\pm$ 0.37                 & 20.64 $\pm$ 0.40                &                       \multicolumn{1}{c}{/}         \\
    NCE+AUL\textsuperscript{\textdaggerdbl} & 68.96 $\pm$ 0.16              & 59.25 $\pm$ 0.23               & 23.03 $\pm$ 0.64                & 38.59 $\pm$ 0.48                &                        \multicolumn{1}{c}{/}        \\
    \midrule
    AGCE                                    & 49.27 $\pm$ 1.03              & 47.76 $\pm$ 1.75               & 16.03 $\pm$ 0.59                & 33.40 $\pm$ 1.57                & 30.45 $\pm$ 1.50               \\
    AGCE shift                              & 69.39 $\pm$ 0.84              & 48.21 $\pm$ 1.06               & 14.49 $\pm$ 0.17                & 40.76 $\pm$ 0.74                & 48.71 $\pm$ 0.45               \\
    AGCE scale                              & 70.57 $\pm$ 0.62              & 56.69 $\pm$ 0.33               & 14.64 $\pm$ 0.79                & 39.71 $\pm$ 0.17                & 50.85 $\pm$ 0.11               \\
    \midrule
    MAE                                     & 3.69  $\pm$ 0.59              & 1.29 $\pm$ 0.50                & 1.00 $\pm$ 0.00                 & 2.53  $\pm$ 1.34                & 2.09 $\pm$ 0.55                \\
    MAE shift                               & 68.57  $\pm$ 0.54             & 49.95  $\pm$ 0.16              & 13.10  $\pm$ 0.41               & 39.83  $\pm$ 0.18               & 47.91  $\pm$ 0.36              \\
    MAE scale                               & \textbf{70.97  $\pm$ 0.41}             & \textbf{60.57  $\pm$ 1.04}              & \textbf{24.44  $\pm$ 0.73}               & \textbf{44.48  $\pm$ 1.05}               & \textbf{54.70  $\pm$ 0.48}              \\
  \end{tabular}
  \caption{Shifting or scaling $w(\Delta_y)$ mitigates underfitting on CIFAR100 under different label noise. We report test accuracies with 3 different runs. Results from \citet{asymmetric} are included as context (denoted with \textdaggerdbl). See \cref{app:dynamics:underfit} for hyperparameter $\tau$  and results with more noise rates.}
  \label{table:correct}
\end{table}
\subsubsection{Addressing Underfitting from Marginal Initial Sample Weights}
\label{sec:dynamics:underfit:fix}
Our analysis suggests that the fixed sample-weighting function $w(\Delta_y)$ is to blame for underfitting. Assuming $\mathbb{E}[\Delta_y] < 0$ at initialization, to address underfitting from marginal initial sample weights, we can simply scale
\begin{equation*}
  w^{*}(\Delta_y) = w(\Delta_y^*) = w(\Delta_y / |\mathbb{E}[\Delta_y]| \cdot \tau)
\end{equation*}
or shift
\begin{equation*}
  w^{+}(\Delta_y) = w(\Delta_y^+) = w(\Delta_y + |\mathbb{E}[\Delta_y]| - \tau)
\end{equation*}
the sample-weighting functions, where $\tau$ is a hyperparameter. Intuitively, both approaches cancel the effect of large $k$ on the weight of $\mathbb{E}[\Delta_y]$ at initialization. A small $\tau$ thus leads to high initial sample weights regardless of $k$. See \cref{app:dynamics:underfit:fix} for visualizations of the scaled and shifted sample-weighting functions and discussions on the robustness of loss functions they induce.

We report results on CIFAR100 with different label noise in \cref{table:correct}, and results on the noisy large-scale dataset WebVision in \cref{table:correct_real}. In summary, shifting and scaling alleviate underfitting, making MAE and AGCE comparable to the previous state-of-the-art (NCE+AUL; \citealt{asymmetric}). Notably, $w^{*}(\Delta_y)$ leads to dramatic improvements for MAE under all settings. Interestingly, although $w^{*}(\Delta_y)$ and $w^{+}(\Delta_y)$ are both agnostic to the number of classes at initialization, their performances differ significantly. Intuitively, $w^{+}(\Delta_y)$ diminishes much faster than $w^{*}(\Delta_y)$ with increased $\Delta_y$, which can lead to insufficient training of clean samples and thus inferior performance.

\subsection{Understanding Noise Robustness of Loss Functions}
\label{sec:dynamics:robust}
We show that robust loss functions following \cref{eq:normalized} \emph{implicitly} assign larger weights to clean samples. The underlying reasons are explored by examining how $\Delta_y$ distributions change during training. Notably, similar sample-weighting rules are \emph{explicitly} adopted by curriculums for noise robust training \citep{method:weight3}. We leave NCE to future work as it involves an additional regularizer.

\textbf{Robust loss functions assign larger weights to clean samples.} We use the ratio between the average weights of clean ($\bar{w}_{\mathrm{clean}}$) and noisy ($\bar{w}_{\mathrm{noise}}$) samples, $\mathrm{snr}=\bar{w}_{\mathrm{clean}}/\bar{w}_{\mathrm{noise}}$, to characterize their relative contribution during training. See \cref{app:dynamics:robust} for the exact formulas. Noise robustness is characterized by differences in test accuracy compared to results with clean labels ($\mathrm{diff}$). We report $\mathrm{diff}$ and $\mathrm{snr}$ under different label noise on CIFAR10 in \cref{table:snr}. Loss functions with higher $\mathrm{snr}$ have less performance drop with label noise in general, thus being more robust.

\begin{table}[t]
  \centering
  \small
  \begin{tabular}{ l | c | c | c  }
    \multicolumn{1}{c|}{} & \multicolumn{1}{l|}{$k=50$}     & \multicolumn{1}{l|}{$k=200$}    & \multicolumn{1}{l}{$k=400$}    \\
    Settings              & \multicolumn{1}{l|}{$\tau=2.0$} & \multicolumn{1}{l|}{$\tau=1.8$} & \multicolumn{1}{l}{$\tau=1.6$} \\
    \midrule
    CE                    & 66.40                           & 70.26                           & 70.16                          \\
    \midrule
    MAE                   & \hspace{0.5em}3.68              & \hspace{0.5em}0.50              & \hspace{0.5em}0.25             \\
    MAE shift             & 60.76                           & 59.31                           & 47.32                          \\
    MAE scale             & \textbf{66.72}                           & \textbf{71.92}                           & \textbf{71.87}                          \\
  \end{tabular}
  \caption{Shifting or scaling $w(\Delta_y)$ mitigates underfitting on WebVision subsampled with different numbers of classes. $k=50$ is the standard ``mini'' setting in previous work \citep{actpass,asymmetric}. We report test accuracy with a single run due to a limited computation budget.}
  \label{table:correct_real}
\end{table}

\begin{table}[t]
  \centering
  \small
  \begin{tabular}{ c | c | c c | c c | c c | c c | c c }
         & Clean & \multicolumn{2}{c|}{Asymmetric}& \multicolumn{6}{c|}{Symmetric}    & \multicolumn{2}{c}{Human} \\
    \midrule
         &       & \multicolumn{2}{c|}{$\eta=0.2$} & \multicolumn{2}{c|}{$\eta=0.2$} & \multicolumn{2}{c|}{$\eta=0.4$} & \multicolumn{2}{c|}{$\eta=0.8$} & \multicolumn{2}{c}{$\eta=0.4$}                                                   \\
    Loss & Acc   & $\mathrm{diff}$         & $\mathrm{snr}$                  & $\mathrm{diff}$         & $\mathrm{snr}$                  & $\mathrm{diff}$        & $\mathrm{snr}$ & $\mathrm{diff}$ & $\mathrm{snr}$ & $\mathrm{diff}$ & $\mathrm{snr}$ \\
    \midrule
    CE & 90.64 & -7.06 & 0.32 & -15.47 & 0.39 & -31.95 & 0.57 & -50.87 & 0.77 & -28.51 & 0.53\\
    \midrule
    SCE & 89.87 & -5.39 & 0.51 & -3.84 & 0.99 & -10.47 & 1.27 & -27.25 & 1.51 & -15.84 & 0.86\\
    GCE & 90.44 & -7.42 & 0.36 & -6.80 & 0.96 & -23.23 & 0.89 & -45.32 & 1.04 & -21.94 & 0.72\\
    \midrule
    AUL & 89.90 & -2.51 & 0.81 & -2.07 & 3.10 & -5.87 & 2.96 & -13.90 & 2.83 & -12.08 & 1.11\\
    MAE & 89.29 & -2.21 & 1.00 & -1.92 & 3.56 & -4.36 & 3.33 & -11.53 & 3.22 & -10.35 & 1.32\\
    AGCE & 82.62 & -9.42 & 0.92 & -1.55 & 3.02 & -19.90 & 2.16 & -41.11 & 1.83 & -21.73 & 1.28\\
  \end{tabular}
  \caption{Robust loss functions assign larger weights to clean samples. We report $\mathrm{snr}$ and $\mathrm{diff}$ from the best of 5 runs on CIFAR10 under each noise setting, as inferior initialization can heavily degrade the performance. Hyperparameters listed in \cref{table:robust_param} are selected to cover more variants of sample-weighting functions (plotted in \cref{fig:snr_weight}), which are not necessarily optimal.}
  \label{table:snr}
  \vspace{-2mm}
\end{table}

\begin{figure}
  \centering
  \begin{subfigure}[b]{0.32\textwidth}
    \centering
    \includegraphics[width=\textwidth]{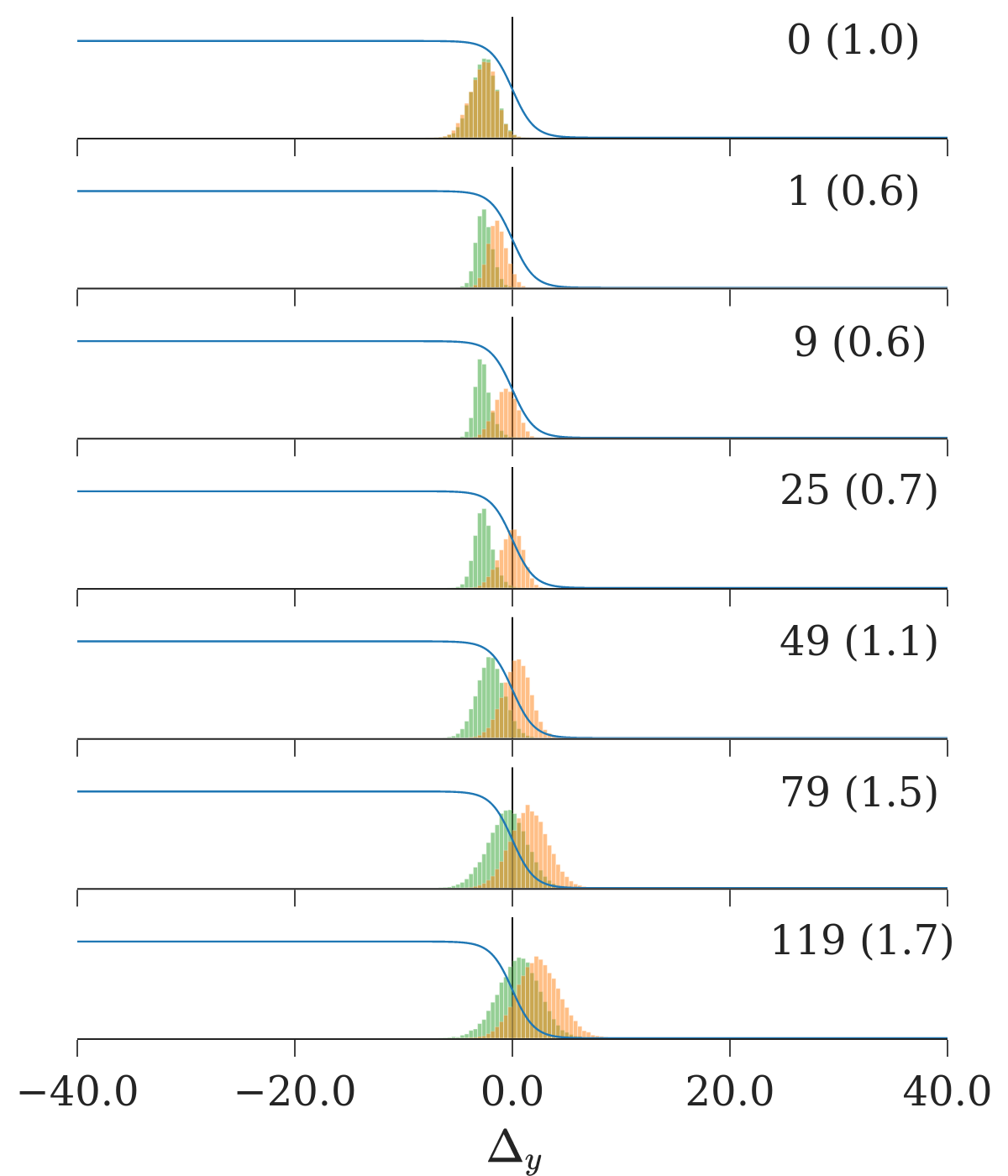}
    \caption{CE: 57.74}
    \label{fig:robust:ce}
  \end{subfigure}
  \begin{subfigure}[b]{0.32\textwidth}
    \centering
    \includegraphics[width=\textwidth]{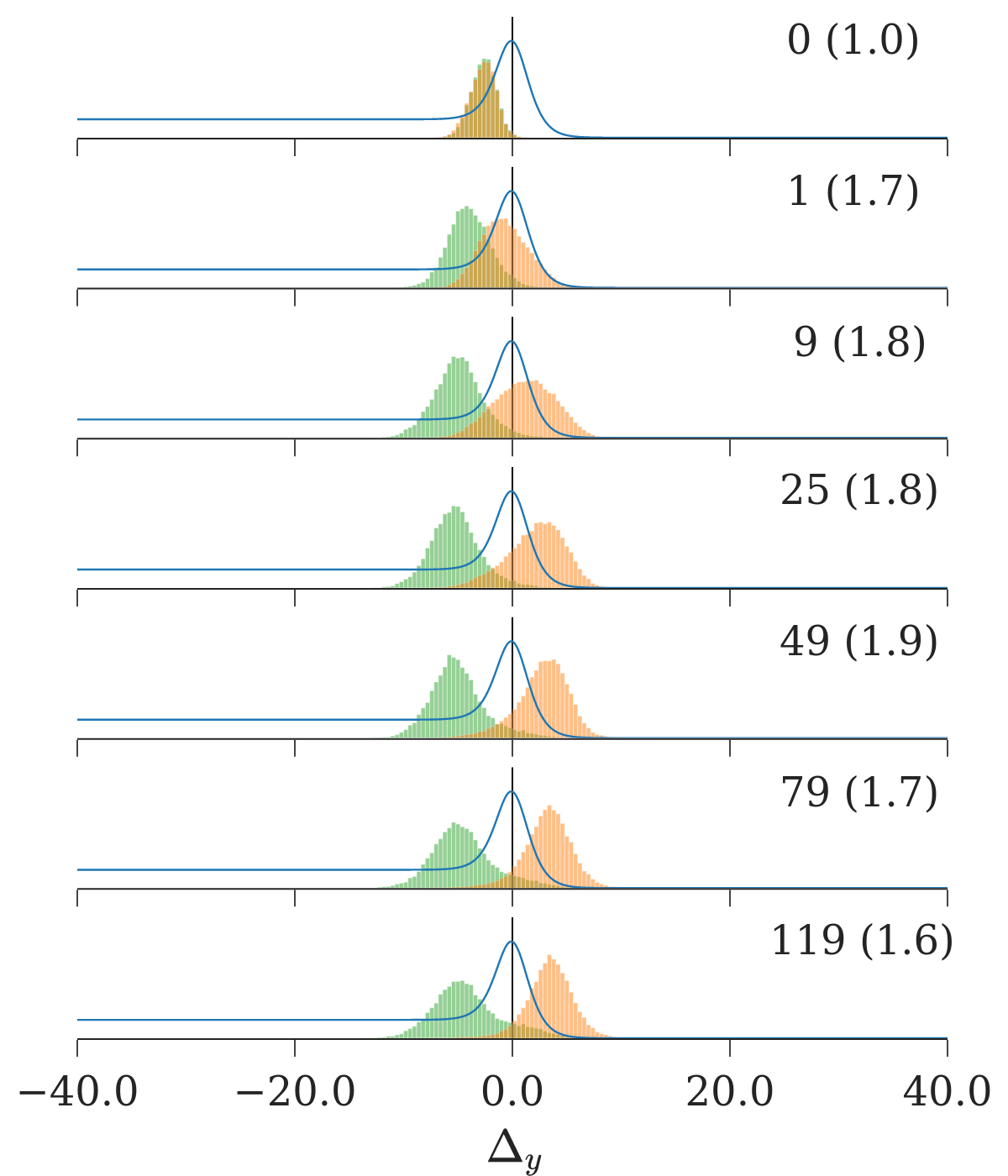}
    \caption{SCE: 79.56}
    \label{fig:robust:sce}
  \end{subfigure}
  \begin{subfigure}[b]{0.32\textwidth}
    \centering
    \includegraphics[width=\textwidth]{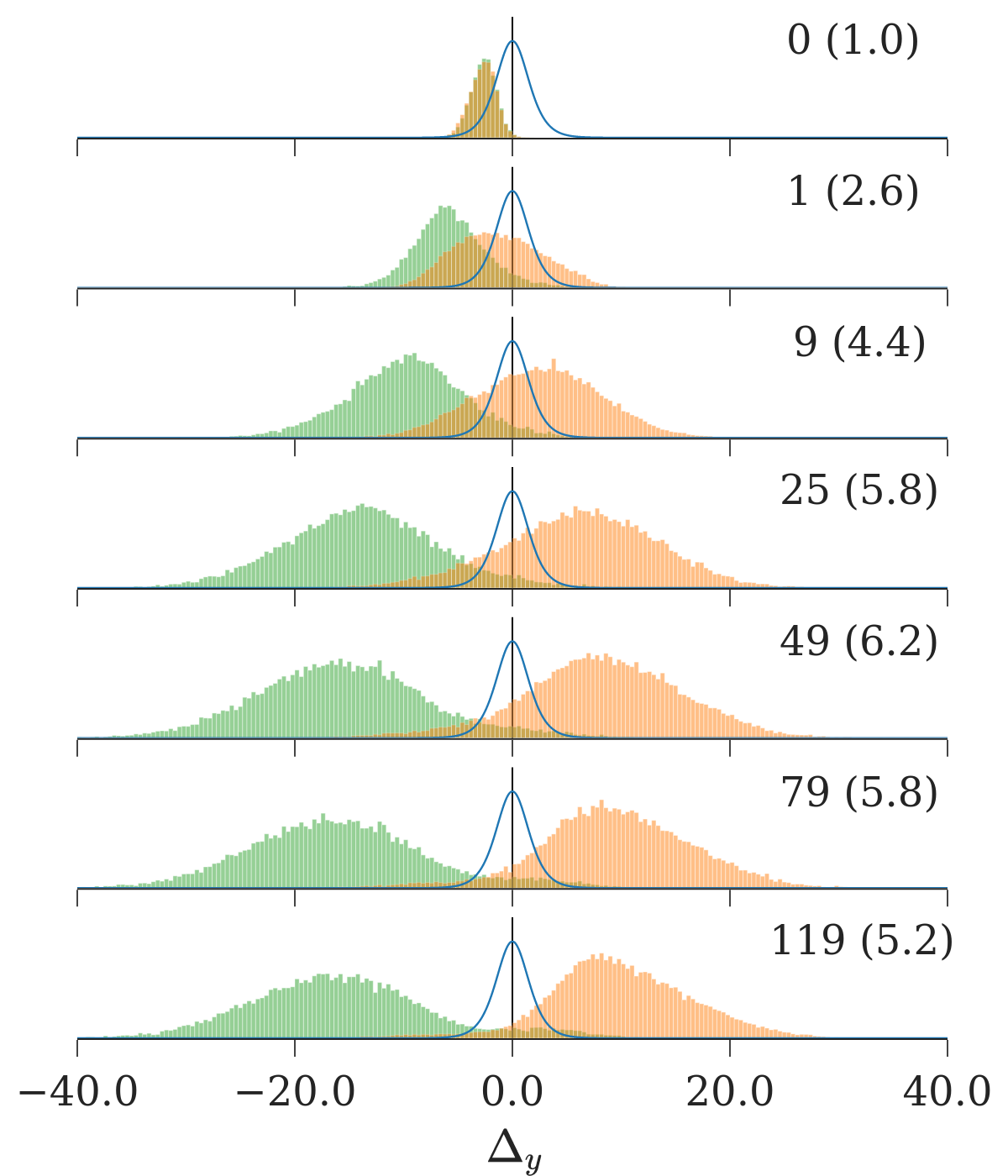}
    \caption{MAE: 83.88}
    \label{fig:robust:mae}
  \end{subfigure}
  \caption{How $\Delta_y$ distributions of noisy (green, left) and clean (orange, right) samples change on CIFAR10 during training  with symmetric label noise and $\eta=0.4$. Vertical axes denoting probability density are scaled to the peak of histograms for readability, with epoch number (axis scaling factor) denoted on the right of each subplot. We plot $w(\Delta_y)$ and report the test accuracy of each setting for reference. See \cref{app:dynamics:robust} for results with additional types of label noise and loss functions.}
  \label{fig:robust}
\end{figure}

To explain what leads to a large $\mathrm{snr}$, we plot changes of $\Delta_y$ distributions during training on CIFAR10 with symmetric label noise in \cref{fig:robust}. When trained with loss functions that are more robust against label noise (\cref{fig:robust:sce,fig:robust:mae}), $\Delta_y$ distributions of noisy and clean samples spread wider and get better separated. In addition, the consistent decrease of $\Delta_y$ for noisy samples suggests that they can be \emph{unlearned}. In contrast, training with CE (\cref{fig:robust:ce}) results in more compact and less separated $\Delta_y$ distributions. Furthermore, $\Delta_y$ of noisy samples consistently increases.

\textbf{Dynamics of SGD suppress learning of noisy samples.} As shown in \cref{fig:robust:ce}, noisy samples are learned slower than clean samples as measured by improvements of $\Delta_y$, which can be explained by more coherent gradients among clean samples \citep{generalization_mys}. Similar results have been reported \citep{generalzation,memorize} and utilized in curriculum-based robust training \citep{memorization:application1,memorization:application2}. In addition, noisy samples can be unlearned as shown in \cref{fig:robust:sce,fig:robust:mae}, which can stem from generalization with clean samples. Both dynamics suppress the learning of noisy samples but clean ones, thus leading to robustness against label noise.

\textbf{Robust $w(\Delta_y)$ synergizes with SGD dynamics for noise robustness.} In \cref{fig:loss_weight}, the bell-shaped $w(\Delta_y)$ of robust loss functions only assigns large weights to samples with moderate $\Delta_y$. Since $\Delta_y$ distributions initially concentrate at the monotonically increasing interval of $w(\Delta_y)$, (1) samples with faster improving $\Delta_y$, due to either larger initial weights or faster learning as clean samples, are weighted more during early training and learned faster. The magnified learning pace difference explains the widely spread distributions in \cref{fig:robust:sce,fig:robust:mae}. In addition, (2) the unlearned samples with small $\Delta_y$ receive diminishing weights from $w(\Delta_y)$, which hampers their pace of learning. Noisy samples in \cref{fig:robust:sce,fig:robust:mae} are consistently unlearned and ignored with marginal sample weights, leading to a consistent decrease in $\Delta_y$.
In addition to the SGD dynamics, (1) and (2) further suppress the learning of noisy samples and enhance that of clean samples, thus leading to increased robustness against label noise. In contrast, the monotonically decreasing $w_{\mathrm{CE}}(\Delta_y)$ emphasizes samples with smaller $\Delta_y$, essentially acting against the SGD dynamics for noise robustness. Thus training with CE results in increased vulnerability to label noise as shown in \cref{table:snr}.

\subsubsection{Training Schedules Affect Noise Robustness}
\label{sec:dynamics:robust:schedule}
Although the learning pace of noisy samples gets initially suppressed, the expected gradient will eventually be dominated by noisy samples, since well-learned clean samples receive marginal sample weights thanks to the monotonically decreasing interval of $w(\Delta_y)$. Models with extended training\footnote{Enough training steps without early stopping or diminishing learning rates for a small training loss.} thus risk overfitting noisy samples during the late training stage. Adjusting the training schedules to enable or avoid such overfitting can therefore affect the noise robustness of models. Based on this intuition, we present two surprising examples that deviate from existing theoretical results:

\begin{figure}
  \centering
  \begin{subfigure}[b]{0.48\textwidth}
    \centering
    \includegraphics[width=\textwidth]{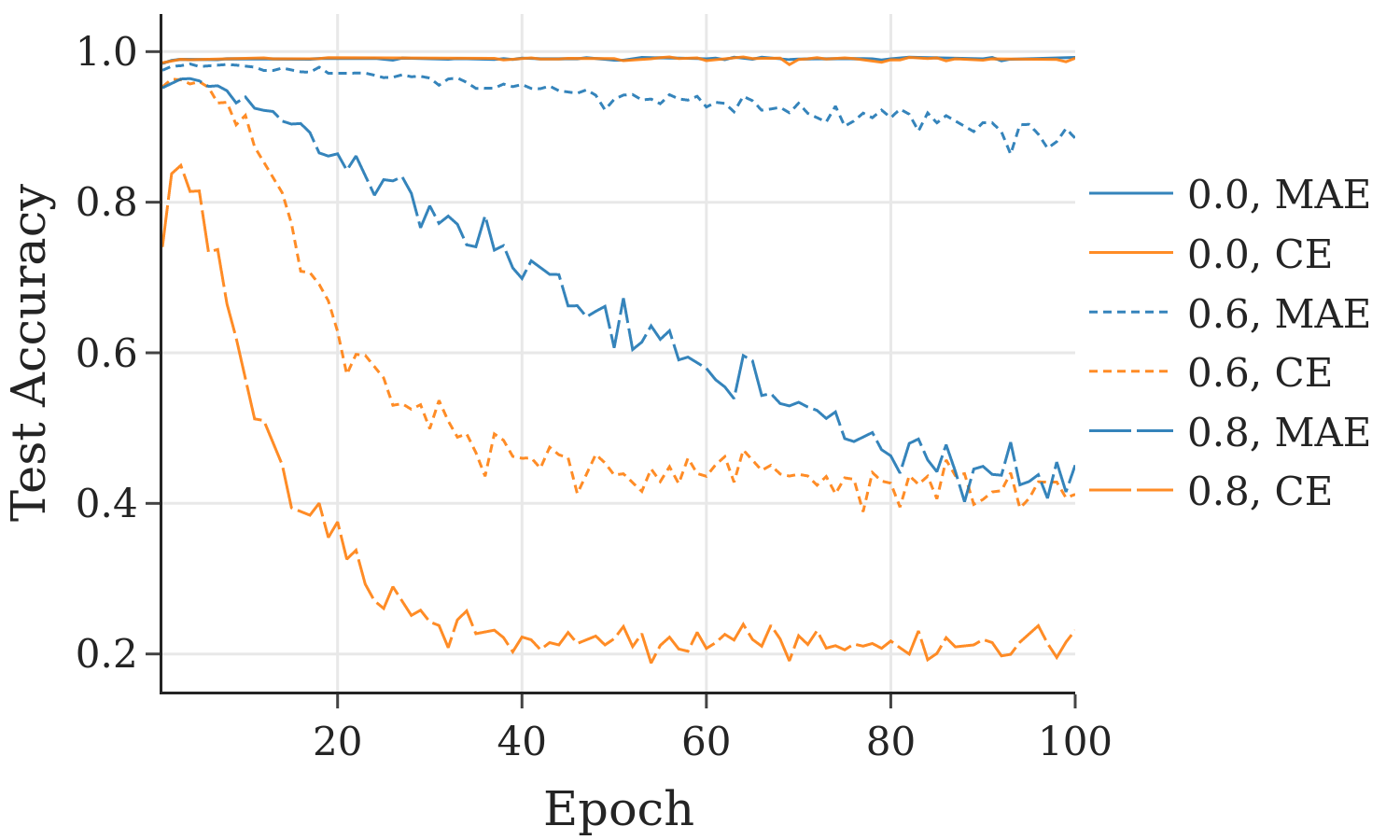}
    \caption{$\alpha=0.01$ with different ($\eta$, loss).}
    \label{fig:robust:overfitmae}
  \end{subfigure}
  \hfill
  \begin{subfigure}[b]{0.48\textwidth}
    \centering
    \includegraphics[width=\textwidth]{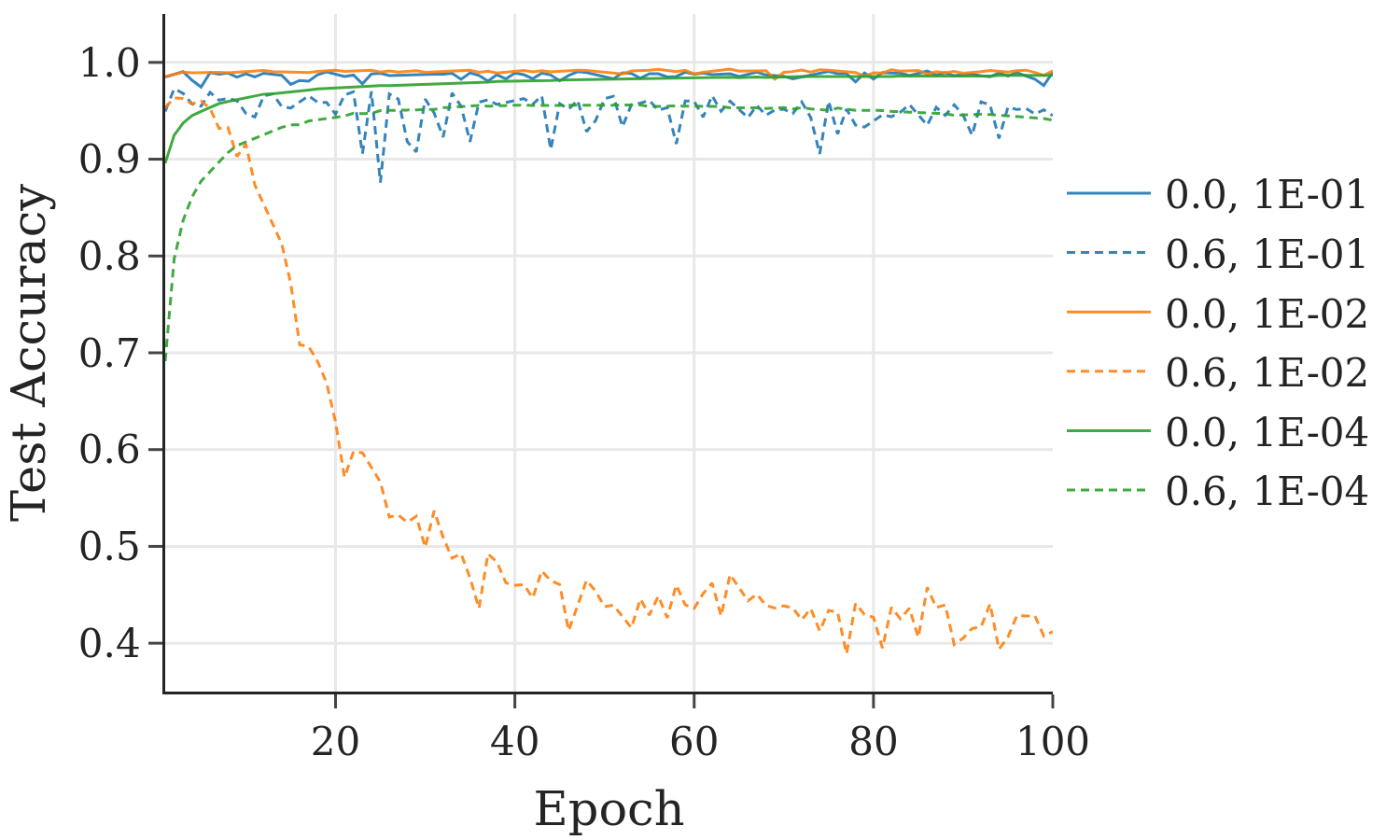}
    \caption{CE with different ($\eta$, $\alpha$).}
    \label{fig:robust:underfitce}
  \end{subfigure}
  \caption{Learning curves with fixed learning rate and extended training epochs on MNIST under symmetric label noise, where $\alpha$ is the learning rate and $\eta$ the noise rate.}
  \vspace{-1.2mm}
\end{figure}

\textbf{Extended training can make robust loss functions vulnerable to label noise.}  The learning curves of CE and MAE with \emph{constant} learning rates on MNIST are shown in \cref{fig:robust:overfitmae}. Despite the theoretically guaranteed noise robustness \citep{mae}, similar to CE, with extended training, MAE eventually overfits noisy samples, resulting in vulnerability to label noise.

\textbf{CE can become robust by adjusting the learning rate schedule.} To avoid overfitting noisy samples, we can avoid learning when noisy samples dominate the expected gradient. It can be achieved with either early stopping \citep{earlystop}, or a constrained learning pace that prevents sufficient learning of clean samples, which avoids diminishing weights for them. We show the learning curve of CE using fixed learning rates under symmetric label noise on MNIST in \cref{fig:robust:underfitce}. By simply increasing or decreasing the learning rate, which strengthens the implicit regularization of SGD \citep{sgd_reg} or directly slows down the learning pace, CE can become robust against label noise.

\vspace{-1.3mm}
\section{Conclusion and Discussions}
\vspace{-1mm}
We extend the understanding of robust loss functions by considering the training dynamics to reach the risk minimizers. By rewriting numerous loss functions into the same class-score margin and varied sample-weighting functions, we explicitly connect the design of loss functions to the design of sample-weighting curriculums and unify a broad array of loss functions. Based on the curriculum view, we gain more insights into how robust loss functions work and propose effective fixes to address the underfitting issue of robust loss functions.

\bibliography{iclr2023_conference}
\bibliographystyle{iclr2023_conference}

\appendix

\section{Extended Review of Loss Functions}
\label{app:losses}
\begin{figure}
  \centering
  \begin{subfigure}[b]{0.32\textwidth}
    \centering
    \includegraphics[width=\textwidth]{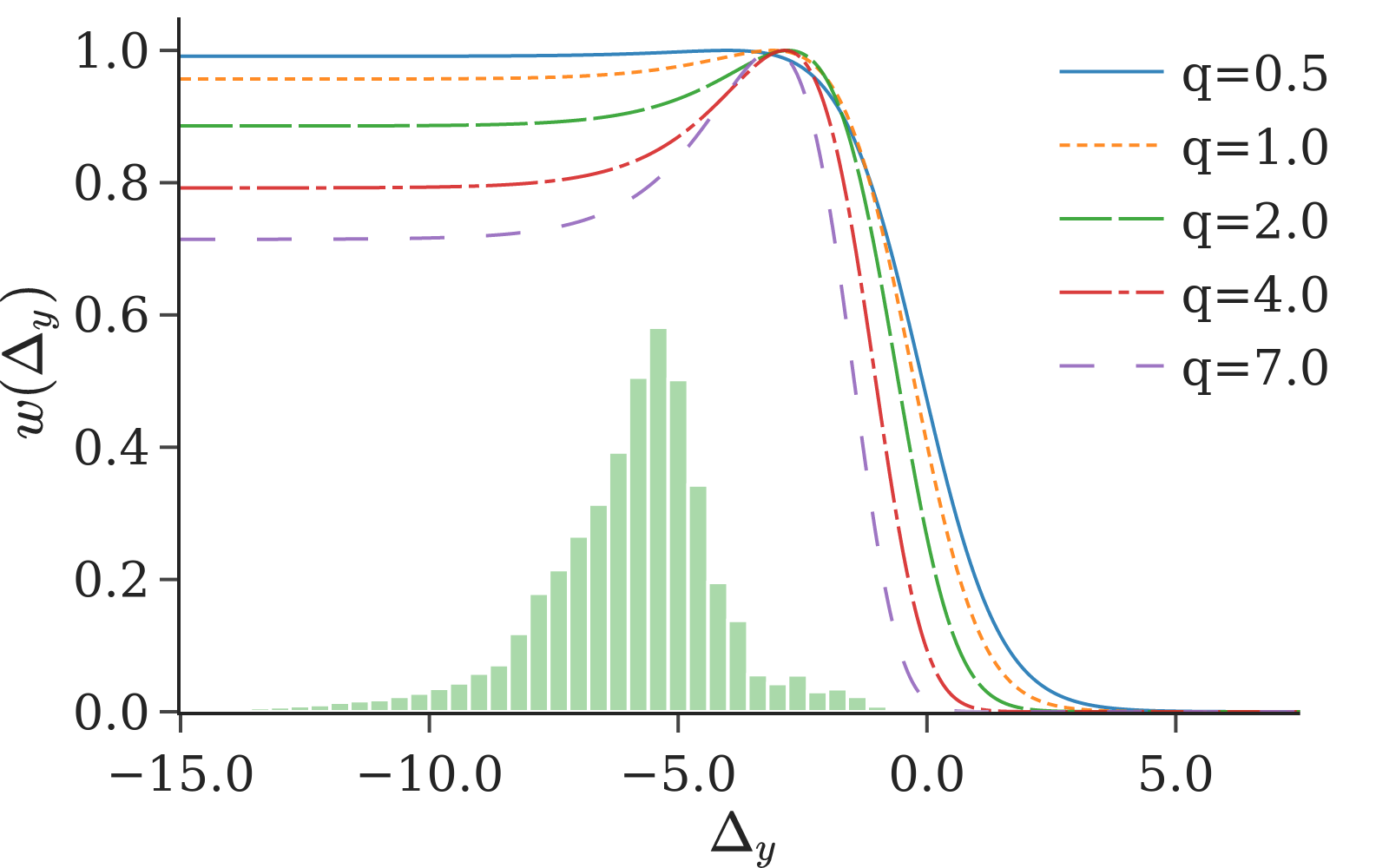}
    \caption{FL}
  \end{subfigure}
  \hfill
  \begin{subfigure}[b]{0.32\textwidth}
    \centering
    \includegraphics[width=\textwidth]{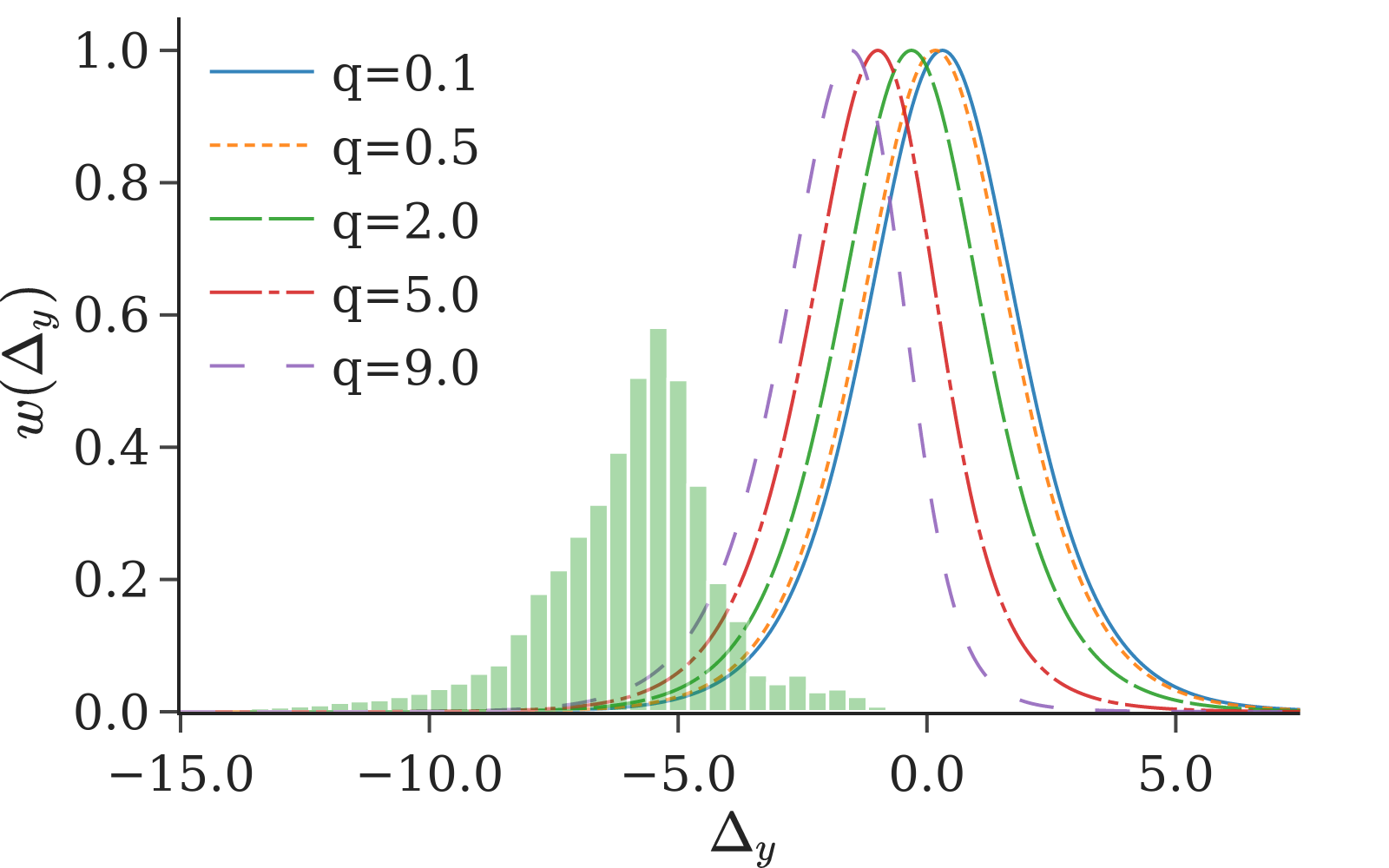}
    \caption{AUL, $a=2.0$}
  \end{subfigure}
  \hfill
  \begin{subfigure}[b]{0.32\textwidth}
    \centering
    \includegraphics[width=\textwidth]{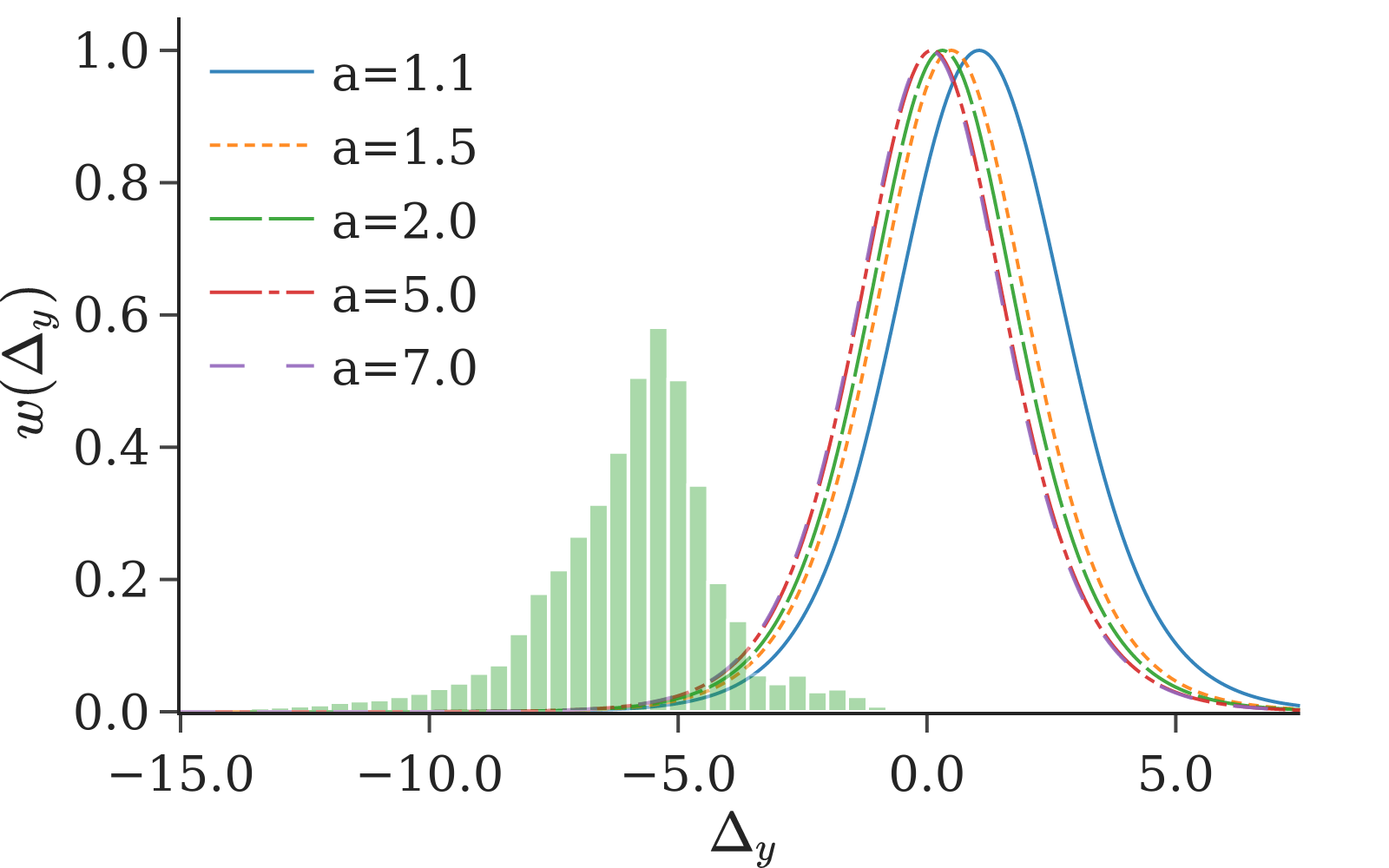}
    \caption{AUL, $q=0.1$}
  \end{subfigure}

  \begin{subfigure}[b]{0.32\textwidth}
    \centering
    \includegraphics[width=\textwidth]{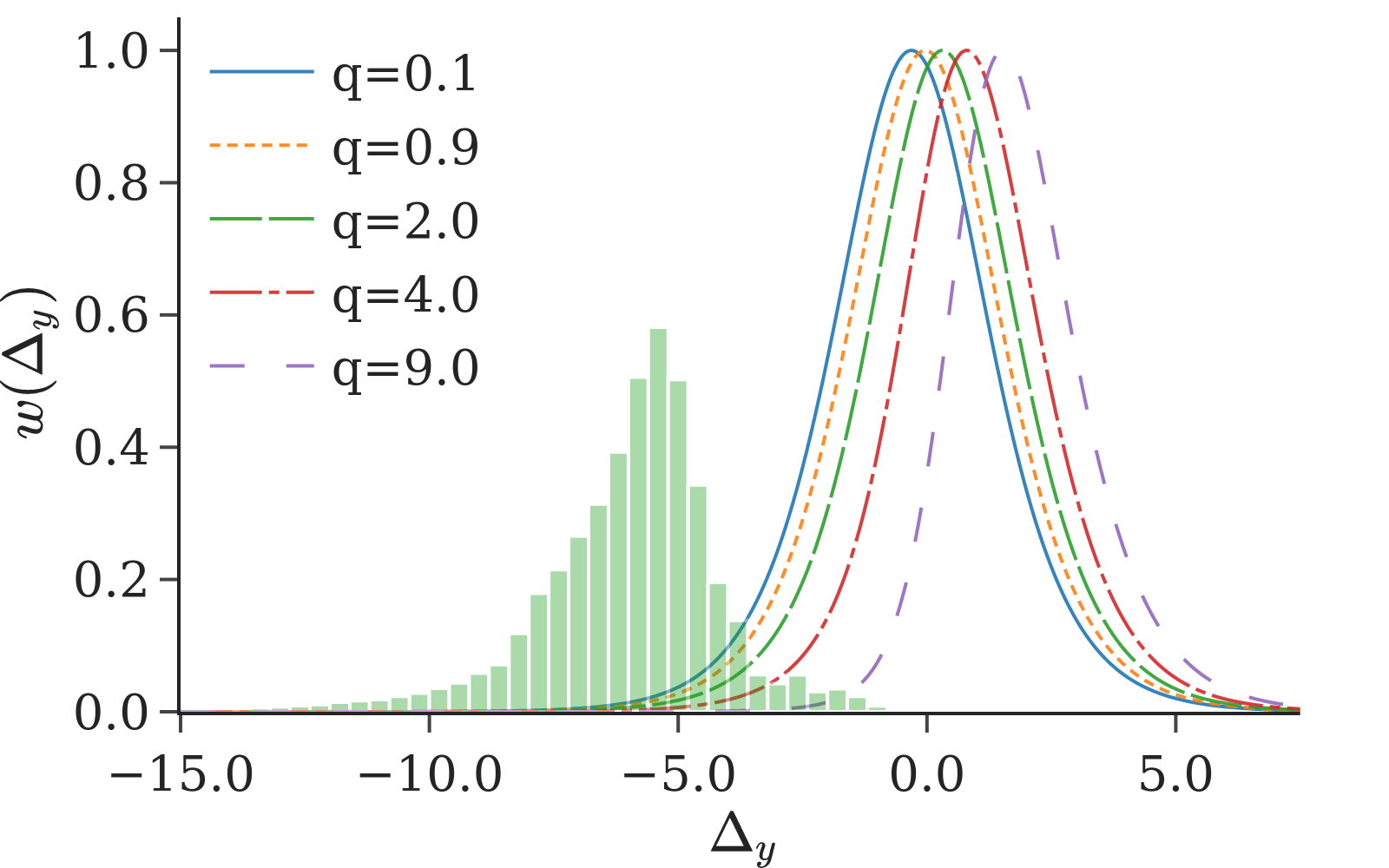}
    \caption{AGCE, $a=1.0$}
  \end{subfigure}
  \hfill
  \begin{subfigure}[b]{0.32\textwidth}
    \centering
    \includegraphics[width=\textwidth]{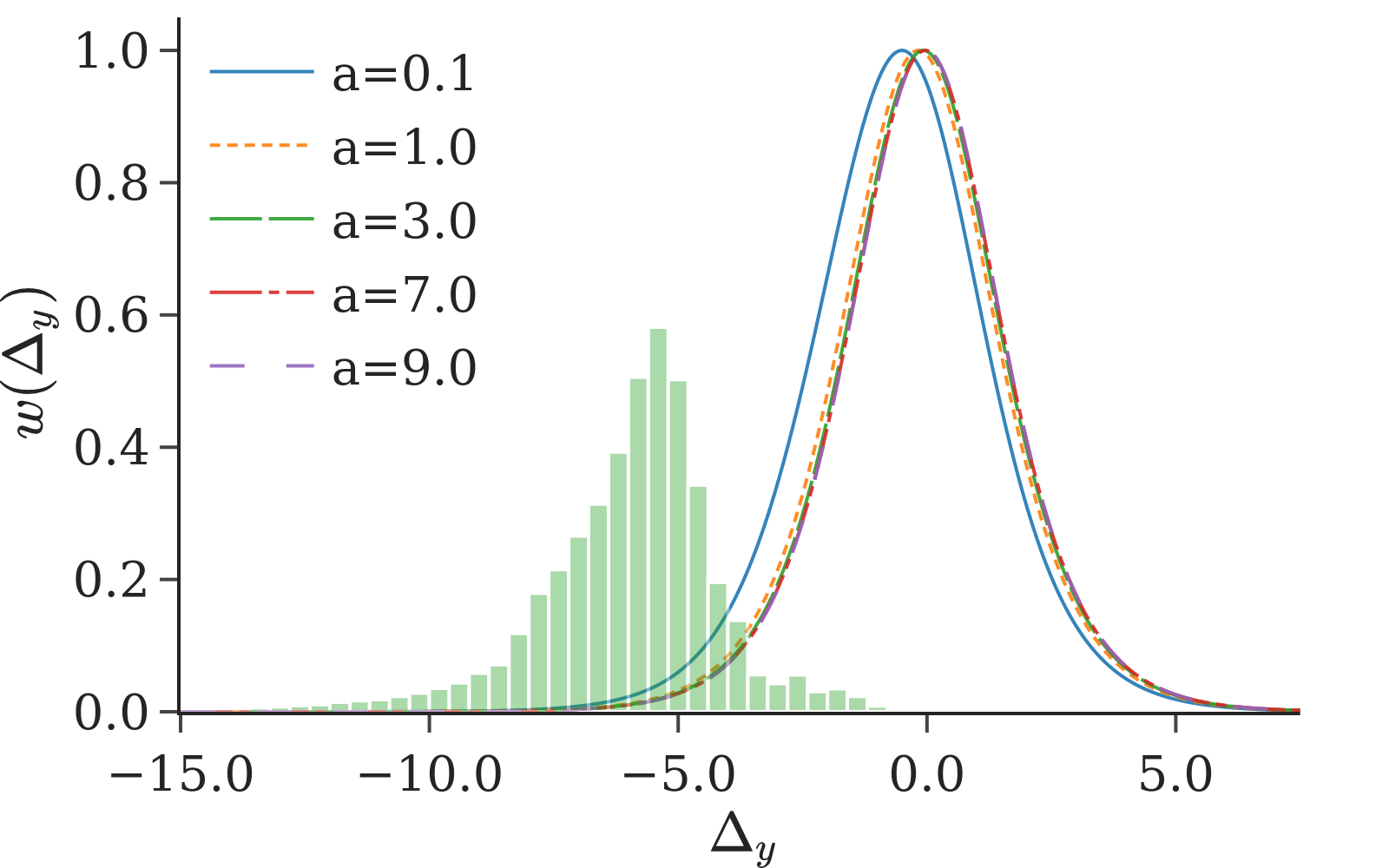}
    \caption{AGCE, $q=0.5$}
  \end{subfigure}
  \hfill
  \begin{subfigure}[b]{0.32\textwidth}
    \centering
    \includegraphics[width=\textwidth]{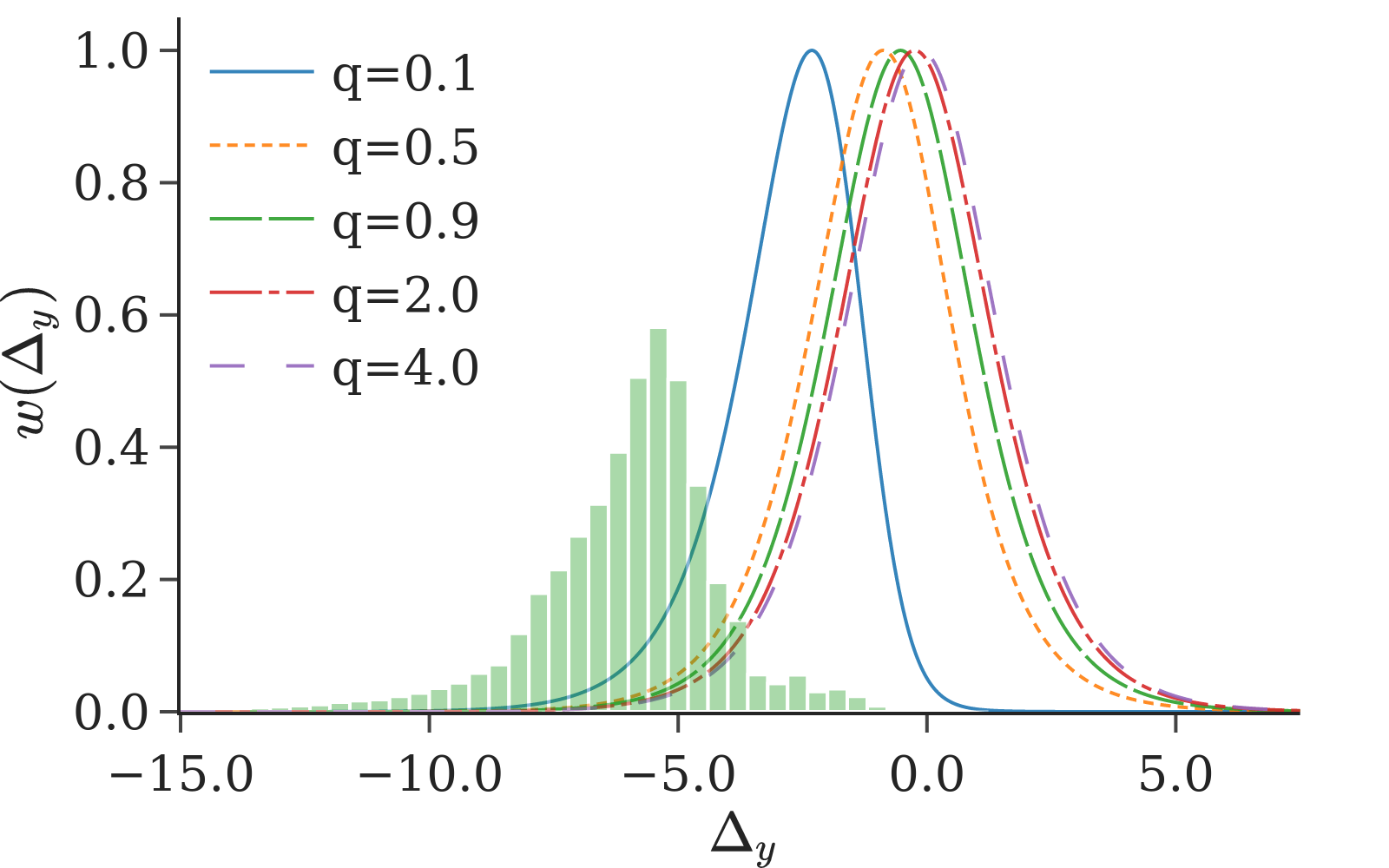}
    \caption{AEL}
  \end{subfigure}

  \begin{subfigure}[b]{0.32\textwidth}
    \centering
    \includegraphics[width=\textwidth]{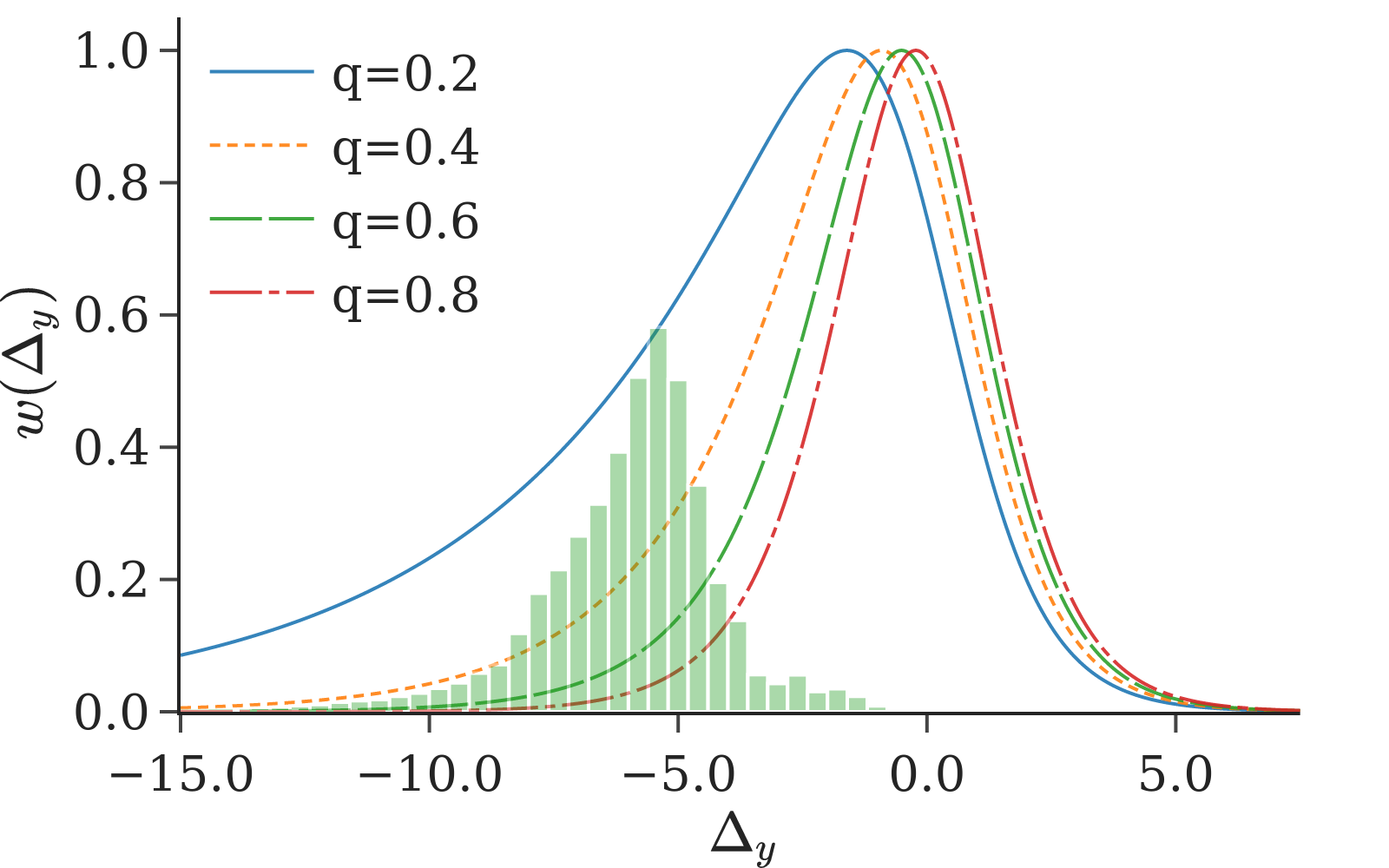}
    \caption{GCE}
  \end{subfigure}
  \hfill
  \begin{subfigure}[b]{0.32\textwidth}
    \centering
    \includegraphics[width=\textwidth]{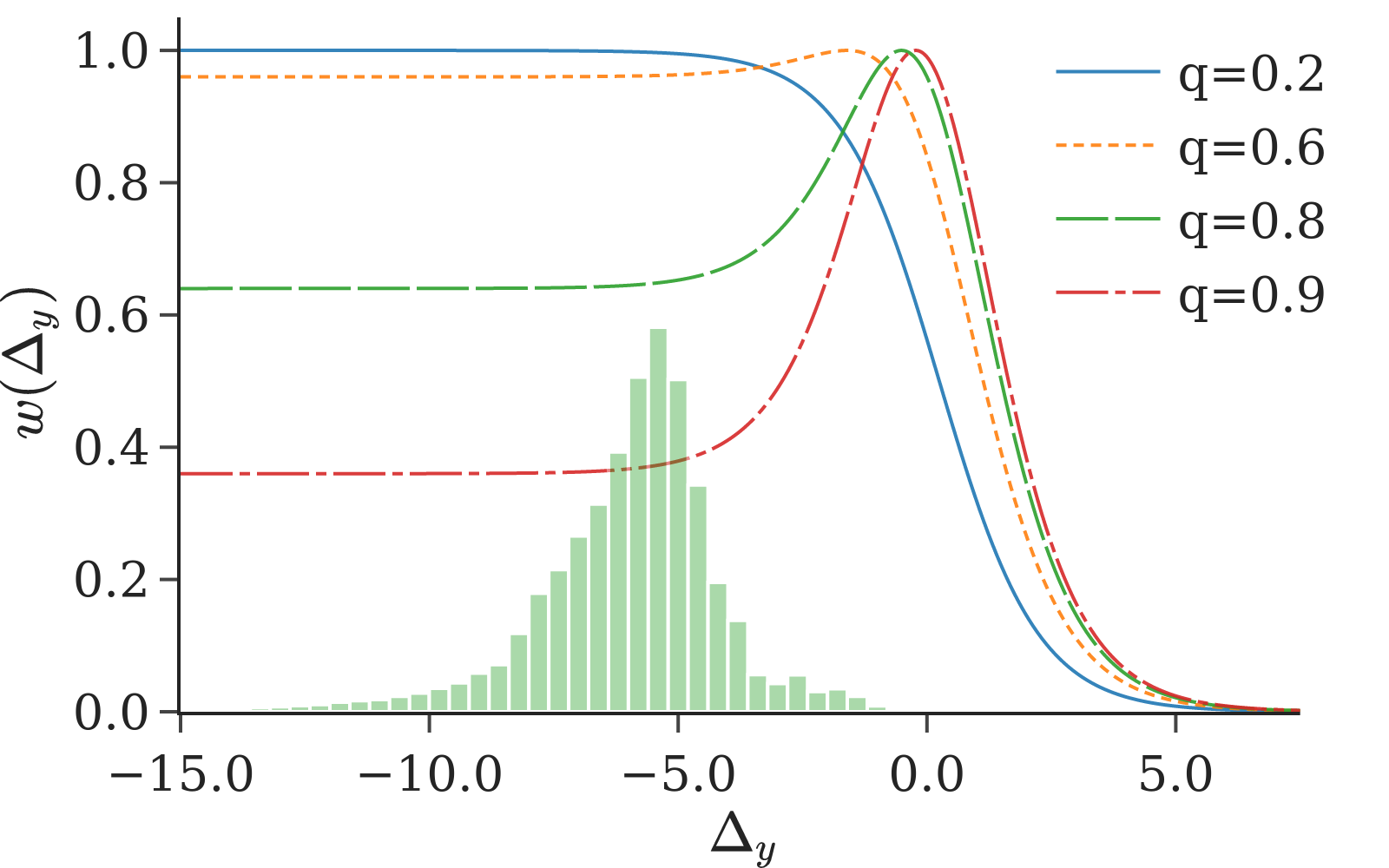}
    \caption{SCE}
  \end{subfigure}
  \hfill
  \begin{subfigure}[b]{0.32\textwidth}
    \centering
    \includegraphics[width=\textwidth]{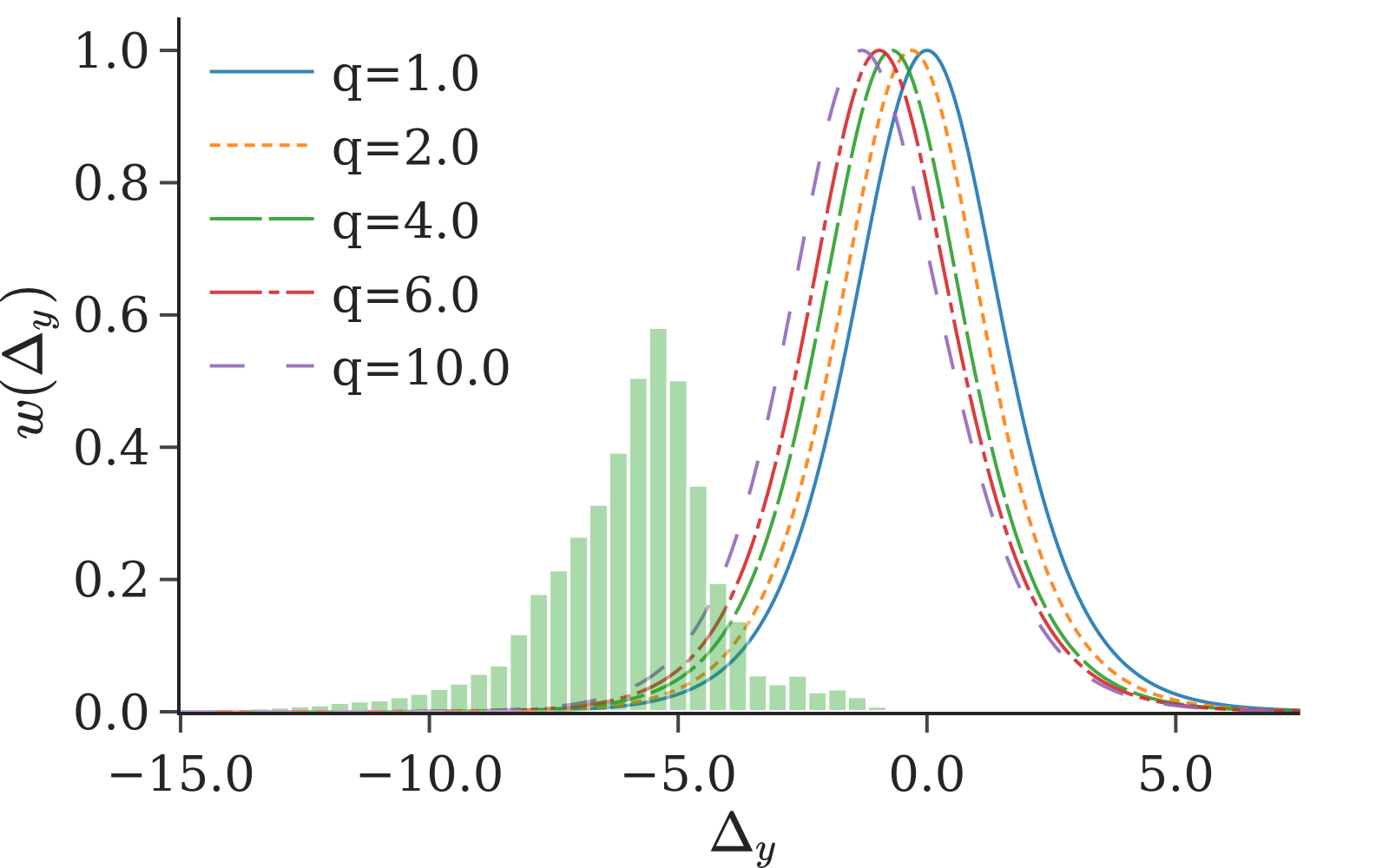}
    \caption{TCE}
  \end{subfigure}

  \caption{How hyperparameters affect the sample-weighting functions in \cref{table:weight_extend}. The initial $\Delta_y$ distributions of CIFAR100 extracted with a randomly initialized model are included as reference.}
  \label{fig:loss_params}
\end{figure}

Due to limited space, we only briefly describe typical robust loss functions in \cref{sec:background:loss}. As a general reference, here we provide a comprehensive review of loss functions related to the standard form \cref{eq:normalized}. Similar to \cref{sec:background:loss}, we ignore the differences in constant scaling factors and additive bias. Loss functions and their sample-weighting functions are summarized in \cref{table:weight_extend}. We examine how hyperparameters affect their sample-weighting functions in \cref{fig:loss_params}.

\subsection{Loss Functions without Robustness Guarantees}
\textbf{Cross Entropy} (CE)
\begin{equation*}
  L_{\mathrm{CE}}(\bm{s}, y) = -\log p_y
\end{equation*}
is the standard loss function for classification.

\textbf{Focal Loss} (FL; \citealt{focal})
\begin{equation*}
  L_{\mathrm{FL}}(\bm{s}, y) = -(1-p_y)^{q} \log p_y
\end{equation*}
aims to address label imbalance when training object detection models. Both CE and FL are neither symmetric \citep{actpass} nor asymmetric \citep{asymmetric}.

\subsection{Symmetric Loss Functions}
\label{app:losses:symmetric}
\textbf{Mean Absolute Error} (MAE; \citealt{mae})
\begin{equation*}
  L_{\mathrm{MAE}}(\bm{s}, y) = \sum_{i=1}^k |\mathbb{I}(i=y) - p_i| = 2 - 2p_y \propto 1-p_y
\end{equation*}
is a classic symmetric loss function, where $\mathbb{I}(i=y)$ is the indicator function.

\textbf{Reverse Cross Entropy} (RCE; \citealt{sce})
\begin{equation*}
  L_{\mathrm{RCE}}(\bm{s}, y) = \sum_{i=1}^k p_i \log \mathbf{1}(i=y) = \sum_{i \neq y} p_i A = (1 - p_y) A  \propto 1-p_y = L_{\mathrm{MAE}}(\bm{s}, y)
\end{equation*}
is equivalent to MAE in implementation, where $\log 0$ is truncated to a negative constant $A$ to avoid numerical overflow.

\citet{actpass} argued that any generic loss functions with $L(\bm{s}, i) > 0, \forall i \in \{1..k\}$ can become symmetric by simply normalizing them. As an example,

\textbf{Normalized Cross Entropy} (NCE; \citealt{actpass})
\begin{equation*}
  L_{\mathrm{NCE}}(\bm{s}, y) = \frac{L_{\mathrm{CE}}(\bm{s}, y)}{\sum_{i=1}^{k} L_{\mathrm{CE}}(\bm{s}, i)} = \frac{-\log p_y}{\sum_{i=1}^{k} -\log p_i}
  \label{eq:nce}
\end{equation*}
is a symmetric loss function. However, NCE does not follow the standard form of \cref{eq:normalized} as it additionally depends on $p_i$, $i\neq y$. It involves an additional regularizer, thus being more relevant to discussions in \cref{app:losses:regularizer}.

\begin{table}[t]
  \centering
  \small
  \renewcommand{\arraystretch}{1.8}
  \begin{tabular}{ c | c | c | c  }
    Name    & Function                          & Sample Weight $w$                     & Constraints      \\
    \midrule
    CE      & $-\log p_y$                       & $1-p_y$                               &                  \\
    FL      & $-(1-p_y)^{q} \log p_y$           & $(1-p_y)^{q}(1 - p_y - qp_y\log p_y)$ & $q>0$            \\
    \midrule
    MAE/RCE & $1- p_y$                          & $p_y(1-p_y)$                          &                  \\
    \midrule

    AUL     & $\frac{(a+1) - (a + p_y)^q}{q}$   & $p_y(1-p_y)(a - p_y)^{q-1}$           & $a > 1$, $q > 0$ \\
    AGCE    & $\frac{(a - p_y)^q - (a-1)^q}{q}$ & $p_y(a+p_y)^{q-1}(1-p_y) $            & $a > 0$, $q > 0$ \\
    AEL     & $e^{-p_y/q}$                      & $\frac{1}{q}p_y (1-p_y) e^{-p_y/q}$   & $q>0$            \\
    \midrule
    GCE     & $(1 - p_y^q)/q$                   & $p_y^q(1-p_y)$                        & $ 0 < q \leq 1$  \\
    SCE     & $ -(1-q) \log p_y + q (1-p_y)$    & $ (1-q + q \cdot p_y )(1-p_y)$        & $0 < q < 1$      \\
    TCE     & $\sum_{i=1}^{q}(1 - p_y)^i/i$     & $ p_y \sum_{i=1}^{q} (1-p_y)^{i}$     & $q \geq 1$       \\
  \end{tabular}
  \caption{Expressions, constraints of hyperparameters and sample-weighting functions of loss functions reviewed in \cref{app:losses} that follow the standard form \cref{eq:normalized}.}
  \renewcommand{\arraystretch}{1}
  \label{table:weight_extend}
\end{table}

\subsection{Asymmetric Loss Functions}
\citet{asymmetric} derived the asymmetric condition for noise robustness and propose numerous asymmetric loss functions:

\textbf{Asymmetric Generalized Cross Entropy} (AGCE)
\begin{equation*}
  L_{\mathrm{AGCE}}(\bm{s}, y) = \frac{(a+1) - (a + p_y)^q}{q}
\end{equation*}
where $a > 0$ and $q > 0$. It is asymmetric when $\mathbb{I}(q \leq 1) (\frac{a+1}{a})^{1-q} + \mathbb{I} (q > 1) \leq 1/ \tilde{r}$.

\textbf{Asymmetric Unhinged Loss} (AUL)
\begin{equation*}
  L_{\mathrm{AUL}}(\bm{s}, y) = \frac{(a - p_y)^q - (a-1)^q}{q}
\end{equation*}
where $a > 1$ and $q > 0$. It is asymmetric when $\mathbb{I}(q \leq 1) (\frac{a}{a-1})^{q-1} + \mathbb{I} (q \leq 1) \leq 1/ \tilde{r}$.

\textbf{Asymmetric Exponential Loss} (AEL)
\begin{equation*}
  L_{\mathrm{AEL}}(\bm{s}, y) = e^{-p_y/q}
\end{equation*}
where $q > 0$. It is asymmetric when $ e^{1/q} \leq 1/ \tilde{r}$.

\subsection{Combined Loss Functions}
\textbf{Generalized Cross Entropy} (GCE; \citealt{gce})
\begin{equation*}
  L_{\mathrm{GCE}}(\bm{s}, y) = \frac{1 - p_y^q}{q}
\end{equation*}
can be viewed as a smooth interpolation between CE and MAE, where $0 < q \leq 1$. CE or MAE can be recovered by setting $q \to 0$ or $q=1$.

\textbf{Symmetric Cross Entropy} (SCE; \citealt{sce})
\begin{align*}
  L_{\mathrm{SCE}}(\bm{s}, y) &= a \cdot L_{\mathrm{CE}}(\bm{s}, y) + b \cdot  L_{\mathrm{RCE}}(\bm{s}, y) \\
  &\propto (1-q) \cdot  (-\log p_i) + q \cdot  (1 - p_i)
\end{align*}
is a weighted average of CE and RCE (MAE), where $a >  0$, $b > 0$, and $0 < q < 1$.

\textbf{Taylor Cross Entropy} (TCE; \citealt{taylor})
\begin{equation*}
  L_{\mathrm{TCE}}(\bm{s}, y) = \sum_{i=1}^{q}\frac{(1 - p_y)^i}{i}
\end{equation*}
is derived from Taylor series of the $\log$ function. It reduces to MAE when $q=1$. Interestingly, the summand of TCE $(1 - p_y)^i/i$ with $i>2$ is proportional to AUL with $a = 1$ and $q = i$. Thus TCE can be viewed as a combination of symmetric and asymmetric loss functions.

\textbf{Active-Passive Loss} (APL; \citealt{actpass})

\citet{actpass} propose weighted combinations of active and passive loss functions. We include NCE+MAE as an example:
\begin{align*}
  L_{\mathrm{NCE+MAE}}(\bm{s}, y) &= a \cdot L_{\mathrm{NCE}}(\bm{s}, y) + b \cdot L_{\mathrm{MAE}}(\bm{s}, y) \\
  &\propto (1-q) \cdot \frac{-\log p_y}{\sum_{i=1}^{k} -\log p_i} + q \cdot (1-p_y)
\end{align*}
where $a >  0$, $b > 0$, and $0 < q < 1$.

\subsection{Loss Functions with Additional Regularizers}
\label{app:losses:regularizer}
We additionally review loss functions that implicitly involve a regularizer and a primary loss function following the standard form \cref{eq:normalized}. See \cref{table:weight_regularizer} for a summary. In addition to the sample-weighting curriculums implicitly defined by the primary loss function, the additional regularizer complicates the analysis of the training dynamics. We leave investigations on how these regularizers affect noise robustness for future work.

\begin{table}[t]
  \centering
  \small
  \renewcommand{\arraystretch}{2}
  \begin{tabular}{ c | c | c | c  }
    Name & Original                                                                      & Primary Loss                                                            & Regularizer                                  \\
    \midrule
    MSE  & $1 - 2p_y +  \sum_{i=1}^k p_i^2$                                              & $1-p_y$                                                                 & $\sum_{i=1}^k p_i^2$                         \\
    PL(CR)   & $-\log p_y + \log p_{y_{n}|\bm{x}_{m}}$                                   & $-\log p_y$                                                             & $\sum_{i=1}^{k} P(\tilde{y} = i) \log p_{i}$ \\
    CE+GLS  & $-\sum_{i=1}^{k} [\mathbb{I}(i=y) (1 - \alpha) +  \frac{\alpha}{k}] \log p_i$ & $-\log p_y$                                                             & $\pm \sum_{i=1}^k \frac{1}{k}\log p_i$       \\
    NCE  & $-\log p_y /(\sum_{i=1}^{k} -\log p_i)$                                  & $-\gamma_{\mathrm{NCE}} \cdot \log p_i$ & $\sum_{i=1}^{k} \frac{1}{k}\log p_i$         \\
  \end{tabular}
  \caption{Original expressions, primary loss functions in the standard form \cref{eq:normalized} and regularizers for loss functions reviewed in \cref{app:losses:regularizer}. We view PL in its expectation to derive its regularizer. $p_{y_{n}|\bm{x}_{m}}$ is the softmax probability of a random label $y_{n}$ with a random input $\bm{x}_{m}$ sampled from the noisy data. $\gamma_{\mathrm{NCE}} = 1/(\sum_{i=1}^{k} -\log p_i)$ is a scalar wrapped with the stop-gradient operator.}
  \renewcommand{\arraystretch}{1}
  \label{table:weight_regularizer}
\end{table}

\textbf{Mean Square Error} (MSE; \citealt{mae})
\begin{align*}
  L_{\mathrm{MSE}}(\bm{s}, y) & = \sum_{i=1}^k (\mathbb{I}(i=y) - p_i)^2                 = 1 - 2p_y +  \sum_{i=1}^k p_i^2                                              \\
                              & \propto 1 - p_y + \frac{1}{2} \cdot \sum_{i=1}^k p_i^2           = L_{\mathrm{MAE}}(\bm{s}, y) + \alpha \cdot R_{\mathrm{MSE}}(\bm{s})
\end{align*}
is more robust than CE \citep{mae}, where $\alpha=0.5$ and the regularizer
\begin{equation}
  R_{\mathrm{MSE}}(\bm{s}) = \sum_{i=1}^k p_i^2
\end{equation}
increases the entropy of the softmax output. We can generalize $\alpha$ to a hyperparamter, making MSE a combination of MAE and an entropy regularizer $R_{\mathrm{MSE}}$.

\textbf{Peer Loss} (PL; \citealt{peer})
\begin{equation*}
  L_{\mathrm{PL}}(\bm{s}, y) = L(\bm{s}, y) - L(\bm{s}_{n}, y_{m})
\end{equation*}
makes a generic loss function $L(\bm{s}, y)$ robust against label noise, where $\bm{s}_{n}$ denotes the score of an input $\bm{x}_{n}$ and $y_{m}$ a label, both randomly sampled from the noisy data. Its noise robustness is theoretically established for binary classification and extended to multi-class setting \citep{peer}.

\textbf{Confidence Regularizer} (CR; \citealt{peer2})
\begin{equation*}
  R_{\mathrm{CR}}(\bm{s}) =  - \mathbb{E}_{\tilde{y}}[L(\bm{s}, \tilde{y})]
\end{equation*}
is shown \citep{peer2} to be the regularizer induced by PL in expectation. Substituting $L$ with cross entropy leads to
\begin{equation}
  R_{\mathrm{CR}}(\bm{s}) =  - \mathbb{E}_{\tilde{y}}[-\log p_{\tilde{y}}] = \sum_{i=1}^{k} P(\tilde{y} = i) \log p_{i}
\end{equation}
Minimizing $R_{\mathrm{CR}}(\bm{s})$ thus makes the softmax output distribution $\bm{p}$ deviate from the prior label distribution of the noisy dataset $P(\tilde{y} = i)$, reducing the entropy of the softmax output.

\textbf{Generalized Label Smoothing} (GLS; \citealt{gls})

\citet{ls} show that label smoothing (LS; \citealt{ls_origin}) can mitigate overfitting with label noise, which is later extended to GSL. Cross entropy with GLS is
\begin{align*}
  L_{\mathrm{CE+GLS}}(\bm{s}, y) & =  \sum_{i=1}^{k} - [\mathbb{I}(i=y) (1 - \alpha) + \frac{\alpha}{k}] \log p_i                                                                                               \\
                                 & = - (1 - \alpha) \log p_y - \alpha \cdot \frac{1}{k} \sum_{i=1}^k \log p_i                                                                                                   \\
                                 & \propto - \log p_y - \frac{\alpha}{1 - \alpha} \cdot \frac{1}{k} \sum_{i=1}^k \log p_i                 = L_{\mathrm{CE}}(\bm{s}, y) + \alpha' \cdot R_{\textrm{GLS}}(\bm{s})
\end{align*}
where $\alpha' = \alpha/(1-\alpha)$, has regularizer $R_{\textrm{GLS}}$
\begin{equation}
  R_{\mathrm{GLS}}(\bm{s}) = -\sum_{i=1}^k \frac{1}{k}\log p_i
\end{equation}
With $\alpha' > 0$, $R_{\mathrm{GLS}}$ corresponds to the original label smoothing, which increases the entropy of softmax outputs. In contrast, $\alpha' < 0$ corresponding to negative label smoothing \citep{gls}, which decreases the output entropy similar to $R_{\mathrm{CR}}$.

\subsubsection{Derivations for NCE}
\label{app:losses:regularizer:nce}

\textbf{Deriving \cref{eq:curriculum:nce} \ } With equivalent derivatives, since
\begin{align*}
  \nabla_{\bm{s}} L_{\mathrm{NCE}}(\bm{s}, y)
   & = \frac{\nabla_{\bm{s}} L_{\mathrm{CE}}(\bm{s}, y) \cdot \left[\sum_{i=1}^{k} L_{\mathrm{CE}}(\bm{s}, i)\right] - \nabla_{\bm{s}} \left[\sum_{i=1}^{k} L_{\mathrm{CE}}(\bm{s}, i)\right] \cdot L_{\mathrm{CE}}(\bm{s}, y)}{\left[\sum_{i=1}^{k} L_{\mathrm{CE}}(\bm{s}, i)\right]^2} \\
   & = \frac{1}{\sum_{i=1}^{k} L_{\mathrm{CE}}(\bm{s}, i)} \left\{ \nabla_{\bm{s}} L_{\mathrm{CE}}(\bm{s}, y) + \frac{k L_{\mathrm{CE}}(\bm{s}, y)}{\sum_{i=1}^{k} L_{\mathrm{CE}}(\bm{s}, i)} \cdot \nabla_{\bm{s}} \left[\sum_{i=1}^{k}  -\frac{1}{k} L_{\mathrm{CE}}(\bm{s}, i)\right]  \right\} \\
   & = \gamma_{\mathrm{NCE}} \cdot \left[ \nabla_{\bm{s}} L_{\mathrm{CE}}(\bm{s}, y) + \epsilon_{\mathrm{NCE}} \cdot \nabla_{\bm{s}} R_{\mathrm{NCE}}(\bm{s}) \right],
\end{align*}
NCE can be rewritten as
\begin{equation*}
  L_{\mathrm{NCE}}(\bm{s}, y) = \gamma_{\mathrm{NCE}} \cdot  L_{\mathrm{CE}}(\bm{s}, y) + \gamma_{\mathrm{NCE}} \cdot  \epsilon_{\mathrm{NCE}} \cdot  R_{\mathrm{NCE}}(\bm{s})
\end{equation*}
where $\gamma_{\mathrm{NCE}} = 1/(\sum_{i=1}^{k} -\log p_i)$ and $\epsilon_{\mathrm{NCE}} = k (-\log p_y)/(\sum_{i=1}^{k}-\log p_i)$ are scalar weights wrapped with the stop-gradient operator as discussed in \cref{sec:curriculum:nce}. The regularizer
\begin{equation*}
  R_{\mathrm{NCE}}(\bm{s}) = \sum_{i=1}^{k} \frac{1}{k}\log p_i
\end{equation*}
has a form similar to $R_{\mathrm{GLS}}$.

\textbf{Deriving $\hat{w}_{\mathrm{NCE}}$ of \cref{eq:nce:weight}\ \ } Here we derive the upperbound of $\|\nabla_{\bm{s}} L_{\mathrm{NCE}}(\bm{s}, y)\|_1$ discussed in \cref{sec:curriculum:nce}:
\begin{align*}
  \|\nabla_{\bm{s}} L_{\mathrm{NCE}}(\bm{s}, y)\|_1  & \leq \gamma_{\mathrm{NCE}} \cdot \left(\|\nabla_{\bm{s}} L_{\mathrm{CE}}(\bm{s}, y)\|_1 + \epsilon_{\mathrm{NCE}} \cdot  \|\nabla_{\bm{s}} R_{\mathrm{NCE}}(\bm{s})\|_1 \right)                      \\
                   & \leq \gamma_{\mathrm{NCE}} \cdot \left(\|\nabla_{\bm{s}} L_{\mathrm{CE}}(\bm{s}, y)\|_1 + \epsilon_{\mathrm{NCE}} \cdot  \frac{1}{k}\sum_{i=1}^k \|\nabla_{\bm{s}} L_{\mathrm{CE}}(\bm{s}, i)\|_1 \right) \\
                   & = \gamma_{\mathrm{NCE}} \cdot \left(w_{\mathrm{CE}} \cdot  \|\nabla_{\bm{s}} \Delta_{y}\|_1 + \epsilon_{\mathrm{NCE}} \cdot  \frac{1}{k}\sum_{i=1}^k w_{\mathrm{CE}}  \cdot \|\nabla_{\bm{s}} \Delta_{i}\|_1 \right) \\
                   & = 2 \gamma_{\mathrm{NCE}} \cdot w_{\mathrm{CE}} \left(1 + \epsilon_{\mathrm{NCE}} \right) \\
                   & = \hat{w}_{\mathrm{NCE}}
\end{align*}
The derivation is based on the inequality $ |x \pm y| \leq |x| + |y|$ and the fact that $\|\nabla_{\bm{s}} \Delta_{i}\|_1 = 2$. The latter can be proved by straightforward calculations. Given
\begin{equation*}
  \frac{\partial \Delta_i}{\partial s_j} = \left\{ \begin{array}{lr}1,                                                             & j=i      \\
             -\frac{e^{s_{j}}}{\sum_{k \neq i}e^{s_{k}}} = -\frac{p_j}{1 - p_i}, & j \neq i\end{array}
  \right.
\end{equation*}
we then have
\begin{equation*}
  \|\nabla_{\boldsymbol{s}} \Delta_{i}\|_1 = \sum_j \left|\frac{\partial \Delta_i}{\partial s_j}\right|  = 1 + \sum_{j \neq i} \frac{p_j}{1-p_i} = 1 + 1 = 2
\end{equation*}

\section{Detailed Experimental Settings}
\label{app:setting}

\textbf{Label noise\ } The synthetic noisy labels are generated following \citep{actpass,asymmetric,loss_corerct}. For symmetric label noise, the training labels are randomly flipped to a different class with probabilities $\eta \in \{0.2, 0.4, 0.6, 0.8\}$. Asymmetric label noise is generated from a class-dependent flipping pattern. On CIFAR100, the 100 classes are grouped into 20 super-classes, each having 5 sub-classes. Each class is flipped within the same super-class into the next in a circular fashion. The flip probabilities are $\eta \in \{0.1, 0.2, 0.3, 0.4\}$. Human label noise for CIFAR10/100 are adopted from \citet{cifarn}. We use the ``worst'' labels of CIFAR10-N and the ``fine'' labels of CIFAR100-N, both leading to $\eta = 0.4$.

\textbf{Models and hyperparameters \ } We use a 4-layer CNN for MNIST, an 8-layer CNN for CIFAR10, a ResNet-34 \citep{resnet} for CIFAR100, and a ResNet-50 \citep{resnet} for WebVision, all with batch normalization \citep{bn}. Data augmentation on CIFAR10/100 include random width/height shift and horizontal flip. On WebVision, we additionally include random cropping and color jittering. Without further specifications, all models are trained using SGD with momentum 0.9 and batch size 128 for 50, 120, 200 and 250 epochs on MNIST, CIFAR10, CIFAR100 and WebVision, respectively. Learning rates with cosine annealing are 0.01 on MNIST and CIFAR10, 0.1 on CIFAR100, and 0.2 on WebVision. Weight decays are $10^{-3}$ on MNIST, $10^{-4}$ on CIFAR10, $10^{-5}$ on CIFAR100 and $3 \times 10^{-5}$ on WebVision. All loss functions are normalized to have unit maximum in sample weights, which is different from \citep{actpass}. Hyperparameters of loss functions are listed in \cref{table:underfit_param,table:robust_param} for different experiments.

\section{Additional Results to Understand Robust Loss Functions}
We complement \cref{sec:dynamics} in the main text with detailed derivations and additional results. \cref{app:dynamics:underfit} extends discussions in  \cref{sec:dynamics:underfit} while \cref{app:dynamics:robust} extends \cref{sec:dynamics:robust}

\begin{table}[t]
  \centering
  \small
  \begin{tabular}{c | c  c   c  c  c | c c }
    & SCE & GCE & NCE+MAE & AUL & AGCE & AUL\textsuperscript{\textdagger} & AGCE\textsuperscript{\textdagger} \\
  \midrule
    $a$ &  /   &    / &   /  & 1.1 & 0.1 & 3.0 & 1.6 \\
  $q$ & 0.7 & 0.3 & 0.3 & 5.0   & 0.1 & 0.7 & 2.0 \\
  \end{tabular}
  \caption{Hyperparameters of different loss functions for results in \cref{sec:dynamics:underfit,app:dynamics:underfit}. They are tuned on CIFAR100 without label noise. Settings with inferior hyperparameters are denoted with \textdagger.}
  \label{table:underfit_param}
\end{table}

\subsection{Understanding Underfitting of Robust Loss Functions}
\label{app:dynamics:underfit}

\textbf{Hyperparameters\ } We list the hyperparameters tuned on CIFAR100 without label noise in \cref{table:underfit_param} for experiments in \cref{sec:dynamics:underfit,app:dynamics:underfit}.

\begin{figure}
  \centering
  \begin{subfigure}[b]{0.32\textwidth}
    \centering
    \includegraphics[width=\textwidth]{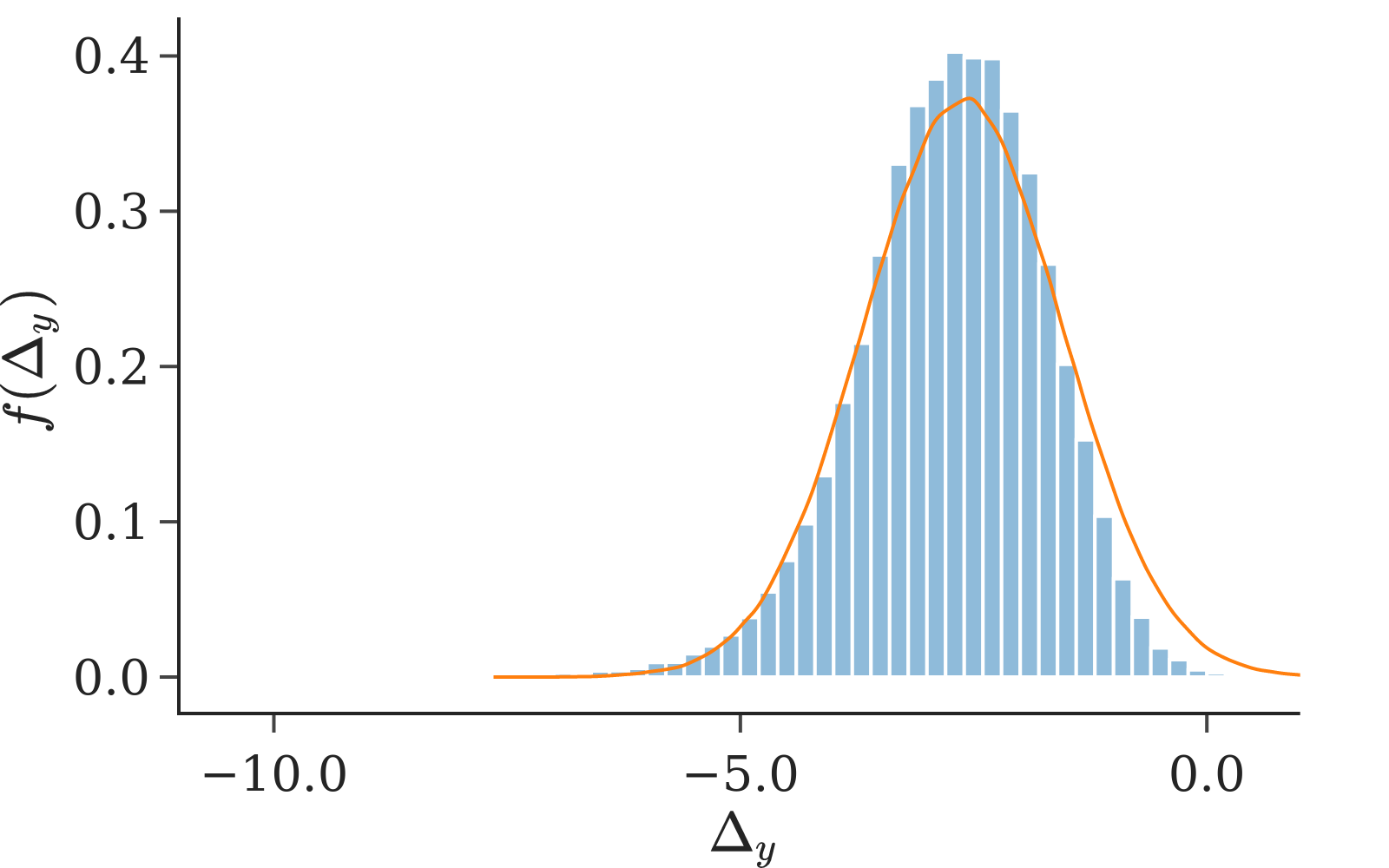}
    \caption{MNIST}
    \label{fig:simulate:mnist}
  \end{subfigure}
  \hfill
  \begin{subfigure}[b]{0.32\textwidth}
    \centering
    \includegraphics[width=\textwidth]{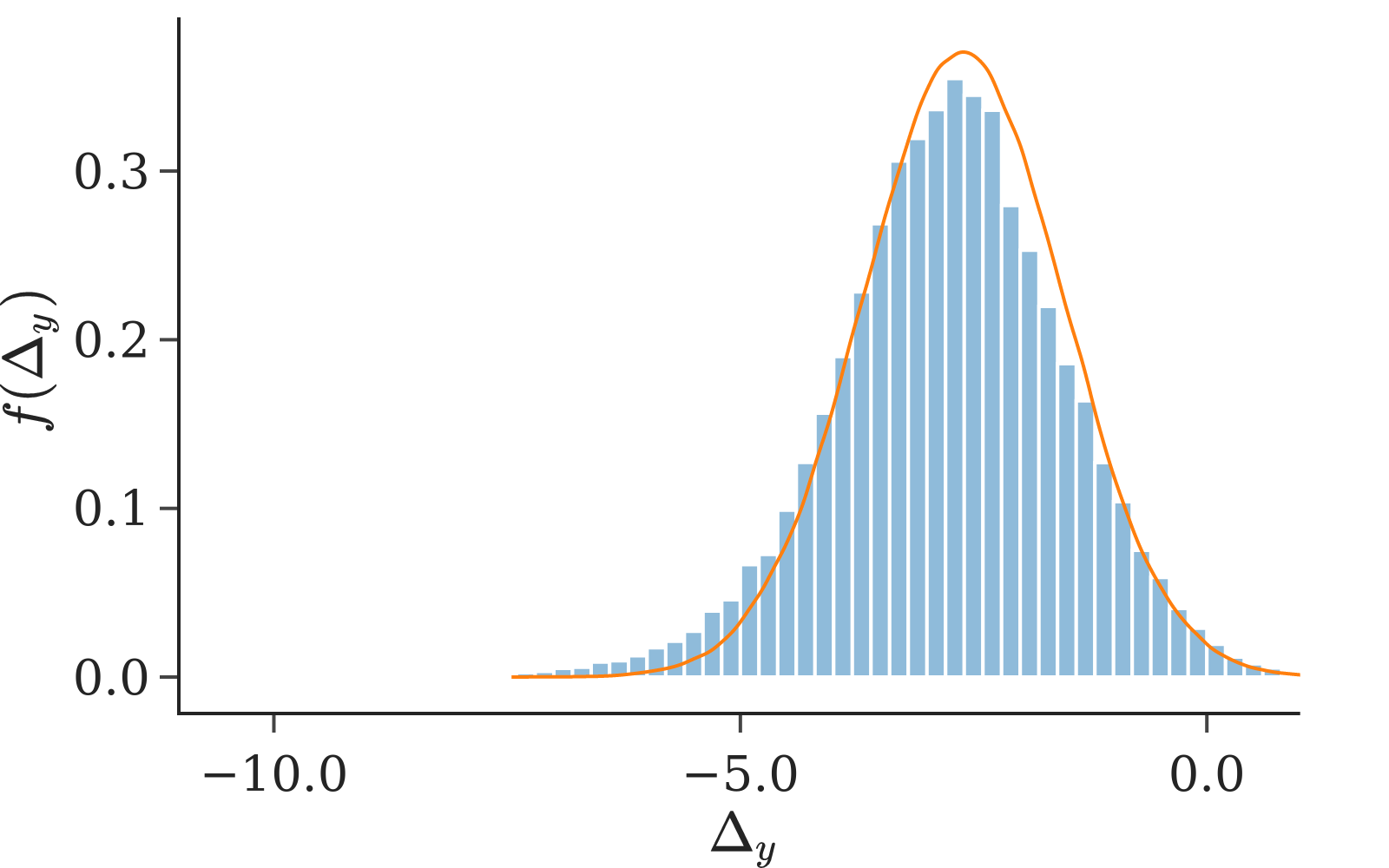}
    \caption{CIFAR10}
    \label{fig:simulate:cifar10}
  \end{subfigure}
  \hfill
  \begin{subfigure}[b]{0.32\textwidth}
    \centering
    \includegraphics[width=\textwidth]{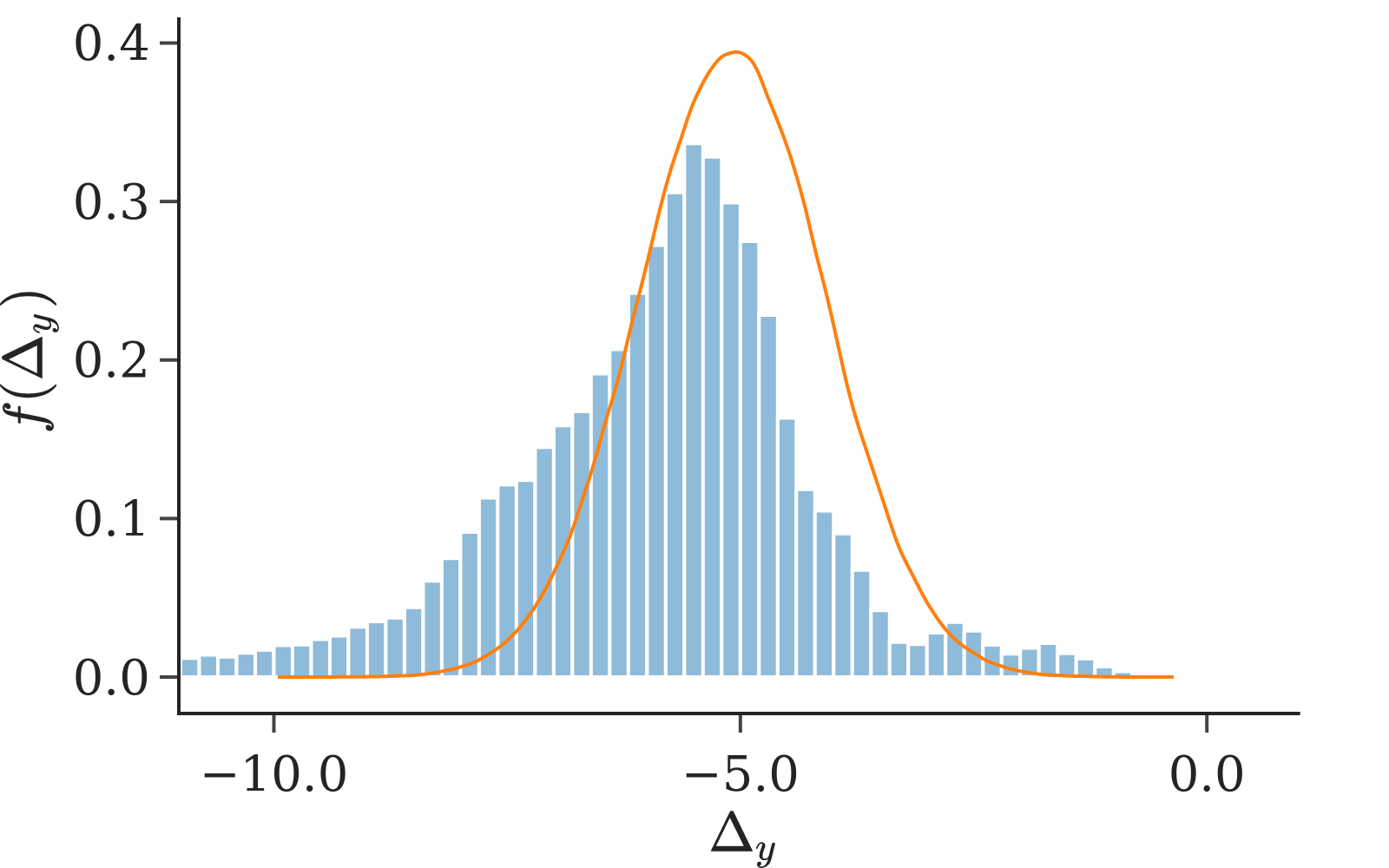}
    \caption{CIFAR100}
    \label{fig:simulate:cifar100}
  \end{subfigure}

  \begin{subfigure}[b]{0.32\textwidth}
    \centering
    \includegraphics[width=\textwidth]{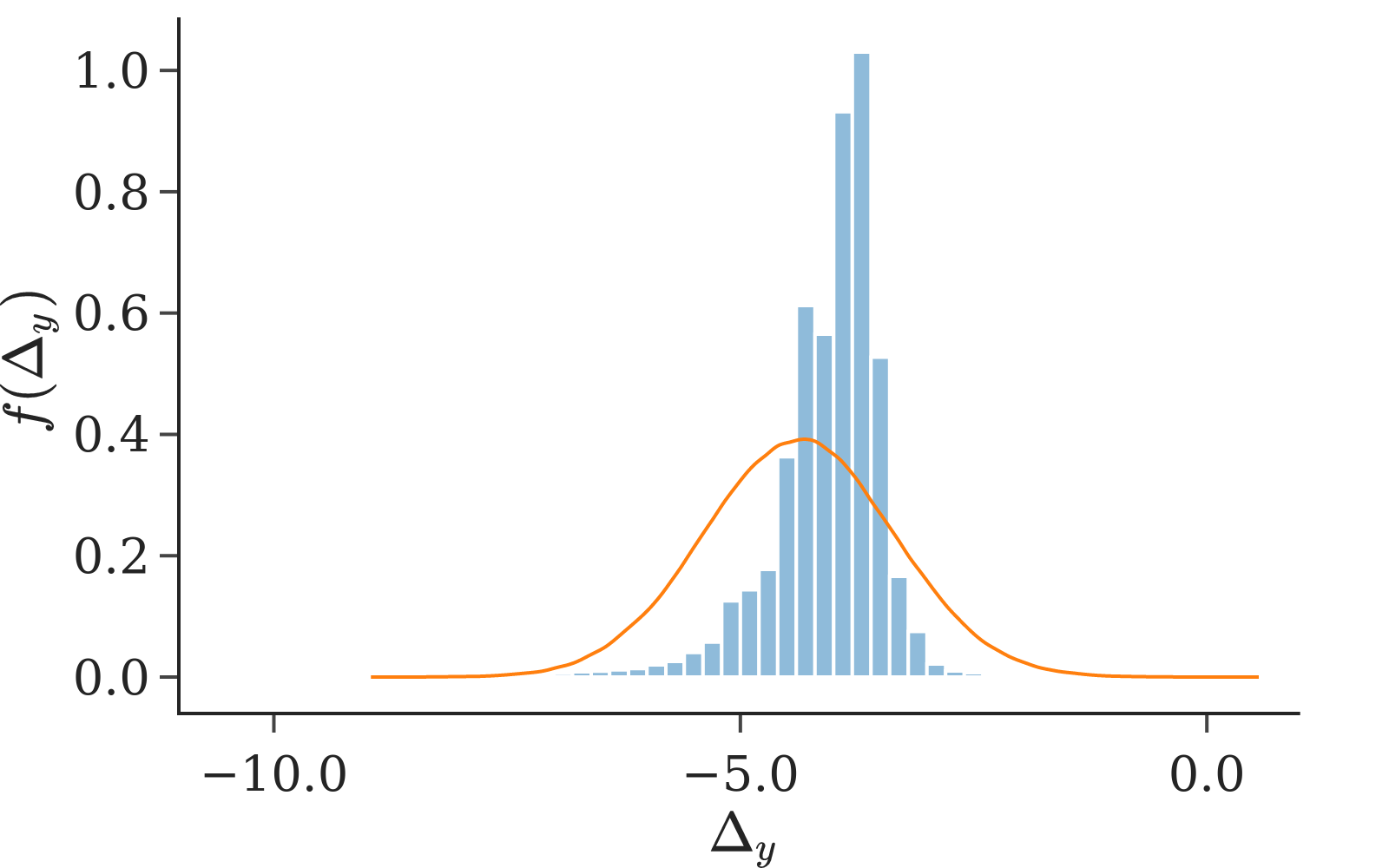}
    \caption{WebVision50}
    \label{fig:simulate:webvision50}
  \end{subfigure}
  \hfill
  \begin{subfigure}[b]{0.32\textwidth}
    \centering
    \includegraphics[width=\textwidth]{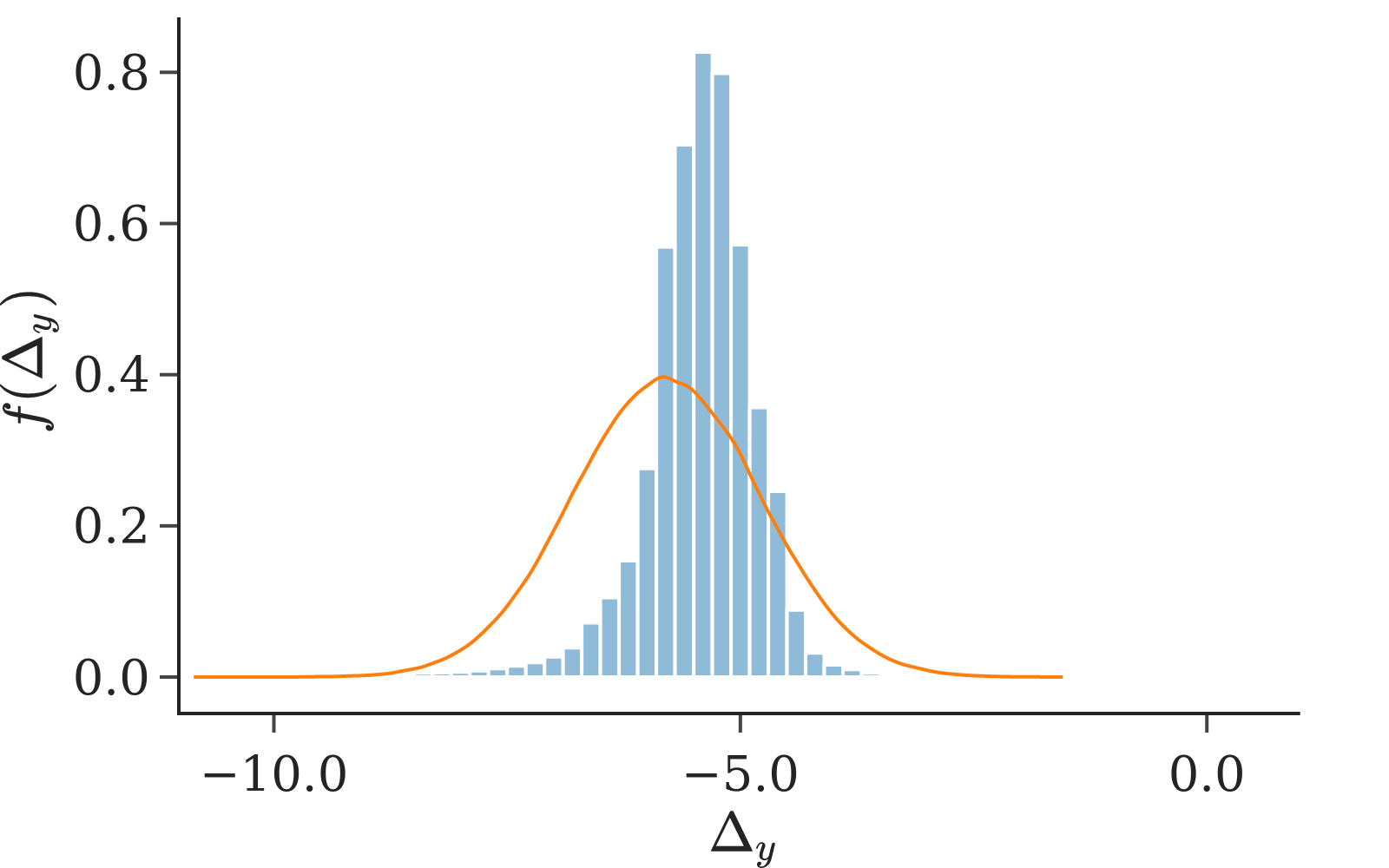}
    \caption{WebVision200}
    \label{fig:simulate:webvision200}
  \end{subfigure}
  \hfill
  \begin{subfigure}[b]{0.32\textwidth}
    \centering
    \includegraphics[width=\textwidth]{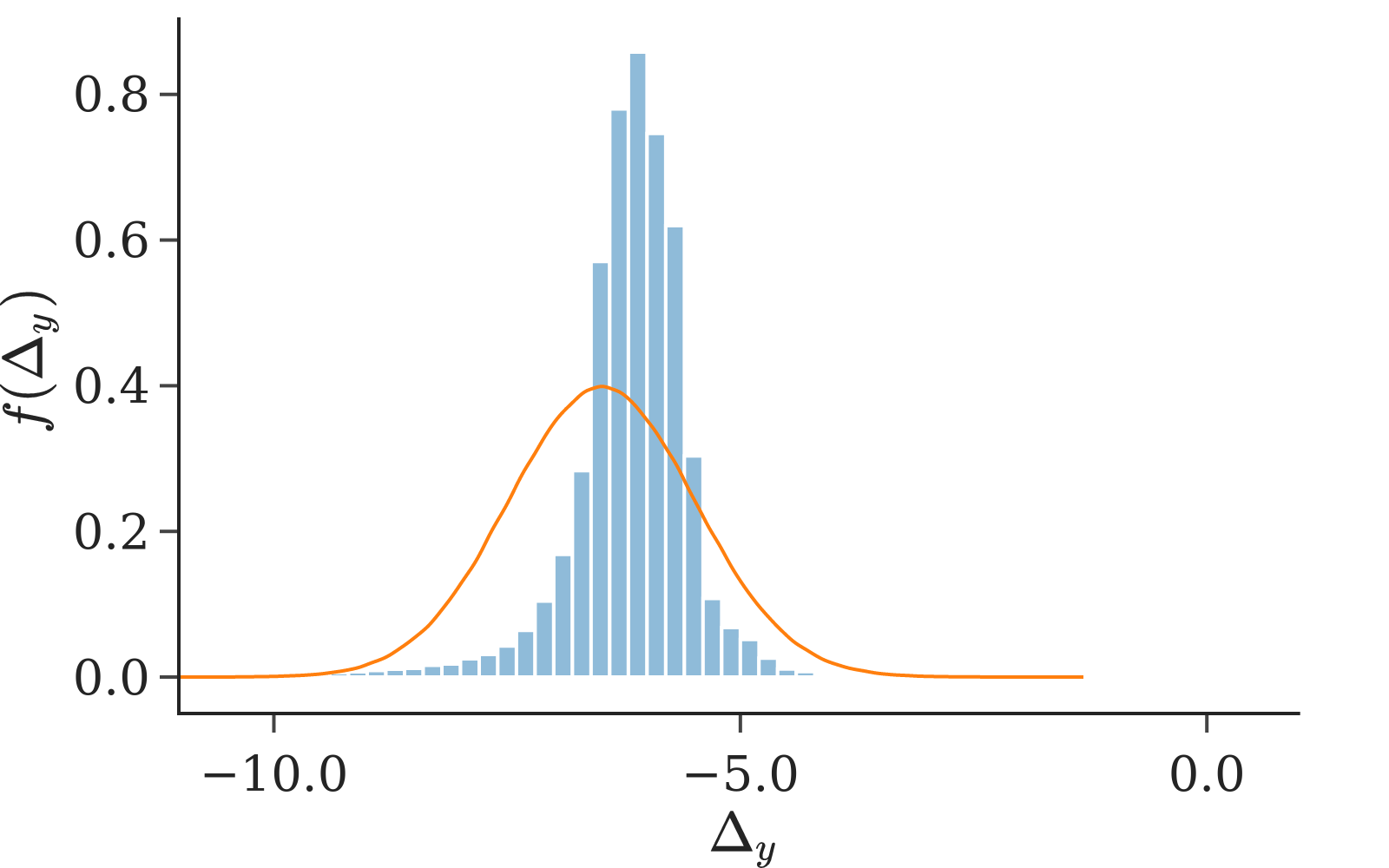}
    \caption{WebVision400}
    \label{fig:simulate:webvision400}
  \end{subfigure}
  \caption{Comparisons between simulated and real $\Delta_y$ distributions at initialization. The simulations are based on the assumption that class scores follow normal distribution $s_i \sim \mathcal{N}(0, 1)$ at initialization and plotted as curves. Real distributions are extracted with randomly initialized models and plotted as histograms. The vertical axis denotes probability density $f(\Delta_y)$.}
  \label{fig:simulate}
\end{figure}

\textbf{Simulated $\Delta_y$ approximate real settings.} We compare the simulated $\Delta_y$ distributions based on our assumptions in \cref{sec:dynamics:underfit} to distributions of real datasets at initialization in \cref{fig:simulate}. The expectations of simulated $\Delta_y$ follow real settings, which supports the analysis in \cref{sec:dynamics:underfit}.

\textbf{Derivation of \cref{eq:difficult}\ }  Assume that class scores $s_i$ at \emph{initialization} are i.i.d. normal variables, i.e., $s_i \sim \mathcal{N}(\mu, \sigma)$,
\begin{align*}
  \mathbb{E}(\Delta_y) & = \mathbb{E}[s_y -\log \sum_{i\neq y} e^{s_i}] =\mu - \mathbb{E}[\log \sum_{i\neq y} e^{s_i}]                                                \\
                       & \approx_1 \mu  - \log \mathbb{E}[\sum_{i\neq y} e^{s_i}] + \frac{\mathbb{V}[\sum_{i\neq y} e^{s_i}]}{2 \mathbb{E}[\sum_{i\neq y} e^{s_i}]^2} \\
                       & =_2 \mu -\log \{(k-1) \mathbb{E}[e^{s_y}]\} + \frac{(k-1)\mathbb{V}[e^{s_y}]}{2 \{(k-1)\mathbb{E}[e^{s_y}]\}^2}                              \\
                       & =_3 \mu -\log [(k-1) e^{\mu + \sigma^2/2}] + \frac{(k-1)(e^{\sigma^2}-1)e^{2\mu + \sigma^2}}{2 [(k-1)e^{\mu + \sigma^2/2}]^2}                \\
                       & = -\log(k-1) - \sigma^2/2 + \frac{e^{\sigma^2}-1}{2 (k-1)}                                                                       \\
\end{align*}
where $\approx_1$ follows the approximation with Taylor expansion $\mathbb{E}[\log X] \approx \log \mathbb{E}[X] - \mathbb{V}[X]/(2\mathbb{E}[X]^2)$ \citep{Elog}, $=_2$ utilizes properties of sum of log-normal variables \citep{lognorm}, and $=_3$ substitutes $\mathbb{E}[e^{s_y}]$ and $\mathbb{V}[e^{s_y}]$ with expressions for log-normal distributions.

\begin{table}[t]
  \centering
  \small
  \begin{tabular}{ l | c | c c c c | c c c c | c }
    \multicolumn{1}{c|}{}         & \multicolumn{1}{c|}{Clean}
                                  & \multicolumn{4}{c|}{Symmetric} & \multicolumn{4}{c|}{Asymmetric} & \multicolumn{1}{c}{Human}                                           \\
    \multicolumn{1}{c|}{Settings} & 0.0                            & 0.2                             & 0.4                       & 0.6 & 0.8 & 0.1 & 0.2 & 0.3 & 0.4 & 0.4 \\
    \midrule
    MAE shift                     & 2.0                            & 2.6                             & 3.0                       & 4.0 & 4.0 & 2.6 & 2.6 & 3.0 & 3.0 & 2.6 \\
    MAE scale                     & 2.0                            & 2.6                             & 2.6                       & 3.0 & 4.0 & 2.6 & 2.6 & 3.0 & 3.0 & 2.6 \\
    \midrule
    AGCE shift                    & 4.0                            & 4.0                             & 4.0                       & 5.0 & 5.0 & 4.0 & 4.0 & 4.0 & 4.0 & 4.0 \\
    AGCE scale                    & 2.6                            & 3.0                             & 4.0                       & 5.0 & 5.0 & 3.0 & 3.0 & 3.0 & 3.0 & 4.0 \\
  \end{tabular}
  \caption{Hyperparameter $\tau$ of $w^{*}(\Delta_y)$ and $w^{+}(\Delta_y)$ for different noise settings on CIFAR100. For reference, $\mathbb{E}[\Delta_y] = 5.097$ on CIFAR100 and $\mathbb{E}[\Delta_y] = 2.717$ on CIFAR10. Better performance can be achieved with a more thorough hyperparameter search. }
  \label{table:correct_param}
\end{table}

\begin{table}[t]
  \centering
  \small
  \begin{tabular}{ l | r | r r r r  }
    \multicolumn{1}{c|}{}                   & \multicolumn{1}{c|}{Clean}    & \multicolumn{4}{c}{Symmetric Noise (Noise Rate $\eta$)}                                                                                                    \\
    Loss                                    & \multicolumn{1}{c|}{$\eta=0$} & \multicolumn{1}{c}{$\eta=0.2$}                          & \multicolumn{1}{c}{$\eta=0.4$} & \multicolumn{1}{c}{$\eta=0.6$} & \multicolumn{1}{c}{$\eta=0.8$} \\
    \midrule
    CE\textsuperscript{\textdaggerdbl}      & 71.33 $\pm$ 0.43              & 56.51 $\pm$ 0.39                                        & 39.92 $\pm$ 0.10               & 21.39 $\pm$ 1.17               & 7.59 $\pm$ 0.20                \\
    \midrule
    GCE\textsuperscript{\textdaggerdbl}     & 63.09 $\pm$ 1.39              & 61.57 $\pm$ 1.06                                        & 56.11 $\pm$ 1.35               & 45.28 $\pm$ 0.61               & 17.42 $\pm$ 0.06               \\
    NCE\textsuperscript{\textdaggerdbl}     & 29.96 $\pm$ 0.73              & 25.27 $\pm$ 0.32                                        & 19.54 $\pm$ 0.52               & 13.51 $\pm$ 0.65               & 8.55 $\pm$ 0.37                \\
    NCE+AUL\textsuperscript{\textdaggerdbl} & 68.96 $\pm$ 0.16              & 65.36 $\pm$ 0.20                                        & 59.25 $\pm$ 0.23               & 46.34 $\pm$ 0.21               & 23.03 $\pm$ 0.64               \\
    \midrule
    AGCE                                    & 49.27 $\pm$ 1.03              & 49.17 $\pm$ 2.15                                        & 47.76 $\pm$ 1.75               & 38.17 $\pm$ 1.43               & 16.03 $\pm$ 0.59               \\
    AGCE shift                              & 69.39 $\pm$ 0.84              & 62.81 $\pm$ 0.42                                        & 48.21 $\pm$ 1.06               & 36.70 $\pm$ 2.89               & 14.49 $\pm$ 0.17               \\
    AGCE scale                              & 70.57 $\pm$ 0.62              & 64.73 $\pm$ 0.98                                        & 56.69 $\pm$ 0.33               & 39.02 $\pm$ 1.20               & 14.64 $\pm$ 0.79               \\
    \midrule
    MAE                                     & 3.69  $\pm$ 0.59              & 2.92 $\pm$ 0.46                                         & 1.29 $\pm$ 0.50                & 2.27 $\pm$ 1.24                & 1.00 $\pm$ 0.00                \\
    MAE shift                               & 69.02 $\pm$ 0.78              & 59.75 $\pm$ 0.84                                        & 44.60 $\pm$ 0.24               & 24.27 $\pm$ 0.26               & 8.08 $\pm$ 0.26                \\
    MAE scale                               & \textbf{70.97  $\pm$ 0.41}             & \textbf{66.83  $\pm$ 0.84}                                       & \textbf{60.57  $\pm$ 1.04}              & \textbf{49.23  $\pm$ 1.22}              & \textbf{24.44  $\pm$ 0.73}              \\
  \end{tabular}
  \caption{Addition results to \cref{table:correct} with more symmetric label noise rates on CIFAR100.}
  \label{table:correct_symmetric}
\end{table}

\begin{table}[t]
  \centering
  \small
  \begin{tabular}{ l | r | r r r r  }
    \multicolumn{1}{c|}{}                   & \multicolumn{1}{c|}{Clean}    & \multicolumn{4}{c}{Asymmetric Noise (Noise Rate $\eta$)}                                                                                                    \\
    Loss                                    & \multicolumn{1}{c|}{$\eta=0$} & \multicolumn{1}{c}{$\eta=0.1$}                           & \multicolumn{1}{c}{$\eta=0.2$} & \multicolumn{1}{c}{$\eta=0.3$} & \multicolumn{1}{c}{$\eta=0.4$} \\
    \midrule
    CE\textsuperscript{\textdaggerdbl}      & 71.33 $\pm$ 0.43              & 64.85 $\pm$ 0.37                                         & 58.11 $\pm$ 0.32               & 50.68 $\pm$ 0.55               & 40.17 $\pm$ 1.31               \\
    \midrule
    GCE\textsuperscript{\textdaggerdbl}     & 63.09 $\pm$ 1.39              & 63.01 $\pm$ 1.01                                         & 59.35 $\pm$ 1.10               & 53.83 $\pm$ 0.64               & 40.91 $\pm$ 0.57               \\
    NCE\textsuperscript{\textdaggerdbl}     & 29.96 $\pm$ 0.73              & 27.59 $\pm$ 0.54                                         & 25.75 $\pm$ 0.50               & 24.28 $\pm$ 0.80               & 20.64 $\pm$ 0.40               \\
    NCE+AUL\textsuperscript{\textdaggerdbl} & 68.96 $\pm$ 0.16              & 66.62 $\pm$ 0.09                                         & 63.86 $\pm$ 0.18               & 50.38 $\pm$ 0.32               & 38.59 $\pm$ 0.48               \\
    \midrule
    AGCE                                    & 49.27 $\pm$ 1.03              & 47.53 $\pm$ 0.73                                         & 46.77 $\pm$ 2.37               & 39.82 $\pm$ 2.70               & 33.40 $\pm$ 1.57               \\
    AGCE shift                              & 69.39 $\pm$ 0.84              & 63.03 $\pm$ 0.42                                         & 55.84 $\pm$ 0.78               & 49.05 $\pm$ 0.81               & 40.76 $\pm$ 0.74               \\
    AGCE scale                              & 70.57 $\pm$ 0.62              & 67.13 $\pm$ 0.60                                         & 59.71 $\pm$ 0.10               & 48.23 $\pm$ 0.29               & 39.71 $\pm$ 0.17               \\
    \midrule
    MAE                                     & 3.69  $\pm$ 0.59              & 3.59  $\pm$ 0.56                                         & 3.19  $\pm$ 0.98               & 2.11  $\pm$ 1.93               & 2.53  $\pm$ 1.34               \\
    MAE shift                               & 68.57  $\pm$ 0.54             & 63.44  $\pm$ 0.32                                        & 56.47  $\pm$ 0.48              & 48.79  $\pm$ 1.22              & 39.83  $\pm$ 0.18              \\
    MAE scale                               & \textbf{70.97  $\pm$ 0.41}             & \textbf{69.50  $\pm$ 0.24}                                        & \textbf{64.80  $\pm$ 0.49}              & \textbf{59.04  $\pm$ 1.52}              & \textbf{44.48  $\pm$ 1.05}              \\
  \end{tabular}
  \caption{Addition results to \cref{table:correct} with more asymmetric label noise rates on CIFAR100.}
  \label{table:correct_asymmetric}
\end{table}

\subsubsection{Addressing Underfitting from Marginal Initial Sample Weights}
\label{app:dynamics:underfit:fix}
\textbf{Hyperparamter $\tau$ for different settings \ } The hyperparameter $\tau$ controlling the shape of modified sample-weighting functions $w^+(\Delta_y)$ and $w^*(\Delta_y)$ can affect the noise robustness. Thus we tune $\tau$ for the best performance under different noise types and noise rates, which are listed in \cref{table:correct_param}.

\textbf{Additional results with $w^*(\Delta_y)$ and $w^+(\Delta_y)$.} We report additional results under symmetric and asymmetric label noise with diverse noise rates $\eta$ in \cref{table:correct_symmetric} and \cref{table:correct_asymmetric}, respectively. Performance of MAE and AGCE gets substantially improved with $w^*(\Delta_y)$ and $w^+(\Delta_y)$.

\begin{figure}
  \centering
    \includegraphics[width=0.5\textwidth]{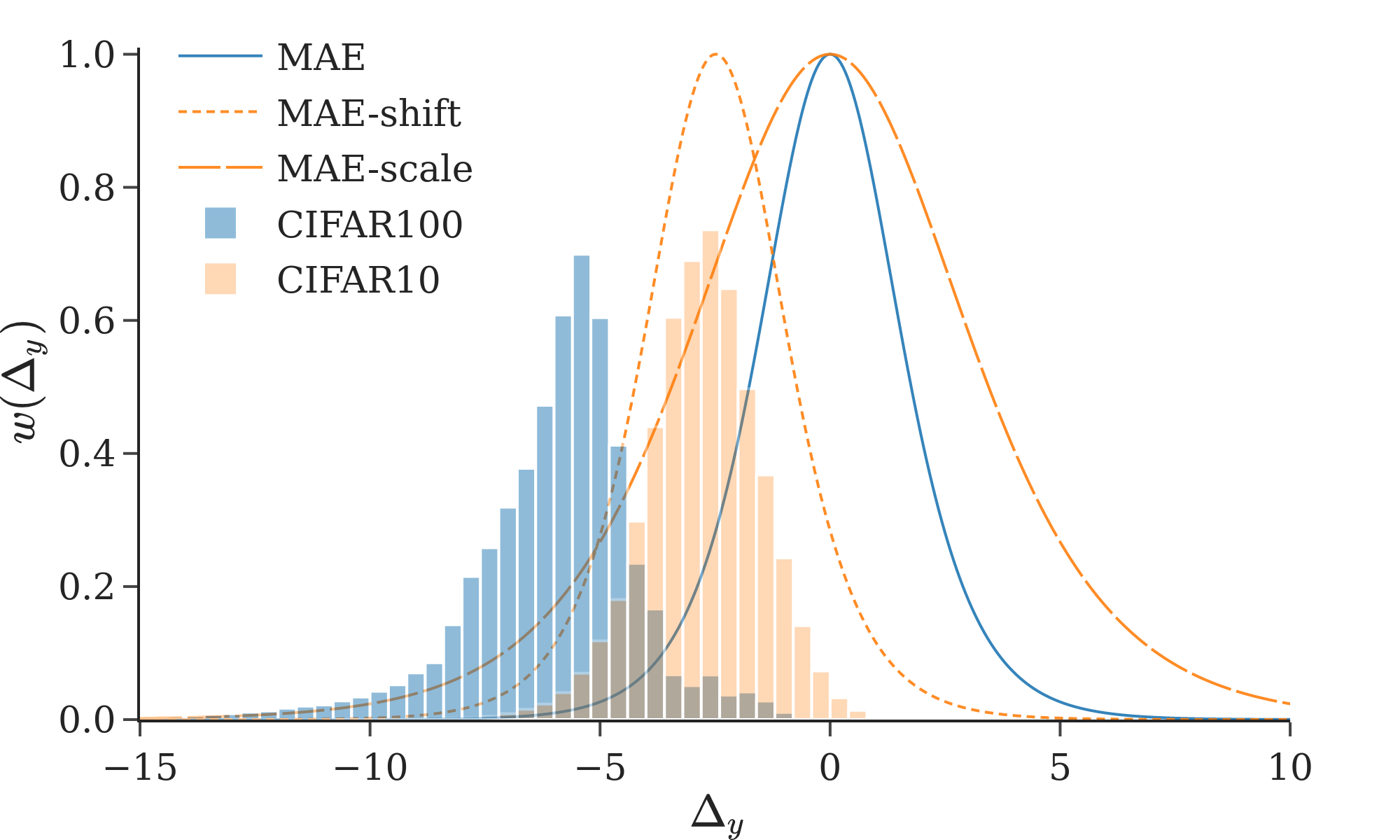}
  \caption{Shifted, scaled and the vanilla sample-weighting functions of MAE on CIFAR100. $\tau$ equals $|\mathbb{E}[\Delta_y]|$ on CIFAR10. We include the initial $\Delta_y$ distributions of CIFAR10/100 extracted with a randomly initialized model as reference.}
  \label{fig:shift_scale}
\end{figure}

\textbf{Visualization of $w^*(\Delta_y)$ and $w^+(\Delta_y)$.} In \cref{fig:shift_scale} we visualize the shifted and scaled sample-weighting functions of MAE on CIFAR100. Although both achieve the same initial sample weights at $|\mathbb{E}[\Delta_y]|$ of CIFAR100, $w^+(\Delta_y)$ diminishes much faster as $\Delta_y$ increases, leading to insufficient learning of training samples, which can explain its inferior performance in \cref{table:correct,table:correct_real,table:correct_symmetric,table:correct_asymmetric}.

\textbf{Robustness of loss functions from $w^*(\Delta_y)$ and $w^+(\Delta_y)$.}  Our proposed $w^*(\Delta_y)$ and $w^+(\Delta_y)$ aim to address the underfitting issue of robust loss functions with marginal initial sample weights. They modify $p_y$ into
\begin{equation*}
  p_y^* = \frac{1}{e^{-(\Delta_y / |\mathbb{E}[\Delta_y]| \cdot \tau)} + 1} =  \frac{1}{e^{-(\Delta_y \cdot \alpha)} + 1}
\end{equation*}
and
\begin{equation*}
  p_y^* = \frac{1}{e^{-(\Delta_y + |\mathbb{E}[\Delta_y]| - \tau)} + 1} =  \frac{1}{e^{-(\Delta_y + \beta)} + 1},
\end{equation*}
where $\alpha = \tau / |\mathbb{E}[\Delta_y]|$ and $\beta = |\mathbb{E}[\Delta_y]| - \tau$, which induces new loss functions $L^*(\bm{s}, y) = l(p_y^*)$ and $L^+(\bm{s}, y) = l(p_y^+)$, respectively. Commonly $\alpha < 1$ and $\beta > 0$ since a small $\tau$ leads to large initial sample weights and underfitting results from small $\mathbb{E}[\Delta_y]$. Notably, $\tau$ can determine the robustness of the induced loss functions. As shown in \cref{table:correct_param}, a larger noise rate $\eta$ requires a larger $\tau$ for better performance, which assigns less weights to samples with small $\Delta_y$ in general. However, our preliminary exploration find no straightforward derivation from $L(\bm{s}, y)$ being symmetric/asymmetric to $L^*(\bm{s}, y)$ and $L^+(\bm{s}, y)$ being symmetric/asymmetric. We leave the theoretical discussions to future work.

\subsection{Noise Robustness of Loss Functions}
\label{app:dynamics:robust}

\textbf{Computation of $\bar{w}_{\mathrm{clean}}$ and $\bar{w}_{\mathrm{noise}}$ for $\mathrm{snr}$ in \cref{table:snr}\ \ } The average weight for clean samples, adjusted by the learning rate at each step $\alpha_t$, can be
\begin{equation*}
  \bar{w}_{\mathrm{clean}} =  \frac{\sum_{i,t}\alpha_t \cdot \mathbb{I}(\tilde{y}_{i,t}=y_{i,t})  w_{i,t}}{\sum_{i,t}\alpha_t \cdot \mathbb{I}(\tilde{y}_{i,t}=y_{i,t})}
\end{equation*}
where $w_{i,t}$ denotes the weight of $i$-th sample of the batch at step $t$, $\tilde{y}_{i,t}$ is the potentially corrupted noisy label and $y_{i,t}$ the uncorrupted label. Similarly, for noisy samples,
\begin{equation*}
  \bar{w}_{\mathrm{noise}} =  \frac{\sum_{i,t}\alpha_t \cdot \mathbb{I}(\tilde{y}_{i,t}\neq y_{i,t})  w_{i,t}}{\sum_{i,t}\alpha_t \cdot \mathbb{I}(\tilde{y}_{i,t}\neq y_{i,t})}
\end{equation*}

\begin{figure}
  \centering
    \includegraphics[width=0.5\textwidth]{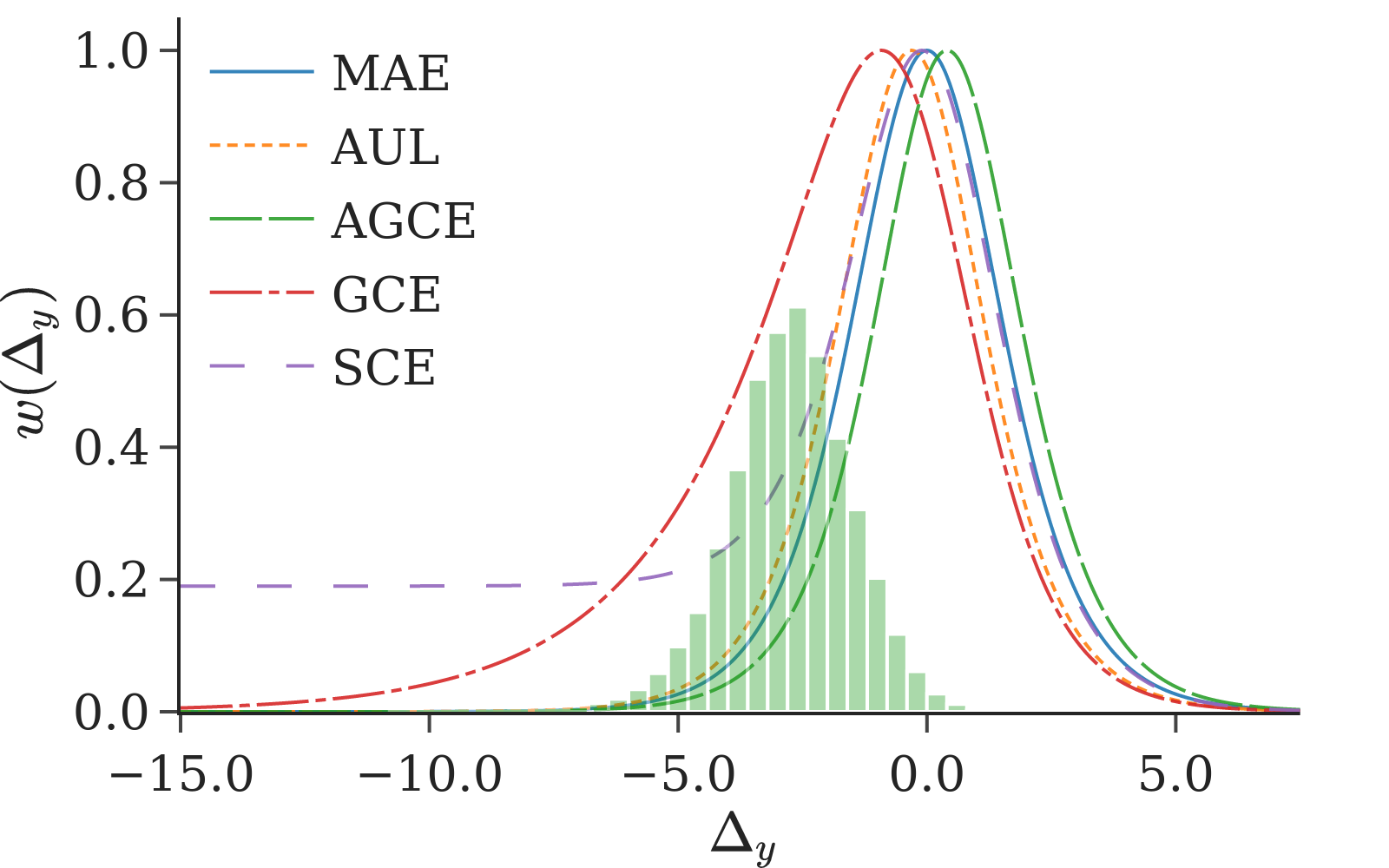}
  \caption{Plots of sample-weighting functions of loss functions used in \cref{table:snr} with hyperparameters in \cref{table:underfit_param}.}
  \label{fig:snr_weight}
\end{figure}

\textbf{Hyperparameters\ \ } We list the hyperparameters for different loss functions in \cref{table:robust_param} for results in \cref{sec:dynamics:robust,app:dynamics:robust}. In \cref{fig:snr_weight}, we plot the sample-weighting functions of different loss functions.

\begin{figure}[t]
  \centering
  \begin{subfigure}[b]{0.32\textwidth}
    \centering
    \includegraphics[width=\textwidth]{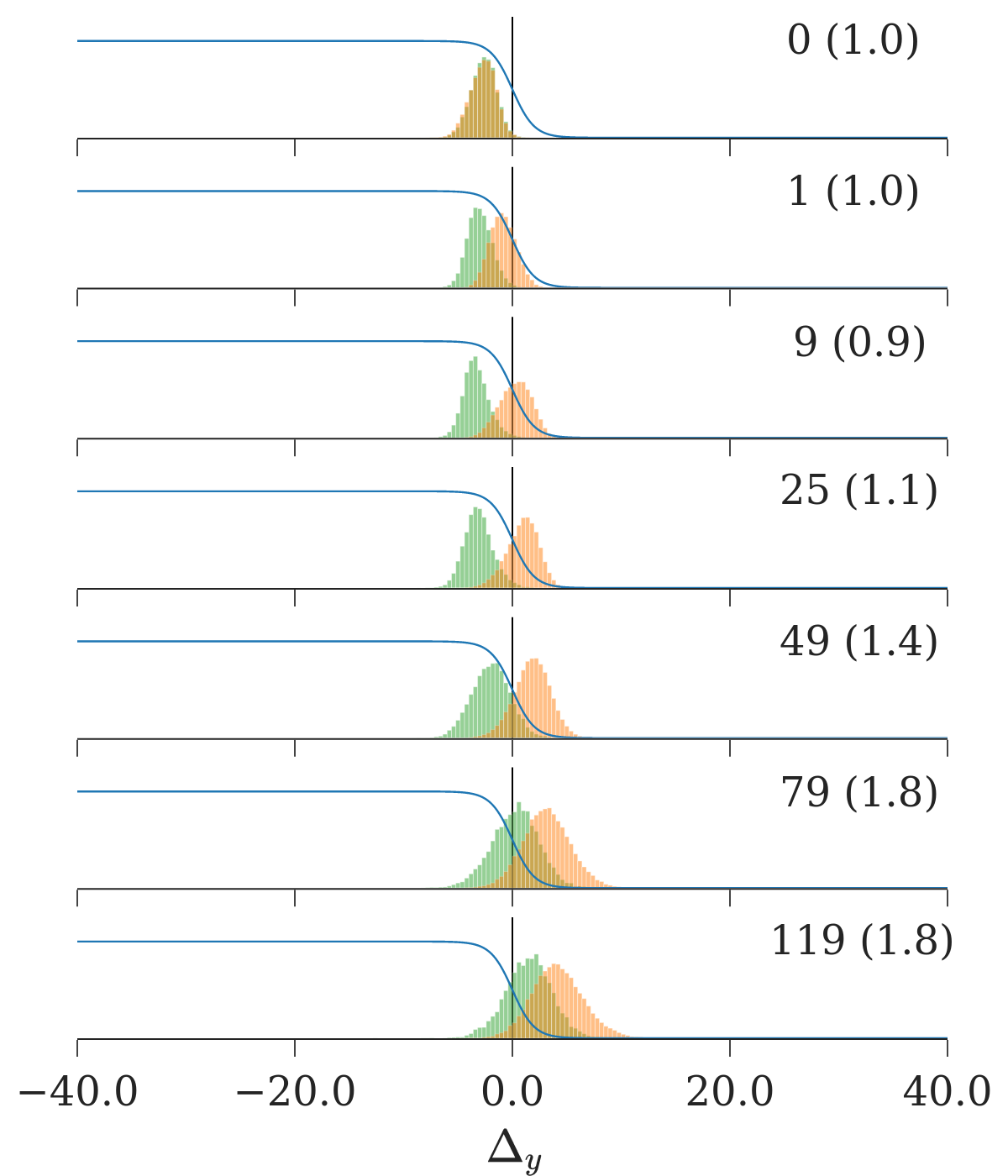}
    \caption{CE, Sym., 0.2: 74.49}
  \end{subfigure}
  \hfill
  \begin{subfigure}[b]{0.32\textwidth}
    \centering
    \includegraphics[width=\textwidth]{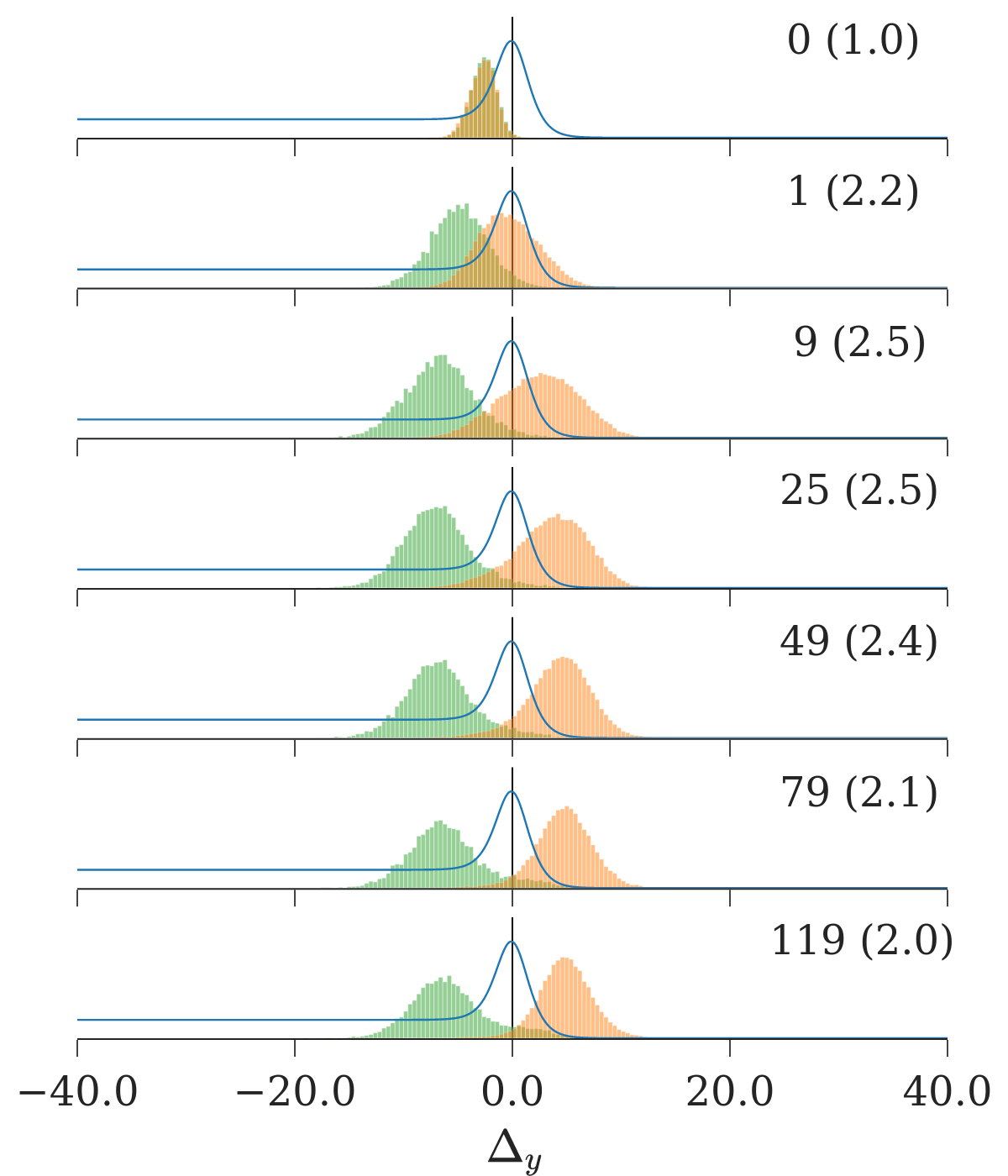}
    \caption{SCE, Sym., 0.2: 85.31}
  \end{subfigure}
  \hfill
  \begin{subfigure}[b]{0.32\textwidth}
    \centering
    \includegraphics[width=\textwidth]{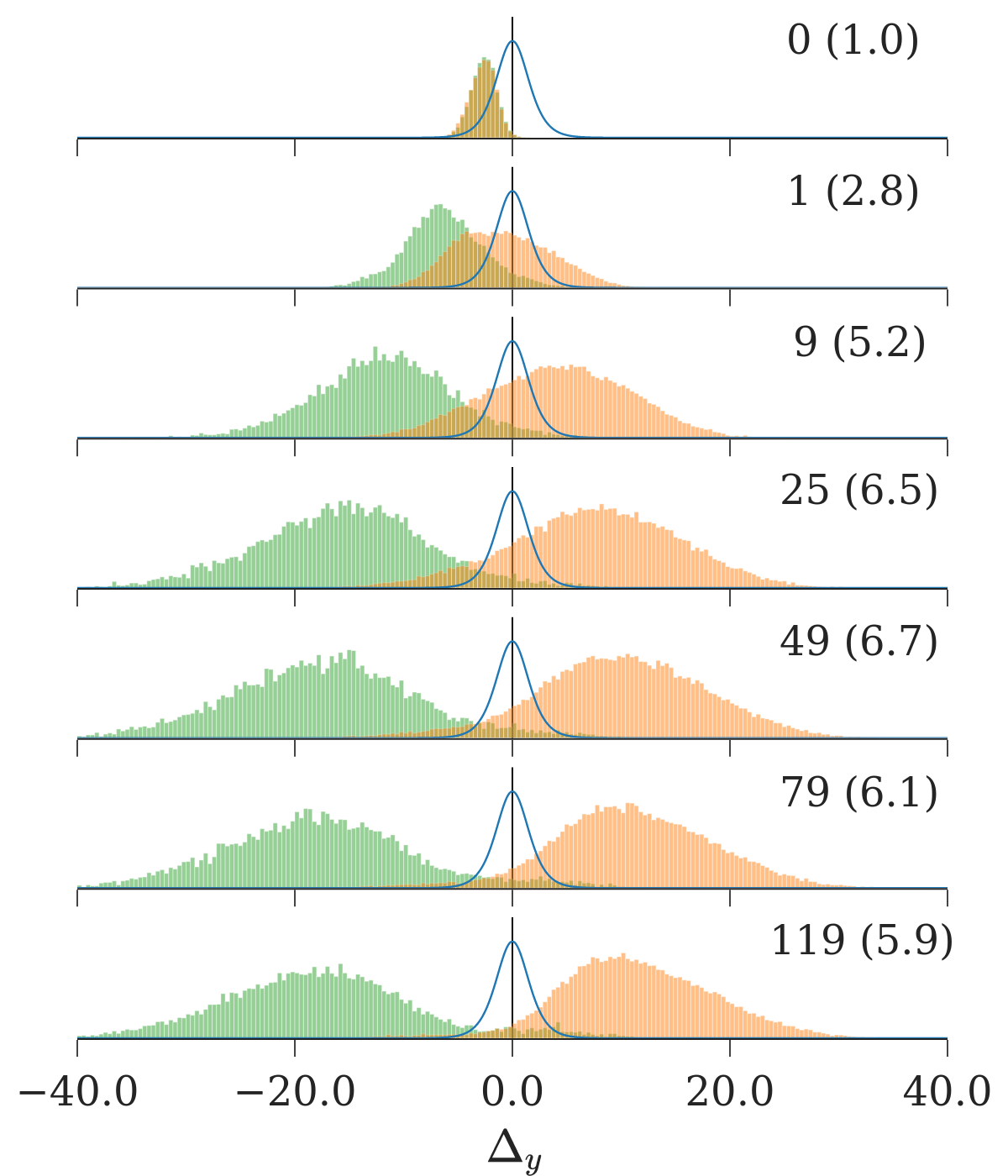}
    \caption{MAE, Sym., 0.2: 86.71}
  \end{subfigure}

  \begin{subfigure}[b]{0.32\textwidth}
    \centering
    \includegraphics[width=\textwidth]{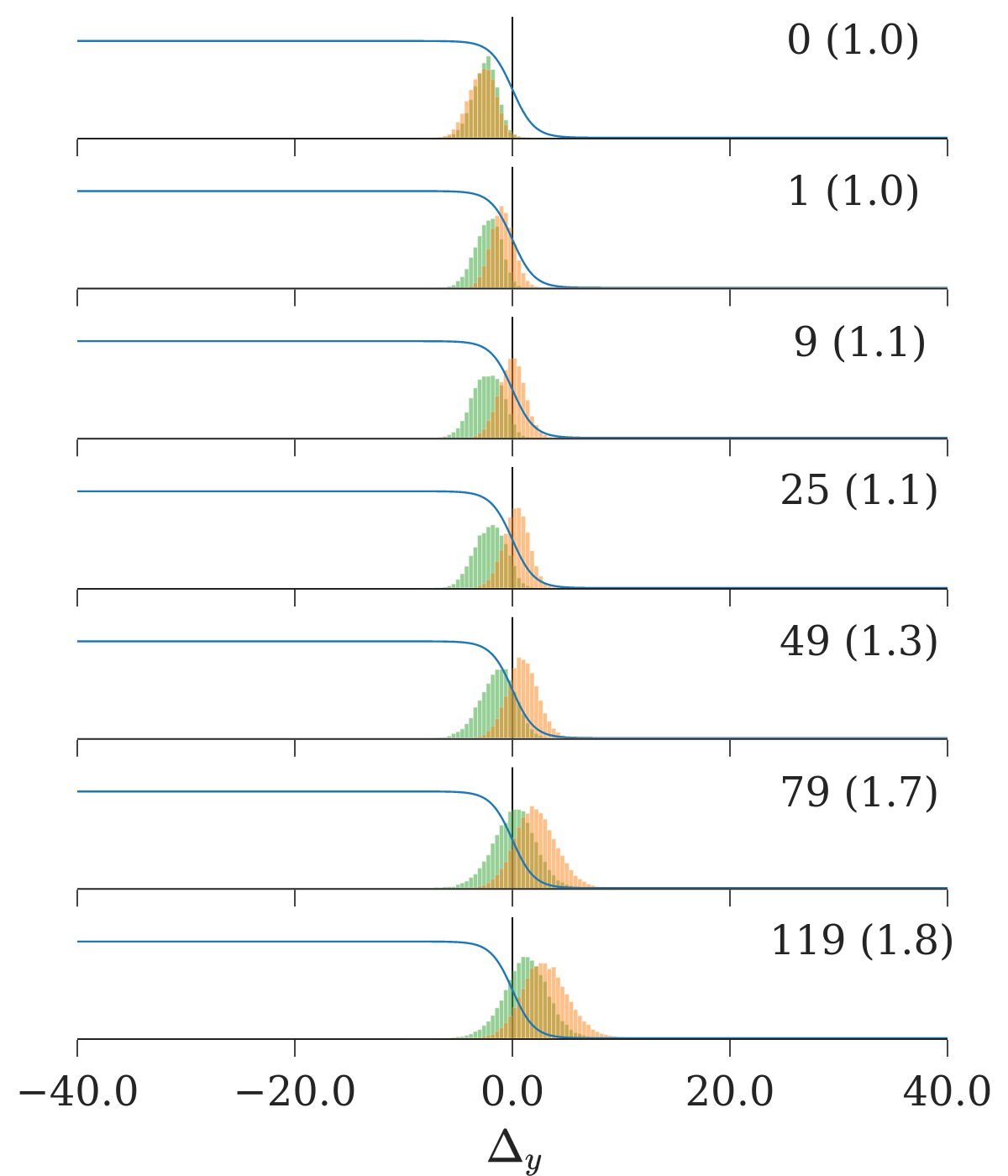}
    \caption{CE, Human, 0.4: 61.12}
  \end{subfigure}
  \hfill
  \begin{subfigure}[b]{0.32\textwidth}
    \centering
    \includegraphics[width=\textwidth]{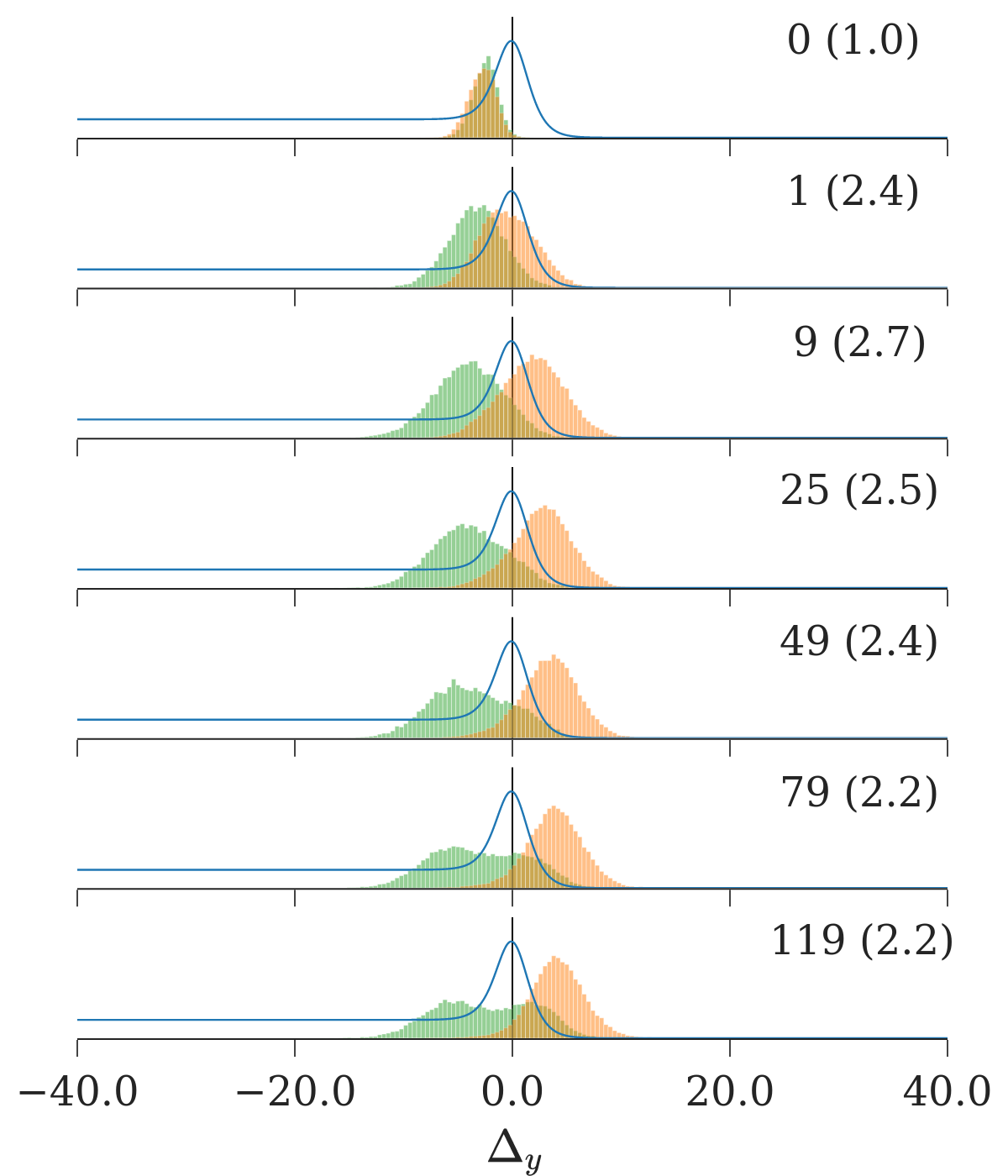}
    \caption{SCE, Human, 0.4: 72.56}
  \end{subfigure}
  \hfill
  \begin{subfigure}[b]{0.32\textwidth}
    \centering
    \includegraphics[width=\textwidth]{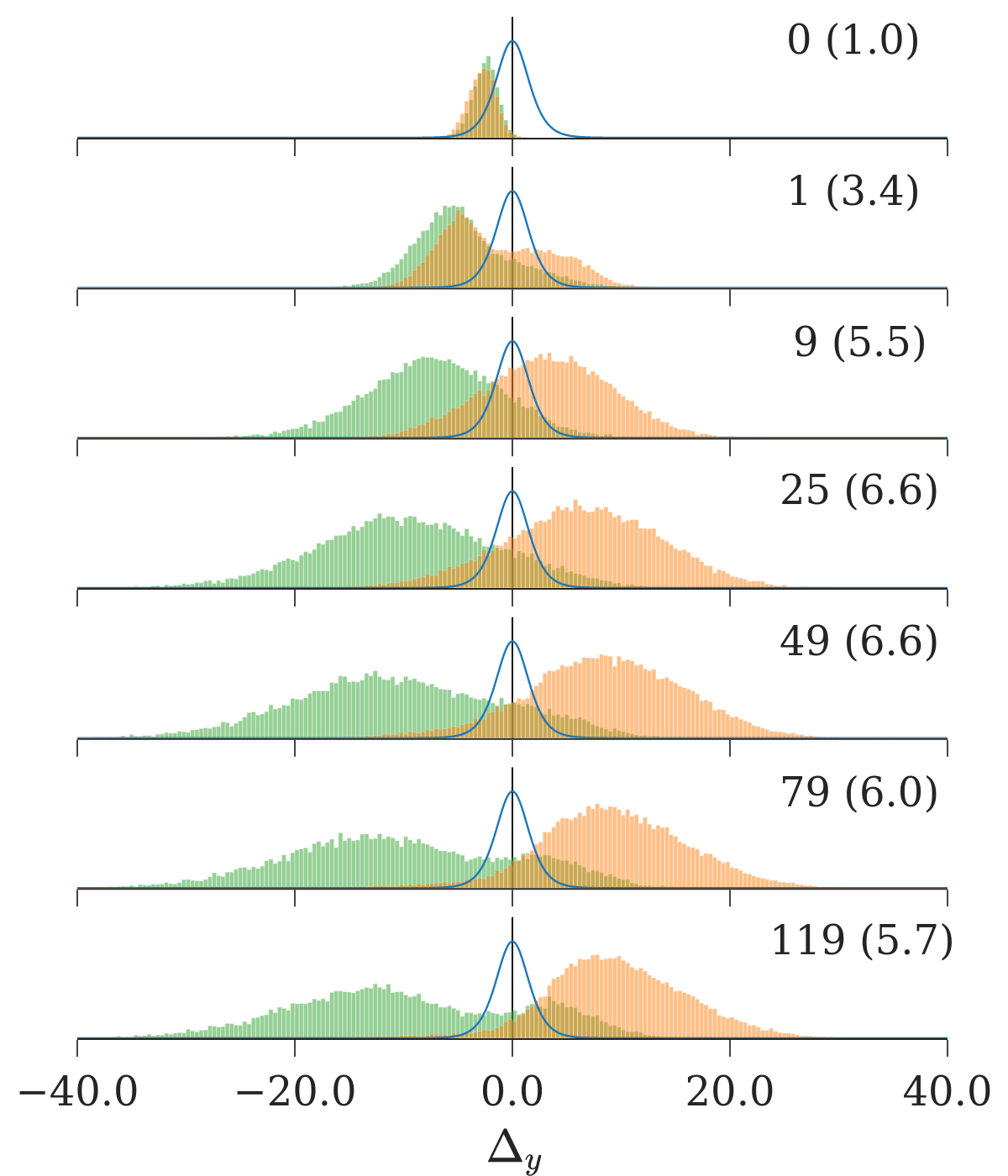}
    \caption{MAE, Human, 0.4: 78.99}
  \end{subfigure}

  \begin{subfigure}[b]{0.32\textwidth}
    \centering
    \includegraphics[width=\textwidth]{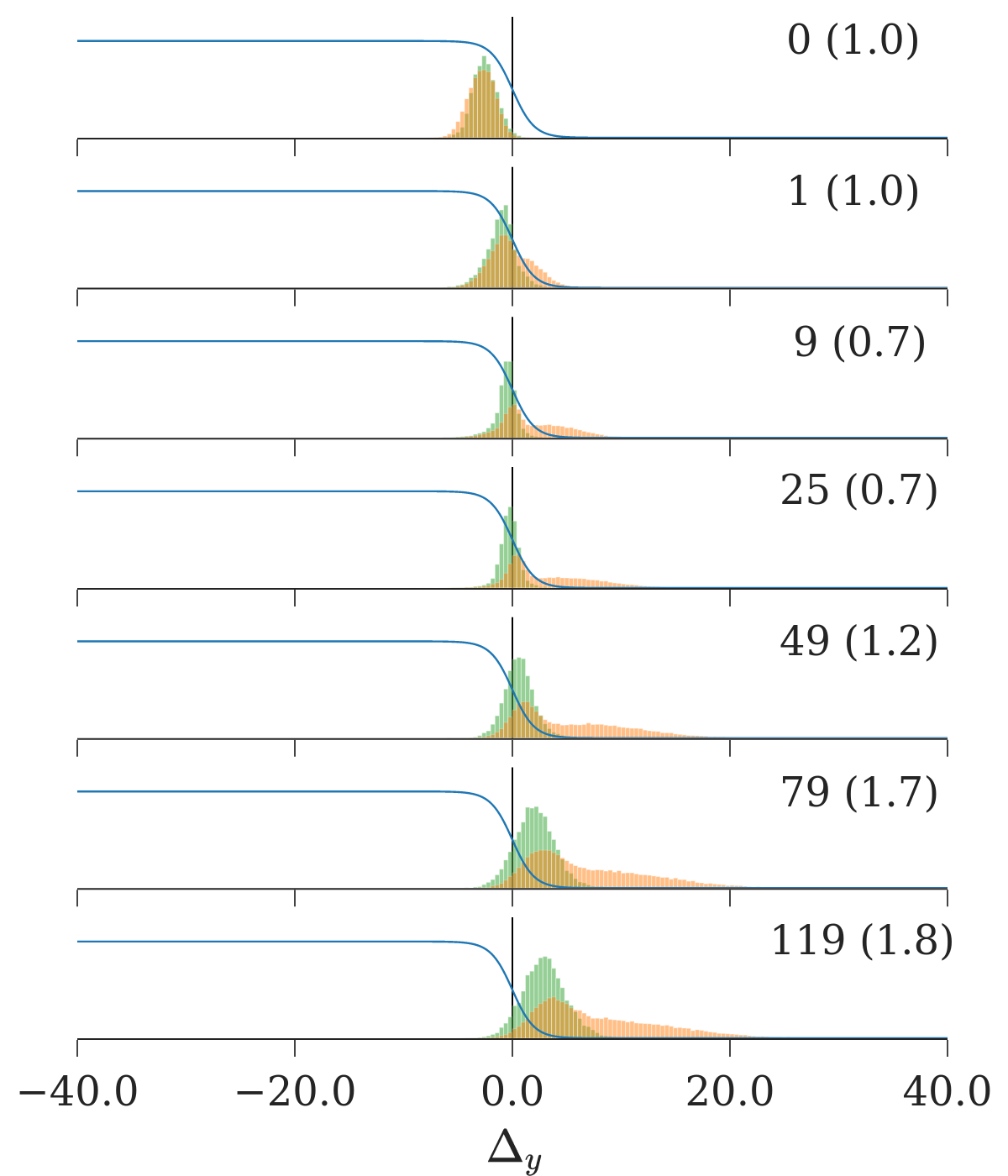}
    \caption{CE, Asym., 0.4: 73.75}
  \end{subfigure}
  \hfill
  \begin{subfigure}[b]{0.32\textwidth}
    \centering
    \includegraphics[width=\textwidth]{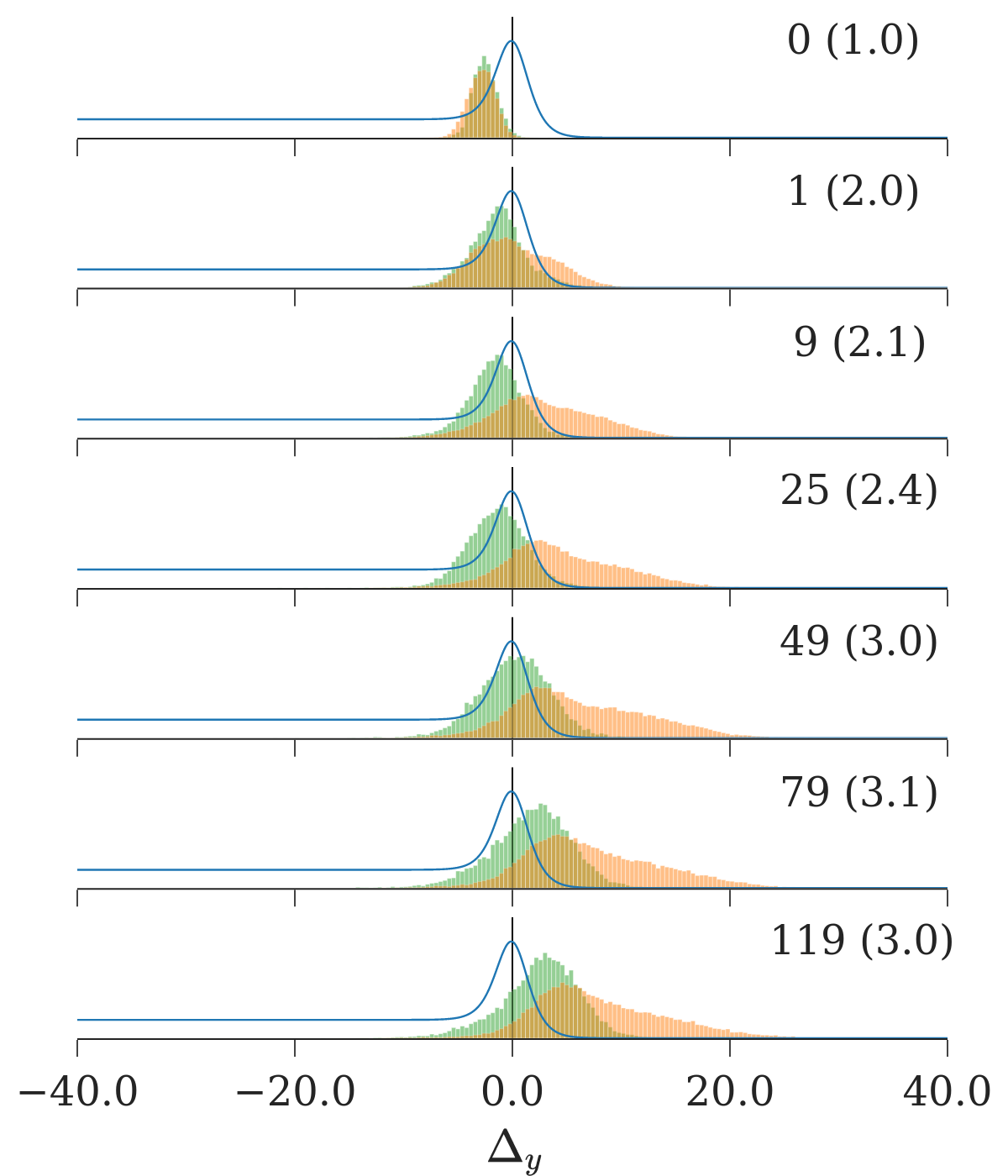}
    \caption{SCE, Asym., 0.4: 72.44}
  \end{subfigure}
  \hfill
  \begin{subfigure}[b]{0.32\textwidth}
    \centering
    \includegraphics[width=\textwidth]{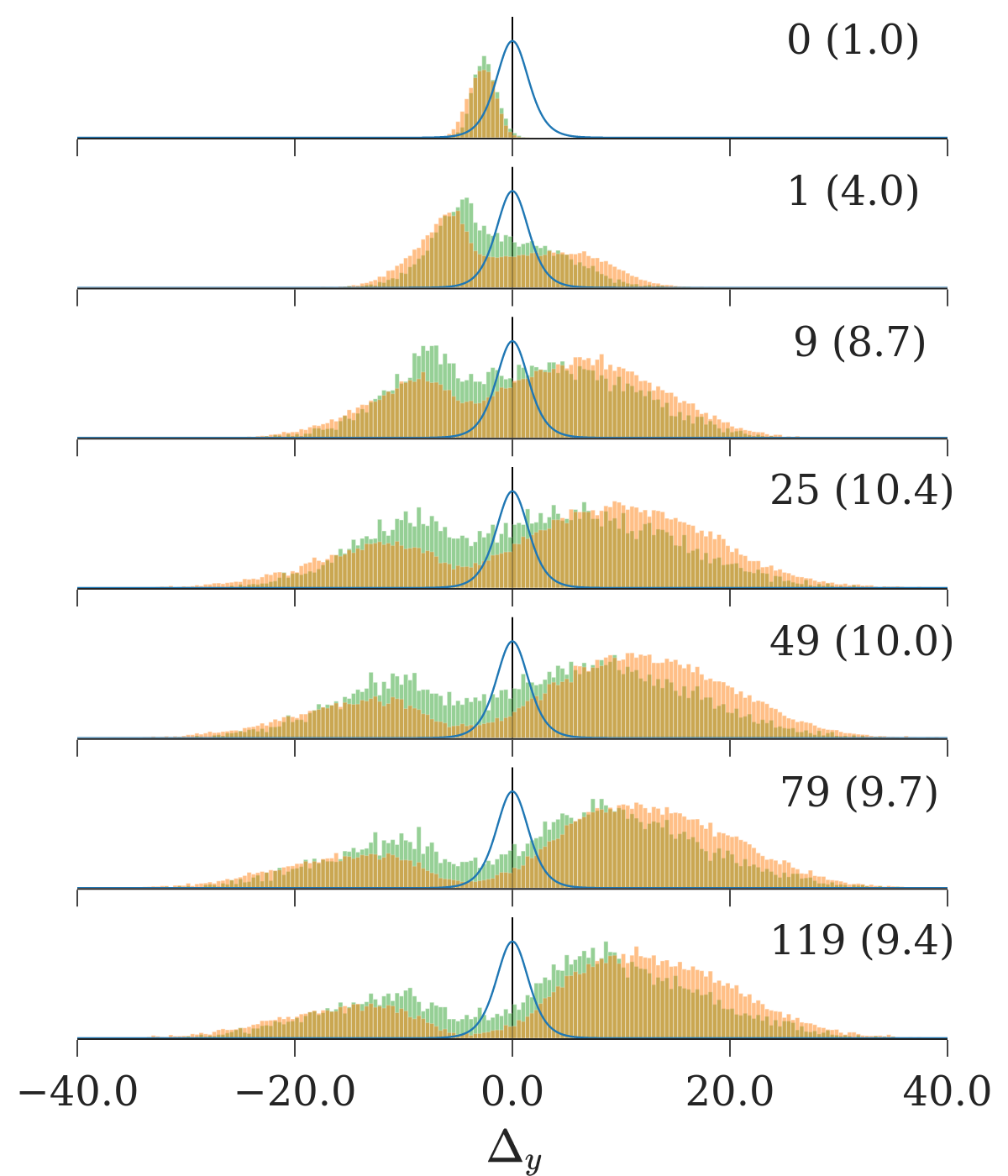}
    \caption{MAE, Asym., 0.4: 61.11}
  \end{subfigure}

  \caption{Additional results to \cref{fig:robust} with different label noise: (a-c) symmetric label noise with $\eta=0.2$; (d-f) human label noise with $\eta=0.4$; (g-i) asymmetric label noise with $\eta=0.4$. Noisy samples are colored green (on the left) and clean samples are orange (on the right). Test accuracies are included in the caption for reference.}
  \label{fig:robust_extend}
\end{figure}

\begin{figure}[t]
  \centering
  \begin{subfigure}[b]{0.32\textwidth}
    \centering
    \includegraphics[width=\textwidth]{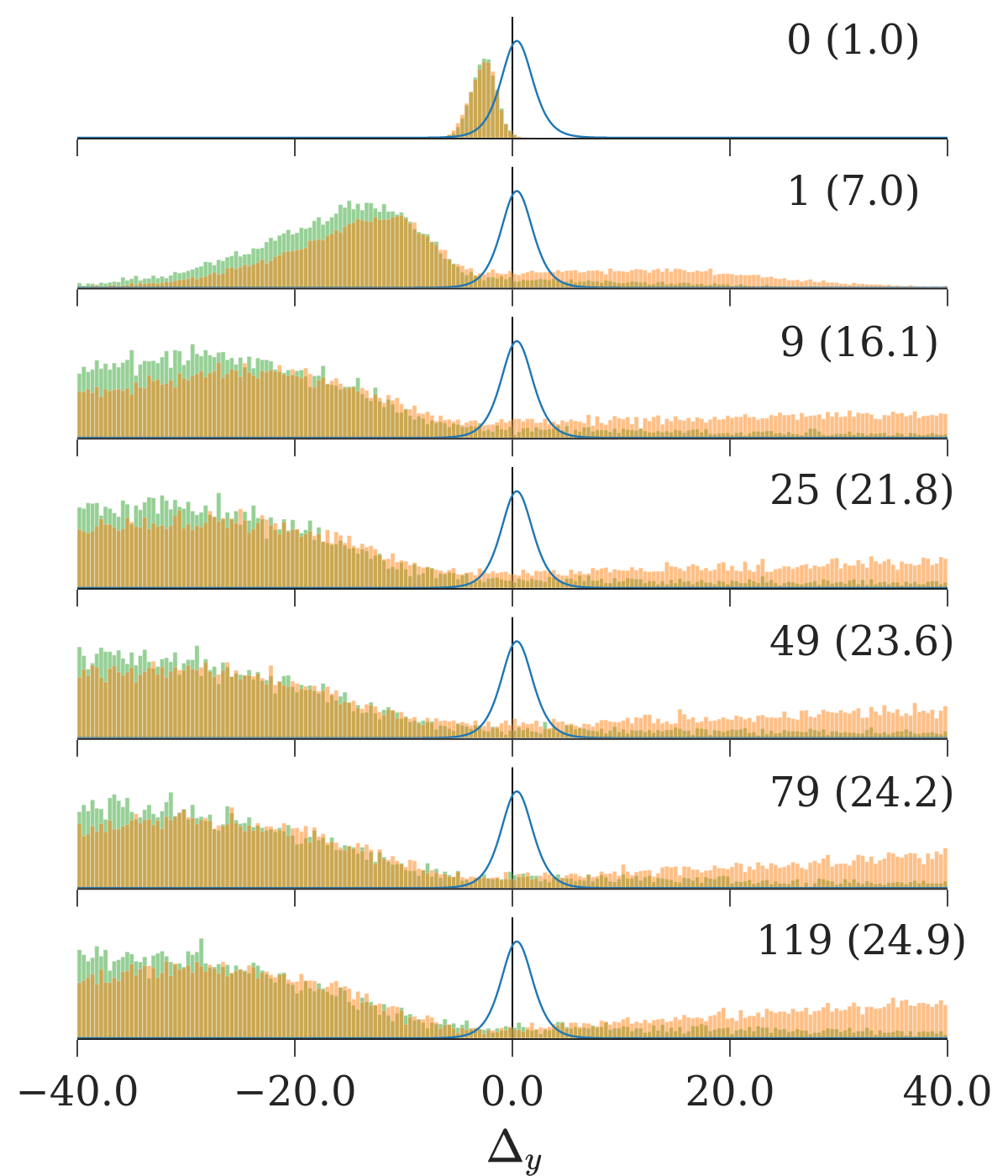}
    \caption{AGCE, Sym., 0.4: 44.52}
  \end{subfigure}
  \hfill
  \begin{subfigure}[b]{0.32\textwidth}
    \centering
    \includegraphics[width=\textwidth]{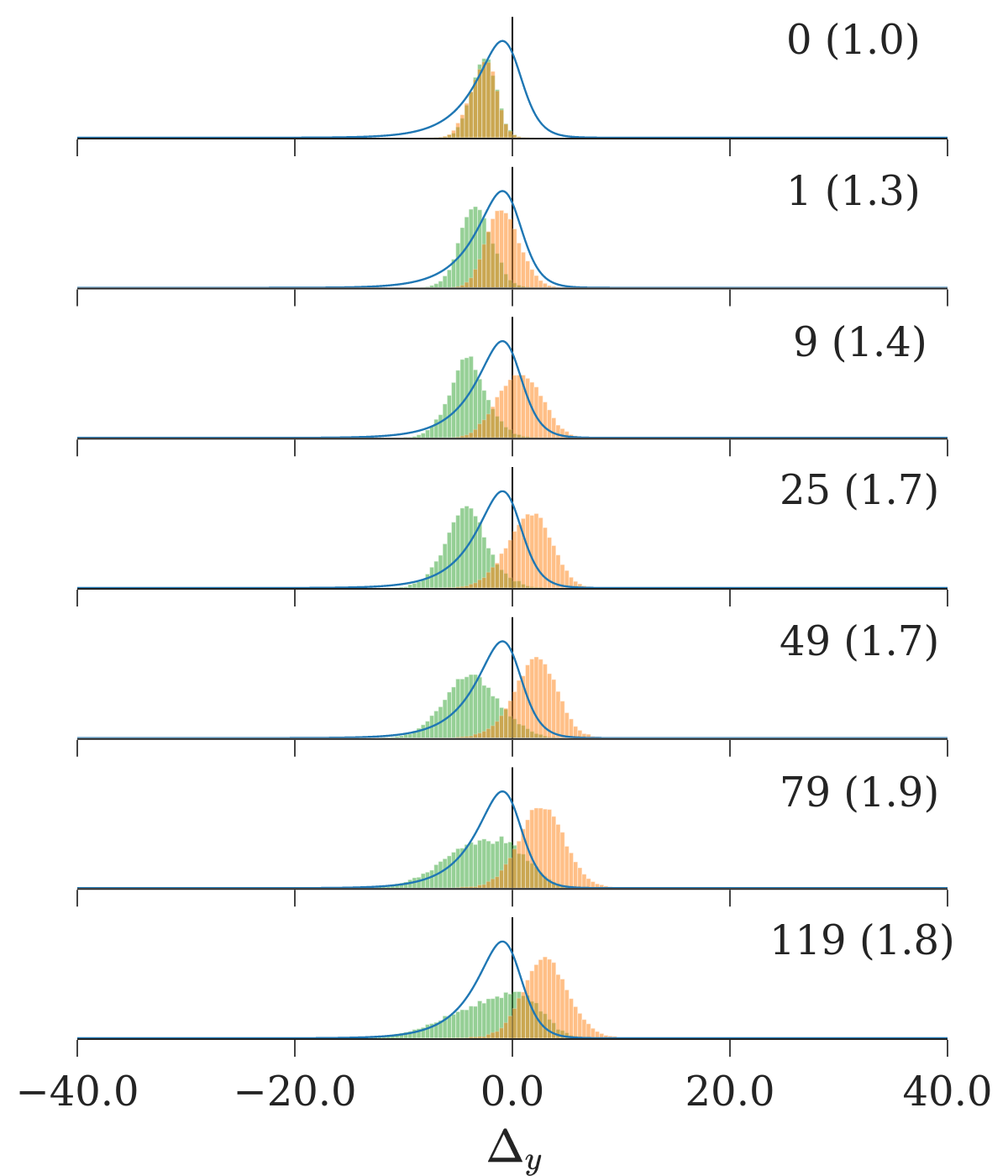}
    \caption{GCE, Sym., 0.4: 66.60}
  \end{subfigure}
  \hfill
  \begin{subfigure}[b]{0.32\textwidth}
    \centering
    \includegraphics[width=\textwidth]{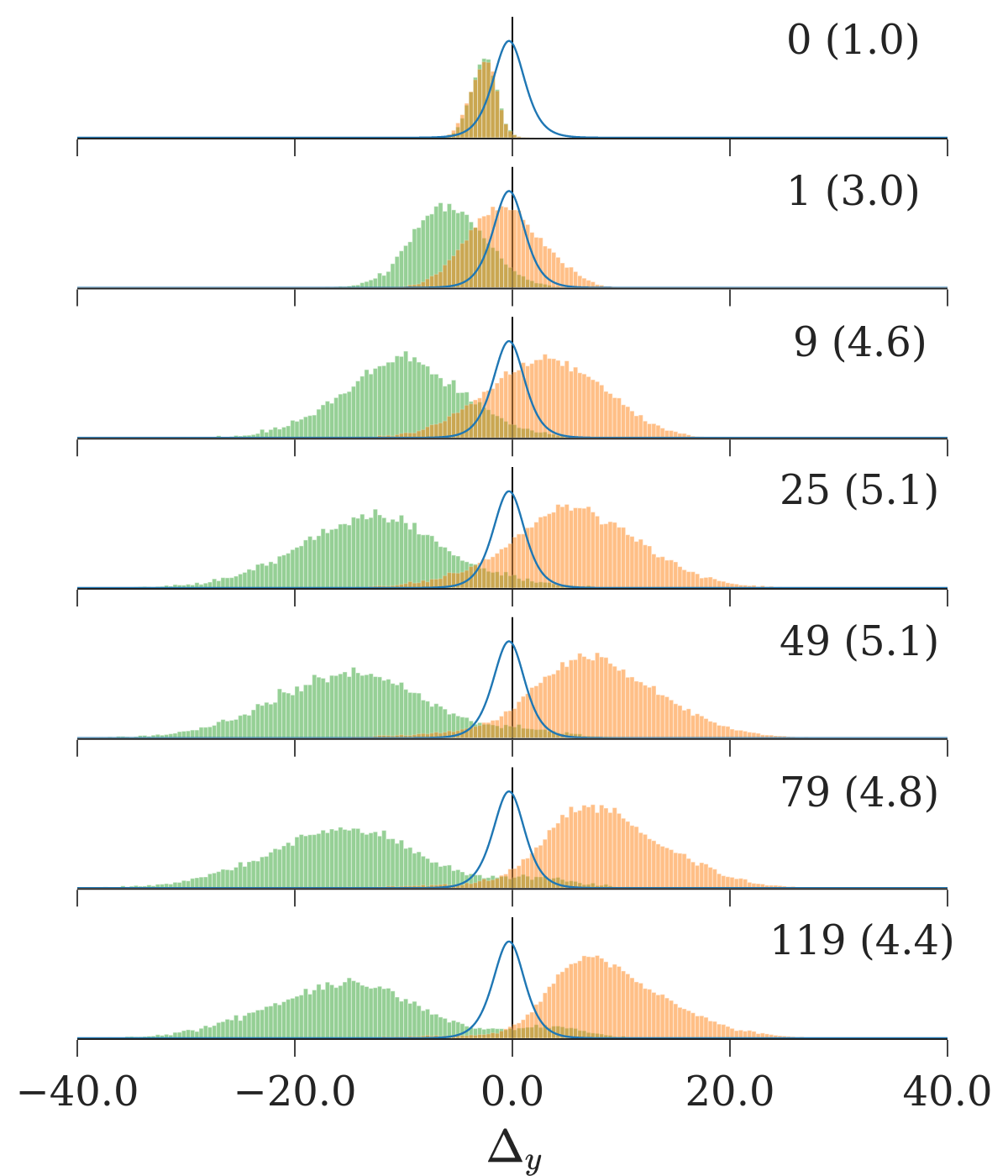}
    \caption{AUL, Sym., 0.4: 84.12}
  \end{subfigure}

  \begin{subfigure}[b]{0.32\textwidth}
    \centering
    \includegraphics[width=\textwidth]{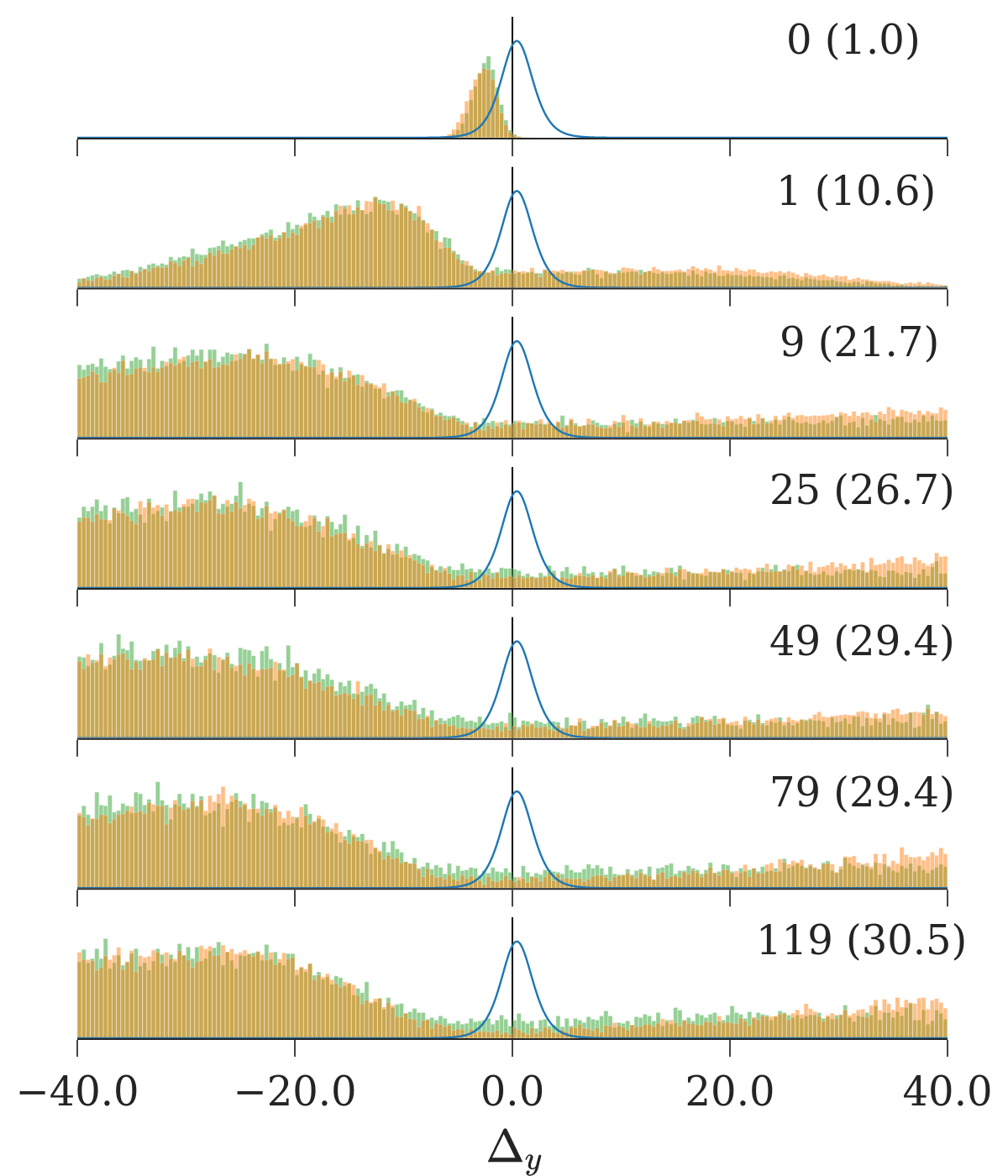}
    \caption{AGCE, Human, 0.4: 35.73}
  \end{subfigure}
  \hfill
  \begin{subfigure}[b]{0.32\textwidth}
    \centering
    \includegraphics[width=\textwidth]{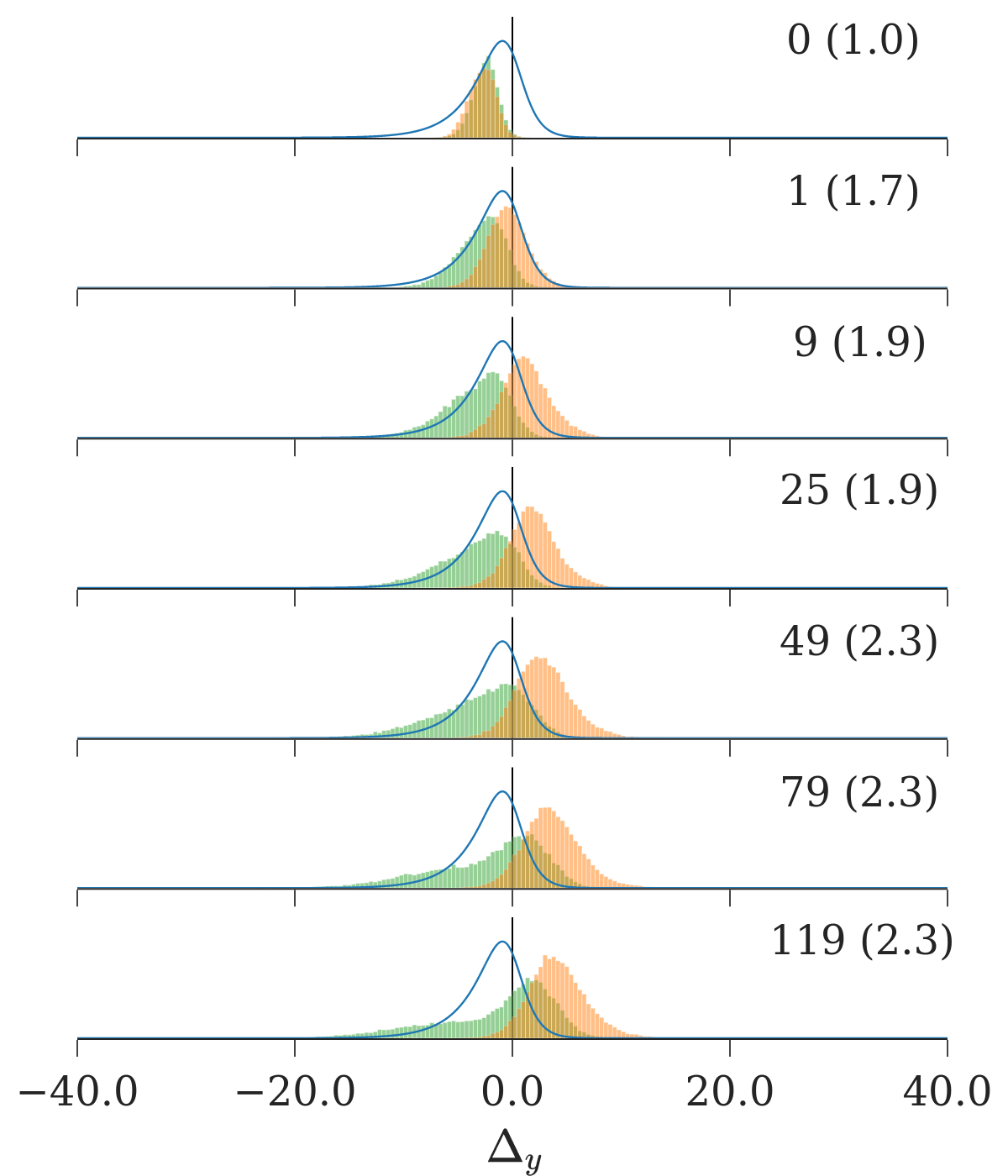}
    \caption{GCE, Human, 0.4: 68.22}
  \end{subfigure}
  \hfill
  \begin{subfigure}[b]{0.32\textwidth}
    \centering
    \includegraphics[width=\textwidth]{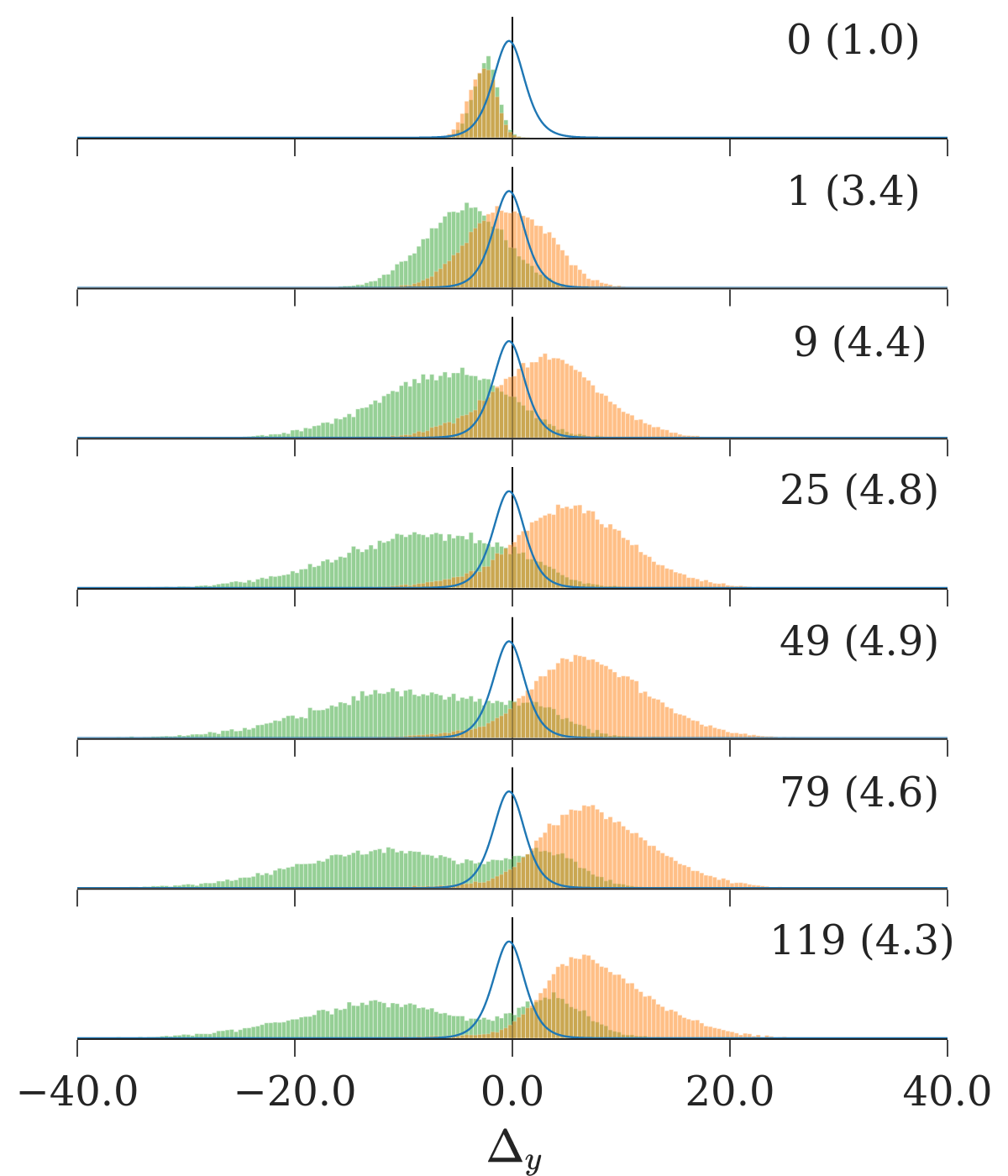}
    \caption{AUL, Human, 0.4: 77.22}
  \end{subfigure}

  \begin{subfigure}[b]{0.32\textwidth}
    \centering
    \includegraphics[width=\textwidth]{fig/cifar10_extract_human_agce.png}
    \caption{AGCE, Asym., 0.4: 46.75}
  \end{subfigure}
  \hfill
  \begin{subfigure}[b]{0.32\textwidth}
    \centering
    \includegraphics[width=\textwidth]{fig/cifar10_extract_human_gce.png}
    \caption{GCE, Asym., 0.4: 73.23}
  \end{subfigure}
  \hfill
  \begin{subfigure}[b]{0.32\textwidth}
    \centering
    \includegraphics[width=\textwidth]{fig/cifar10_extract_human_aul.png}
    \caption{AUL, Asym., 0.4: 67.72}
  \end{subfigure}

  \caption{Additional results to \cref{fig:robust} with more robust loss functions under different label noise: (a-c) symmetric label noise with $\eta=0.4$; (d-f) human label noise with $\eta=0.4$; (g-i) asymmetric label noise with $\eta=0.4$. Test accuracies are included in the caption for reference. Noisy samples are colored green (on the left) and clean samples are orange (on the right). Hyperparameters of these loss functions are selected for broad coverage rather than optimal performance.}
  \label{fig:robust_extend2}
\end{figure}

\textbf{Changes of $\Delta_y$ distributions with different label noise and loss functions\ \ } Complementing \cref{fig:robust}, in \cref{fig:robust_extend} we plot how distributions of $\Delta_y$ change during training on CIFAR10 with additional types of label noise using hyperparameters in \cref{table:robust_param}. They follow similar trends as in \cref{fig:robust}, thus supporting analysis in \cref{sec:dynamics:robust}. As MAE is not robust against asymmetric label noise with high $\eta$ \citep{mae}, it results in inferior performance. We also include results with additional loss functions in \cref{fig:robust_extend2}. Since optimal hyperparameters will result in similar sample-weighting functions, we choose hyperparameters for broad coverage of $w(\Delta_y)$ to better understand how they affect robustness.

\begin{table}[t]
  \centering
  \small
  \begin{tabular}{ c | c  c  c  c   }
    & AUL & AGCE & GCE & SCE \\
    \midrule
    $a$ & 2.0 & 3.0 & / & /  \\
    $q$ & 2.0& 4.0& 0.4& 0.95\\

  \end{tabular}
  \caption{Hyperparameters of different loss functions for results in \cref{sec:dynamics:robust,app:dynamics:robust}. They are selected for broad coverage of shapes, scales and horizontal locations of sample-weighting functions instead of optimal performance on CIFAR10.}
  \label{table:robust_param}
\end{table}

\end{document}

%% file: math_commands.tex
\usepackage{amsmath,amsfonts,bm}

\def\eqref#1{equation~\ref{#1}}

\def\1{\bm{1}}

\DeclareMathAlphabet{\mathsfit}{\encodingdefault}{\sfdefault}{m}{sl}
\SetMathAlphabet{\mathsfit}{bold}{\encodingdefault}{\sfdefault}{bx}{n}

\DeclareMathOperator*{\argmax}{arg\,max}
\DeclareMathOperator*{\argmin}{arg\,min}